\documentclass{article}
\usepackage{iclr2027_conference,times}
\iclrfinalcopy %

\usepackage{amsmath,amsthm,amssymb}
\usepackage{booktabs,multirow,array,tabularx,rotating}
\newcolumntype{Y}{>{\raggedleft\arraybackslash}X}
\usepackage{graphicx}
\usepackage{xcolor}
\usepackage{colortbl}
\usepackage{tikz}
\usetikzlibrary{arrows.meta,positioning,shapes.geometric,calc}
\usepackage{algorithm}
\usepackage{algpseudocode}
\usepackage{hyperref}
\hypersetup{colorlinks=true,linkcolor=blue!55!black,citecolor=green!40!black,urlcolor=blue!55!black}
\usepackage{url}
\usepackage[capitalize,noabbrev]{cleveref}
\usepackage{caption}
\usepackage{enumitem}
\usepackage{wrapfig,placeins,float}
\makeatletter
\renewcommand\section{\@startsection{section}{1}{\z@}{-2.0ex plus -1.5ex minus -.5ex}{1.5ex plus 1.2ex minus .3ex}{\large\sc\raggedright}}
\renewcommand\subsection{\@startsection{subsection}{2}{\z@}{-1.8ex plus -1.2ex minus -.4ex}{0.8ex plus .8ex minus .2ex}{\normalsize\sc\raggedright}}
\makeatother

\graphicspath{{figures/}}

\newtheorem{proposition}{Proposition}
\newtheorem{remark}{Remark}
\newcommand{\pol}{\pi}

\providecommand{\E}{}\renewcommand{\E}{\mathbb{E}}
\definecolor{okgreen}{HTML}{1a7a4d}
\definecolor{violred}{HTML}{c0392b}
\definecolor{passblue}{HTML}{E6EFFB}
\definecolor{failred}{HTML}{FBE7EA}
\definecolor{passgreen}{HTML}{E4F3EA}
\newcommand{\gatecell}[1]{\edef\trbgatetmp{#1}\def\trbgatepass{pass}\ifx\trbgatetmp\trbgatepass\cellcolor{passgreen}\else\cellcolor{failred}\fi\trbnew{#1}}
\makeatletter
\newcommand{\gatecellend}[1]{\ifnum\pdfstrcmp{#1}{pass}=0 \expandafter\@firstoftwo\else\expandafter\@secondoftwo\fi
  {\multicolumn{1}{>{\columncolor{passgreen}[\tabcolsep][0pt]}c@{}}{\trbnew{#1}}}%
  {\multicolumn{1}{>{\columncolor{failred}[\tabcolsep][0pt]}c@{}}{\trbnew{#1}}}}
\makeatother
\newcommand{\reachcell}[2]{\ifnum#1=#2\relax\cellcolor{passgreen}\else\ifnum#1=0\relax\cellcolor{failred}\fi\fi #1/#2}
\newcommand{\RQ}[1]{\vspace{2pt}\noindent\textbf{RQ#1}}
\newcommand{\rv}[1]{#1}
\newcommand{\pend}{[pending]}

\providecommand{\TrbUngatedScore}{}\renewcommand{\TrbUngatedScore}{0.18}%
\providecommand{\TrbUngatedProbe}{}\renewcommand{\TrbUngatedProbe}{0.48}%
\providecommand{\TrbUngatedProbeLat}{}\renewcommand{\TrbUngatedProbeLat}{1.29}%
\providecommand{\TrbLocoSelections}{}\renewcommand{\TrbLocoSelections}{four}%
\providecommand{\TrbPairedFolds}{}\renewcommand{\TrbPairedFolds}{$0.00$, $1.02$, $\ensuremath{-}0.71$ and $0.91$}%
\providecommand{\TrbGatedScoreMin}{}\renewcommand{\TrbGatedScoreMin}{0.00}%
\providecommand{\TrbGatedScoreMax}{}\renewcommand{\TrbGatedScoreMax}{0.18}%
\providecommand{\TrbRandomMin}{}\renewcommand{\TrbRandomMin}{0.66}%
\providecommand{\TrbRandomMax}{}\renewcommand{\TrbRandomMax}{4.34}%
\providecommand{\TrbGatedProbeMin}{}\renewcommand{\TrbGatedProbeMin}{0.67}%
\providecommand{\TrbGatedProbeMax}{}\renewcommand{\TrbGatedProbeMax}{1.69}%
\providecommand{\TrbRefitMin}{}\renewcommand{\TrbRefitMin}{0.00}%
\providecommand{\TrbRefitMax}{}\renewcommand{\TrbRefitMax}{0.65}%
\providecommand{\TrbInfeasibleAdvantage}{}\renewcommand{\TrbInfeasibleAdvantage}{3.45}%
\providecommand{\TrbScoreFirstViolProfiles}{}\renewcommand{\TrbScoreFirstViolProfiles}{2}%
\providecommand{\TrbChampionViolProfiles}{}\renewcommand{\TrbChampionViolProfiles}{4}%
\providecommand{\TrbNProfiles}{}\renewcommand{\TrbNProfiles}{4}%
\providecommand{\TrbNRuns}{}\renewcommand{\TrbNRuns}{four}%
\providecommand{\TrbNPlans}{}\renewcommand{\TrbNPlans}{622}%
\providecommand{\TrbInferPfifty}{}\renewcommand{\TrbInferPfifty}{1.03}%
\providecommand{\TrbInferPninetyfiveLo}{}\renewcommand{\TrbInferPninetyfiveLo}{1.05}%
\providecommand{\TrbInferPninetyfiveHi}{}\renewcommand{\TrbInferPninetyfiveHi}{1.06}%
\providecommand{\TrbCyclePfifty}{}\renewcommand{\TrbCyclePfifty}{1.50}%
\providecommand{\TrbCyclePninetyfiveLo}{}\renewcommand{\TrbCyclePninetyfiveLo}{1.55}%
\providecommand{\TrbCyclePninetyfiveHi}{}\renewcommand{\TrbCyclePninetyfiveHi}{1.59}%
\providecommand{\TrbCycleMin}{}\renewcommand{\TrbCycleMin}{1.40}%
\providecommand{\TrbPreInferPfifty}{}\renewcommand{\TrbPreInferPfifty}{0.12}%
\providecommand{\TrbPostInferPfifty}{}\renewcommand{\TrbPostInferPfifty}{0.34}%
\providecommand{\TrbOutsideModel}{}\renewcommand{\TrbOutsideModel}{0.46}%
\providecommand{\TrbTransportPfifty}{}\renewcommand{\TrbTransportPfifty}{7}%
\providecommand{\TrbSamplerPeriod}{}\renewcommand{\TrbSamplerPeriod}{0.5}%
\providecommand{\TrbTicksPerCycle}{}\renewcommand{\TrbTicksPerCycle}{three}%
\providecommand{\TrbTickShare}{}\renewcommand{\TrbTickShare}{99.5}%
\providecommand{\TrbProbeGpuhLo}{}\renewcommand{\TrbProbeGpuhLo}{0.34}%
\providecommand{\TrbProbeGpuhHi}{}\renewcommand{\TrbProbeGpuhHi}{0.60}%
\providecommand{\TrbProbeNModels}{}\renewcommand{\TrbProbeNModels}{three}%
\providecommand{\TrbTimingGpuMin}{}\renewcommand{\TrbTimingGpuMin}{32}%
\providecommand{\TrbTimingNModels}{}\renewcommand{\TrbTimingNModels}{twelve}%
\providecommand{\TrbRrB}{}\renewcommand{\TrbRrB}{1{,}000}%
\providecommand{\TrbRrBlock}{}\renewcommand{\TrbRrBlock}{20}%
\providecommand{\TrbRrTiDva}{}\renewcommand{\TrbRrTiDva}{0.65 to 0.75}%
\providecommand{\TrbRrRgDva}{}\renewcommand{\TrbRrRgDva}{0.37 to 0.51}%
\providecommand{\TrbRrTiRc}{}\renewcommand{\TrbRrTiRc}{0.80 to 0.88}%
\providecommand{\TrbRrRgRc}{}\renewcommand{\TrbRrRgRc}{0.63 to 0.82}%
\providecommand{\TrbRrTiWz}{}\renewcommand{\TrbRrTiWz}{1.31 to 1.47}%
\providecommand{\TrbRrRgWz}{}\renewcommand{\TrbRrRgWz}{1.08 to 1.28}%
\providecommand{\TrbRrTiWf}{}\renewcommand{\TrbRrTiWf}{0.67 to 0.75}%
\providecommand{\TrbRrRgWf}{}\renewcommand{\TrbRrRgWf}{0.45 to 0.55}%
\providecommand{\TrbRrAgRc}{}\renewcommand{\TrbRrAgRc}{0.161 to 0.198}%
\providecommand{\TrbRrAjRc}{}\renewcommand{\TrbRrAjRc}{0.123 to 0.147}%
\providecommand{\TrbEtwoNTurn}{}\renewcommand{\TrbEtwoNTurn}{40}%
\providecommand{\TrbEtwoNStraight}{}\renewcommand{\TrbEtwoNStraight}{120}%
\providecommand{\TrbEtwoTcpRatioMid}{}\renewcommand{\TrbEtwoTcpRatioMid}{4.37}%
\providecommand{\TrbEtwoTcpRatioMidLo}{}\renewcommand{\TrbEtwoTcpRatioMidLo}{3.69}%
\providecommand{\TrbEtwoTcpRatioMidHi}{}\renewcommand{\TrbEtwoTcpRatioMidHi}{5.16}%
\providecommand{\TrbEtwoTcpRatioMin}{}\renewcommand{\TrbEtwoTcpRatioMin}{3.06}%
\providecommand{\TrbEtwoTcpRatioMax}{}\renewcommand{\TrbEtwoTcpRatioMax}{4.79}%
\providecommand{\TrbEtwoSlRatioMid}{}\renewcommand{\TrbEtwoSlRatioMid}{8.50}%
\providecommand{\TrbEtwoSlRatioMidLo}{}\renewcommand{\TrbEtwoSlRatioMidLo}{4.88}%
\providecommand{\TrbEtwoSlRatioMidHi}{}\renewcommand{\TrbEtwoSlRatioMidHi}{15.76}%
\providecommand{\TrbEtwoSlRatioMin}{}\renewcommand{\TrbEtwoSlRatioMin}{2.39}%
\providecommand{\TrbEtwoSlRatioMax}{}\renewcommand{\TrbEtwoSlRatioMax}{8.50}%
\providecommand{\TrbEtwoSlNoise}{}\renewcommand{\TrbEtwoSlNoise}{0.148}%
\providecommand{\TrbEtwoSlNoiseLo}{}\renewcommand{\TrbEtwoSlNoiseLo}{0.087}%
\providecommand{\TrbEtwoSlNoiseHi}{}\renewcommand{\TrbEtwoSlNoiseHi}{0.222}%
\providecommand{\TrbEtwoSlDownMin}{}\renewcommand{\TrbEtwoSlDownMin}{0.08}%
\providecommand{\TrbEtwoSlDownMax}{}\renewcommand{\TrbEtwoSlDownMax}{0.18}%
\providecommand{\TrbEfiveLtfNTrips}{}\renewcommand{\TrbEfiveLtfNTrips}{four}%
\providecommand{\TrbEfiveLtfNPlans}{}\renewcommand{\TrbEfiveLtfNPlans}{23{,}939}%
\providecommand{\TrbEfiveLtfEtePfifty}{}\renewcommand{\TrbEfiveLtfEtePfifty}{0.77}%
\providecommand{\TrbEfiveLtfEtePninetyfive}{}\renewcommand{\TrbEfiveLtfEtePninetyfive}{0.86}%
\providecommand{\TrbEfiveLtfEteNewPninetyfive}{}\renewcommand{\TrbEfiveLtfEteNewPninetyfive}{0.81}%
\providecommand{\TrbEfiveLtfNOver}{}\renewcommand{\TrbEfiveLtfNOver}{two}%
\providecommand{\TrbEfiveLtfInferPfifty}{}\renewcommand{\TrbEfiveLtfInferPfifty}{0.06}%
\providecommand{\TrbEfiveLtfPrePfifty}{}\renewcommand{\TrbEfiveLtfPrePfifty}{0.70}%
\providecommand{\TrbEfiveLtfDeliveryPfifty}{}\renewcommand{\TrbEfiveLtfDeliveryPfifty}{0.63}%
\providecommand{\TrbEfiveLtfGapPfifty}{}\renewcommand{\TrbEfiveLtfGapPfifty}{0.09}%
\providecommand{\TrbEfivePkLtfN}{}\renewcommand{\TrbEfivePkLtfN}{5{,}617}%
\providecommand{\TrbEfivePkLtfPfifty}{}\renewcommand{\TrbEfivePkLtfPfifty}{0.72}%
\providecommand{\TrbEfivePkLtfPninetyfive}{}\renewcommand{\TrbEfivePkLtfPninetyfive}{0.772}%
\providecommand{\TrbEfivePkLtfOver}{}\renewcommand{\TrbEfivePkLtfOver}{1}%
\providecommand{\TrbEfivePkLtfInfer}{}\renewcommand{\TrbEfivePkLtfInfer}{0.05}%
\providecommand{\TrbEfivePkLtfOld}{}\renewcommand{\TrbEfivePkLtfOld}{--}%
\providecommand{\TrbEfivePkLtfGate}{}\renewcommand{\TrbEfivePkLtfGate}{pass}%
\providecommand{\TrbEfivePkLwamN}{}\renewcommand{\TrbEfivePkLwamN}{2{,}870}%
\providecommand{\TrbEfivePkLwamPfifty}{}\renewcommand{\TrbEfivePkLwamPfifty}{0.82}%
\providecommand{\TrbEfivePkLwamPninetyfive}{}\renewcommand{\TrbEfivePkLwamPninetyfive}{0.884}%
\providecommand{\TrbEfivePkLwamOver}{}\renewcommand{\TrbEfivePkLwamOver}{0}%
\providecommand{\TrbEfivePkLwamInfer}{}\renewcommand{\TrbEfivePkLwamInfer}{0.11}%
\providecommand{\TrbEfivePkLwamOld}{}\renewcommand{\TrbEfivePkLwamOld}{--}%
\providecommand{\TrbEfivePkLwamGate}{}\renewcommand{\TrbEfivePkLwamGate}{pass}%
\providecommand{\TrbEfivePkWzN}{}\renewcommand{\TrbEfivePkWzN}{310}%
\providecommand{\TrbEfivePkWzPfifty}{}\renewcommand{\TrbEfivePkWzPfifty}{0.88}%
\providecommand{\TrbEfivePkWzPninetyfive}{}\renewcommand{\TrbEfivePkWzPninetyfive}{1.007}%
\providecommand{\TrbEfivePkWzOver}{}\renewcommand{\TrbEfivePkWzOver}{7}%
\providecommand{\TrbEfivePkWzInfer}{}\renewcommand{\TrbEfivePkWzInfer}{0.56}%
\providecommand{\TrbEfivePkWzOld}{}\renewcommand{\TrbEfivePkWzOld}{2.01}%
\providecommand{\TrbEfivePkWzGate}{}\renewcommand{\TrbEfivePkWzGate}{pass}%
\providecommand{\TrbEfivePkRcN}{}\renewcommand{\TrbEfivePkRcN}{562}%
\providecommand{\TrbEfivePkRcPfifty}{}\renewcommand{\TrbEfivePkRcPfifty}{1.15}%
\providecommand{\TrbEfivePkRcPninetyfive}{}\renewcommand{\TrbEfivePkRcPninetyfive}{1.204}%
\providecommand{\TrbEfivePkRcOver}{}\renewcommand{\TrbEfivePkRcOver}{38}%
\providecommand{\TrbEfivePkRcInfer}{}\renewcommand{\TrbEfivePkRcInfer}{0.74}%
\providecommand{\TrbEfivePkRcOld}{}\renewcommand{\TrbEfivePkRcOld}{--}%
\providecommand{\TrbEfivePkRcGate}{}\renewcommand{\TrbEfivePkRcGate}{fail}%
\providecommand{\TrbEfivePkDvaN}{}\renewcommand{\TrbEfivePkDvaN}{291}%
\providecommand{\TrbEfivePkDvaPfifty}{}\renewcommand{\TrbEfivePkDvaPfifty}{1.52}%
\providecommand{\TrbEfivePkDvaPninetyfive}{}\renewcommand{\TrbEfivePkDvaPninetyfive}{1.580}%
\providecommand{\TrbEfivePkDvaOver}{}\renewcommand{\TrbEfivePkDvaOver}{291}%
\providecommand{\TrbEfivePkDvaInfer}{}\renewcommand{\TrbEfivePkDvaInfer}{1.03}%
\providecommand{\TrbEfivePkDvaOld}{}\renewcommand{\TrbEfivePkDvaOld}{3.58}%
\providecommand{\TrbEfivePkDvaGate}{}\renewcommand{\TrbEfivePkDvaGate}{fail}%
\providecommand{\TrbEfivePkAlpN}{}\renewcommand{\TrbEfivePkAlpN}{326}%
\providecommand{\TrbEfivePkAlpPfifty}{}\renewcommand{\TrbEfivePkAlpPfifty}{1.97}%
\providecommand{\TrbEfivePkAlpPninetyfive}{}\renewcommand{\TrbEfivePkAlpPninetyfive}{2.102}%
\providecommand{\TrbEfivePkAlpOver}{}\renewcommand{\TrbEfivePkAlpOver}{326}%
\providecommand{\TrbEfivePkAlpInfer}{}\renewcommand{\TrbEfivePkAlpInfer}{1.62}%
\providecommand{\TrbEfivePkAlpOld}{}\renewcommand{\TrbEfivePkAlpOld}{2.39}%
\providecommand{\TrbEfivePkAlpGate}{}\renewcommand{\TrbEfivePkAlpGate}{fail}%
\providecommand{\TrbEfivePkLtfLab}{}\renewcommand{\TrbEfivePkLtfLab}{0.07}%
\providecommand{\TrbEfivePkLtfGateInf}{}\renewcommand{\TrbEfivePkLtfGateInf}{pass}%
\providecommand{\TrbEfivePkLtfPdms}{}\renewcommand{\TrbEfivePkLtfPdms}{84.14}%
\providecommand{\TrbEfivePkLwamLab}{}\renewcommand{\TrbEfivePkLwamLab}{--}%
\providecommand{\TrbEfivePkLwamGateInf}{}\renewcommand{\TrbEfivePkLwamGateInf}{pass}%
\providecommand{\TrbEfivePkLwamPdms}{}\renewcommand{\TrbEfivePkLwamPdms}{--}%
\providecommand{\TrbEfivePkWzLab}{}\renewcommand{\TrbEfivePkWzLab}{0.49}%
\providecommand{\TrbEfivePkWzGateInf}{}\renewcommand{\TrbEfivePkWzGateInf}{pass}%
\providecommand{\TrbEfivePkWzPdms}{}\renewcommand{\TrbEfivePkWzPdms}{90.30}%
\providecommand{\TrbEfivePkRcLab}{}\renewcommand{\TrbEfivePkRcLab}{0.51}%
\providecommand{\TrbEfivePkRcGateInf}{}\renewcommand{\TrbEfivePkRcGateInf}{pass}%
\providecommand{\TrbEfivePkRcPdms}{}\renewcommand{\TrbEfivePkRcPdms}{90.42}%
\providecommand{\TrbEfivePkDvaLab}{}\renewcommand{\TrbEfivePkDvaLab}{0.95}%
\providecommand{\TrbEfivePkDvaGateInf}{}\renewcommand{\TrbEfivePkDvaGateInf}{pass}%
\providecommand{\TrbEfivePkDvaPdms}{}\renewcommand{\TrbEfivePkDvaPdms}{91.22}%
\providecommand{\TrbEfivePkAlpLab}{}\renewcommand{\TrbEfivePkAlpLab}{0.96}%
\providecommand{\TrbEfivePkAlpGateInf}{}\renewcommand{\TrbEfivePkAlpGateInf}{fail}%
\providecommand{\TrbEfivePkAlpPdms}{}\renewcommand{\TrbEfivePkAlpPdms}{--}%
\providecommand{\TrbEfivePkNPass}{}\renewcommand{\TrbEfivePkNPass}{three}%
\providecommand{\TrbEfivePkRcMarginMs}{}\renewcommand{\TrbEfivePkRcMarginMs}{4}%
\providecommand{\TrbEfivePkWzOldFrames}{}\renewcommand{\TrbEfivePkWzOldFrames}{two}%
\providecommand{\TrbEthreeNArms}{}\renewcommand{\TrbEthreeNArms}{16}%
\providecommand{\TrbEthreeNmax}{}\renewcommand{\TrbEthreeNmax}{20{,}000}%
\providecommand{\TrbEthreeNmid}{}\renewcommand{\TrbEthreeNmid}{5{,}000}%
\providecommand{\TrbEthreeNativeTurn}{}\renewcommand{\TrbEthreeNativeTurn}{0.85}%
\providecommand{\TrbEthreePSixMean}{}\renewcommand{\TrbEthreePSixMean}{0.032}%
\providecommand{\TrbEthreeS}{}\renewcommand{\TrbEthreeS}{0.026}%
\providecommand{\TrbEthreePThree}{}\renewcommand{\TrbEthreePThree}{0.063}%
\providecommand{\TrbEthreeKeepsNR}{}\renewcommand{\TrbEthreeKeepsNR}{$0.6$ and $0.3$}%
\providecommand{\TrbEthreeKeepsWithin}{}\renewcommand{\TrbEthreeKeepsWithin}{$0.9$, $0.75$ and $0.45$}%
\providecommand{\TrbEfourNArms}{}\renewcommand{\TrbEfourNArms}{seven}%
\providecommand{\TrbEfourNUsable}{}\renewcommand{\TrbEfourNUsable}{18{,}063}%
\providecommand{\TrbEfourNVision}{}\renewcommand{\TrbEfourNVision}{96}%
\providecommand{\TrbEfourS}{}\renewcommand{\TrbEfourS}{0.0075}%
\providecommand{\TrbEfourRepairA}{}\renewcommand{\TrbEfourRepairA}{0.117}%
\providecommand{\TrbEfourRepairB}{}\renewcommand{\TrbEfourRepairB}{0.125}%
\providecommand{\TrbEfourResSixA}{}\renewcommand{\TrbEfourResSixA}{0.022}%
\providecommand{\TrbEfourResSixB}{}\renewcommand{\TrbEfourResSixB}{0.026}%
\providecommand{\TrbEfourShareSixA}{}\renewcommand{\TrbEfourShareSixA}{19}%
\providecommand{\TrbEfourShareSixB}{}\renewcommand{\TrbEfourShareSixB}{21}%
\providecommand{\TrbEfourLoSixA}{}\renewcommand{\TrbEfourLoSixA}{\ensuremath{-}0.0010}%
\providecommand{\TrbEfourLoSixB}{}\renewcommand{\TrbEfourLoSixB}{\ensuremath{-}0.0005}%
\providecommand{\TrbEfourCfSixA}{}\renewcommand{\TrbEfourCfSixA}{0.900}%
\providecommand{\TrbEfourCfSixB}{}\renewcommand{\TrbEfourCfSixB}{0.882}%
\providecommand{\TrbEfourConcLo}{}\renewcommand{\TrbEfourConcLo}{1.5}%
\providecommand{\TrbEfourConcHi}{}\renewcommand{\TrbEfourConcHi}{3.0}%
\providecommand{\TrbEfourStraightSixLo}{}\renewcommand{\TrbEfourStraightSixLo}{8}%
\providecommand{\TrbEfourStraightSixHi}{}\renewcommand{\TrbEfourStraightSixHi}{14}%
\providecommand{\TrbEfourResSevenFive}{}\renewcommand{\TrbEfourResSevenFive}{0.041}%
\providecommand{\TrbEfourShareSevenFive}{}\renewcommand{\TrbEfourShareSevenFive}{35}%
\providecommand{\TrbEfourLoSevenFive}{}\renewcommand{\TrbEfourLoSevenFive}{0.017}%
\providecommand{\TrbEfourResFourFive}{}\renewcommand{\TrbEfourResFourFive}{0.055}%
\providecommand{\TrbEfourShareFourFive}{}\renewcommand{\TrbEfourShareFourFive}{47}%
\providecommand{\TrbEfourLoFourFive}{}\renewcommand{\TrbEfourLoFourFive}{0.014}%
\providecommand{\TrbEfourCfFourFive}{}\renewcommand{\TrbEfourCfFourFive}{0.837}%
\providecommand{\TrbEfourResThree}{}\renewcommand{\TrbEfourResThree}{0.090}%
\providecommand{\TrbEfourShareThree}{}\renewcommand{\TrbEfourShareThree}{77}%
\providecommand{\TrbEfourLoThree}{}\renewcommand{\TrbEfourLoThree}{0.044}%
\providecommand{\TrbEfourCfThree}{}\renewcommand{\TrbEfourCfThree}{0.757}%
\providecommand{\TrbEfourDamageSevenFive}{}\renewcommand{\TrbEfourDamageSevenFive}{0.087}%
\providecommand{\TrbEsevenNTrain}{}\renewcommand{\TrbEsevenNTrain}{2{,}667}%
\providecommand{\TrbEsevenNHeld}{}\renewcommand{\TrbEsevenNHeld}{992}%
\providecommand{\TrbEsevenNTurn}{}\renewcommand{\TrbEsevenNTurn}{282}%
\providecommand{\TrbEsevenNNovel}{}\renewcommand{\TrbEsevenNNovel}{653}%
\providecommand{\TrbEsevenNSeeds}{}\renewcommand{\TrbEsevenNSeeds}{three}%
\providecommand{\TrbEsevenAdeZs}{}\renewcommand{\TrbEsevenAdeZs}{1.12}%
\providecommand{\TrbEsevenAdeFtLo}{}\renewcommand{\TrbEsevenAdeFtLo}{0.84}%
\providecommand{\TrbEsevenAdeFtHi}{}\renewcommand{\TrbEsevenAdeFtHi}{0.88}%
\providecommand{\TrbEsevenDeltaLo}{}\renewcommand{\TrbEsevenDeltaLo}{\ensuremath{-}0.27}%
\providecommand{\TrbEsevenDeltaHi}{}\renewcommand{\TrbEsevenDeltaHi}{\ensuremath{-}0.24}%
\providecommand{\TrbEsevenDeltaUpMax}{}\renewcommand{\TrbEsevenDeltaUpMax}{\ensuremath{-}0.11}%
\providecommand{\TrbEsevenNovelDeltaLo}{}\renewcommand{\TrbEsevenNovelDeltaLo}{\ensuremath{-}0.20}%
\providecommand{\TrbEsevenNovelDeltaHi}{}\renewcommand{\TrbEsevenNovelDeltaHi}{\ensuremath{-}0.18}%
\providecommand{\TrbEsevenNovelUpMax}{}\renewcommand{\TrbEsevenNovelUpMax}{\ensuremath{-}0.04}%
\providecommand{\TrbEsevenTurnDeltaLo}{}\renewcommand{\TrbEsevenTurnDeltaLo}{\ensuremath{-}0.32}%
\providecommand{\TrbEsevenTurnDeltaHi}{}\renewcommand{\TrbEsevenTurnDeltaHi}{\ensuremath{-}0.26}%
\providecommand{\TrbEsevenTurnNIncl}{}\renewcommand{\TrbEsevenTurnNIncl}{one}%
\providecommand{\TrbEsevenSpread}{}\renewcommand{\TrbEsevenSpread}{0.036}%
\providecommand{\TrbEsevenGammaRatio}{}\renewcommand{\TrbEsevenGammaRatio}{0.66}%
\providecommand{\TrbEsevenJpegRatio}{}\renewcommand{\TrbEsevenJpegRatio}{0.66}%
\providecommand{\TrbEsevenBlindRatio}{}\renewcommand{\TrbEsevenBlindRatio}{0.99}%
\providecommand{\TrbEsevenTrainMin}{}\renewcommand{\TrbEsevenTrainMin}{16}%
\providecommand{\TrbEsthreeLtfTi}{}\renewcommand{\TrbEsthreeLtfTi}{0.66}%
\providecommand{\TrbEsthreeLtfTiLo}{}\renewcommand{\TrbEsthreeLtfTiLo}{0.63}%
\providecommand{\TrbEsthreeLtfTiHi}{}\renewcommand{\TrbEsthreeLtfTiHi}{0.69}%
\providecommand{\TrbEsthreeLtfRg}{}\renewcommand{\TrbEsthreeLtfRg}{0.48--0.62}%
\providecommand{\TrbEsthreeLtfAg}{}\renewcommand{\TrbEsthreeLtfAg}{0.27}%
\providecommand{\TrbEsthreeLtfAj}{}\renewcommand{\TrbEsthreeLtfAj}{0.09}%
\providecommand{\TrbEsthreeLtfGray}{}\renewcommand{\TrbEsthreeLtfGray}{1.47}%
\providecommand{\TrbEsthreeLtfSign}{}\renewcommand{\TrbEsthreeLtfSign}{0.920}%
\providecommand{\TrbEsthreeLtfGate}{}\renewcommand{\TrbEsthreeLtfGate}{pass}%
\providecommand{\TrbEsthreeLtfPdms}{}\renewcommand{\TrbEsthreeLtfPdms}{84.14}%
\providecommand{\TrbEsthreeLtfRank}{}\renewcommand{\TrbEsthreeLtfRank}{challenger}%
\providecommand{\TrbEsthreeLwamTi}{}\renewcommand{\TrbEsthreeLwamTi}{0.61}%
\providecommand{\TrbEsthreeLwamTiLo}{}\renewcommand{\TrbEsthreeLwamTiLo}{0.57}%
\providecommand{\TrbEsthreeLwamTiHi}{}\renewcommand{\TrbEsthreeLwamTiHi}{0.64}%
\providecommand{\TrbEsthreeLwamRg}{}\renewcommand{\TrbEsthreeLwamRg}{--}%
\providecommand{\TrbEsthreeLwamAg}{}\renewcommand{\TrbEsthreeLwamAg}{0.25}%
\providecommand{\TrbEsthreeLwamAj}{}\renewcommand{\TrbEsthreeLwamAj}{0.03}%
\providecommand{\TrbEsthreeLwamGray}{}\renewcommand{\TrbEsthreeLwamGray}{1.72}%
\providecommand{\TrbEsthreeLwamSign}{}\renewcommand{\TrbEsthreeLwamSign}{0.919}%
\providecommand{\TrbEsthreeLwamGate}{}\renewcommand{\TrbEsthreeLwamGate}{pass}%
\providecommand{\TrbEsthreeLwamPdms}{}\renewcommand{\TrbEsthreeLwamPdms}{--}%
\providecommand{\TrbEsthreeLwamRank}{}\renewcommand{\TrbEsthreeLwamRank}{--}%
\providecommand{\TrbEsthreeWzTi}{}\renewcommand{\TrbEsthreeWzTi}{1.87}%
\providecommand{\TrbEsthreeWzTiLo}{}\renewcommand{\TrbEsthreeWzTiLo}{1.79}%
\providecommand{\TrbEsthreeWzTiHi}{}\renewcommand{\TrbEsthreeWzTiHi}{1.96}%
\providecommand{\TrbEsthreeWzRg}{}\renewcommand{\TrbEsthreeWzRg}{1.08--1.28}%
\providecommand{\TrbEsthreeWzAg}{}\renewcommand{\TrbEsthreeWzAg}{1.11}%
\providecommand{\TrbEsthreeWzAj}{}\renewcommand{\TrbEsthreeWzAj}{0.70}%
\providecommand{\TrbEsthreeWzGray}{}\renewcommand{\TrbEsthreeWzGray}{5.77}%
\providecommand{\TrbEsthreeWzSign}{}\renewcommand{\TrbEsthreeWzSign}{0.959}%
\providecommand{\TrbEsthreeWzGate}{}\renewcommand{\TrbEsthreeWzGate}{pass}%
\providecommand{\TrbEsthreeWzPdms}{}\renewcommand{\TrbEsthreeWzPdms}{90.30}%
\providecommand{\TrbEsthreeWzRank}{}\renewcommand{\TrbEsthreeWzRank}{selected}%
\providecommand{\TrbEsthreeRcTi}{}\renewcommand{\TrbEsthreeRcTi}{0.90}%
\providecommand{\TrbEsthreeRcTiLo}{}\renewcommand{\TrbEsthreeRcTiLo}{0.85}%
\providecommand{\TrbEsthreeRcTiHi}{}\renewcommand{\TrbEsthreeRcTiHi}{0.95}%
\providecommand{\TrbEsthreeRcRg}{}\renewcommand{\TrbEsthreeRcRg}{0.63--0.82}%
\providecommand{\TrbEsthreeRcAg}{}\renewcommand{\TrbEsthreeRcAg}{0.20}%
\providecommand{\TrbEsthreeRcAj}{}\renewcommand{\TrbEsthreeRcAj}{0.13}%
\providecommand{\TrbEsthreeRcGray}{}\renewcommand{\TrbEsthreeRcGray}{2.18}%
\providecommand{\TrbEsthreeRcSign}{}\renewcommand{\TrbEsthreeRcSign}{0.970}%
\providecommand{\TrbEsthreeRcGate}{}\renewcommand{\TrbEsthreeRcGate}{fail}%
\providecommand{\TrbEsthreeRcPdms}{}\renewcommand{\TrbEsthreeRcPdms}{90.42}%
\providecommand{\TrbEsthreeRcRank}{}\renewcommand{\TrbEsthreeRcRank}{--}%
\providecommand{\TrbEsthreeDvaTi}{}\renewcommand{\TrbEsthreeDvaTi}{0.96}%
\providecommand{\TrbEsthreeDvaTiLo}{}\renewcommand{\TrbEsthreeDvaTiLo}{0.91}%
\providecommand{\TrbEsthreeDvaTiHi}{}\renewcommand{\TrbEsthreeDvaTiHi}{1.02}%
\providecommand{\TrbEsthreeDvaRg}{}\renewcommand{\TrbEsthreeDvaRg}{0.37--0.51}%
\providecommand{\TrbEsthreeDvaAg}{}\renewcommand{\TrbEsthreeDvaAg}{0.26}%
\providecommand{\TrbEsthreeDvaAj}{}\renewcommand{\TrbEsthreeDvaAj}{0.18}%
\providecommand{\TrbEsthreeDvaGray}{}\renewcommand{\TrbEsthreeDvaGray}{2.90}%
\providecommand{\TrbEsthreeDvaSign}{}\renewcommand{\TrbEsthreeDvaSign}{0.985}%
\providecommand{\TrbEsthreeDvaGate}{}\renewcommand{\TrbEsthreeDvaGate}{fail}%
\providecommand{\TrbEsthreeDvaPdms}{}\renewcommand{\TrbEsthreeDvaPdms}{91.22}%
\providecommand{\TrbEsthreeDvaRank}{}\renewcommand{\TrbEsthreeDvaRank}{--}%
\providecommand{\TrbEsthreeNWin}{}\renewcommand{\TrbEsthreeNWin}{4{,}359}%
\providecommand{\TrbEsthreeNSamples}{}\renewcommand{\TrbEsthreeNSamples}{4{,}529}%
\providecommand{\TrbEsthreeMin}{}\renewcommand{\TrbEsthreeMin}{38}%
\providecommand{\TrbEsthreeNWinSept}{}\renewcommand{\TrbEsthreeNWinSept}{2{,}842}%
\providecommand{\TrbEsthreeSignMin}{}\renewcommand{\TrbEsthreeSignMin}{0.919}%
\providecommand{\TrbEsthreeSignMax}{}\renewcommand{\TrbEsthreeSignMax}{0.985}%
\providecommand{\TrbEsthreeNOutside}{}\renewcommand{\TrbEsthreeNOutside}{four}%
\providecommand{\TrbEoneTcpNatDs}{}\renewcommand{\TrbEoneTcpNatDs}{85.7}%
\providecommand{\TrbEoneTcpNatDsSd}{}\renewcommand{\TrbEoneTcpNatDsSd}{1.3}%
\providecommand{\TrbEoneTcpNatRc}{}\renewcommand{\TrbEoneTcpNatRc}{86.7}%
\providecommand{\TrbEoneTcpNatCrr}{}\renewcommand{\TrbEoneTcpNatCrr}{0.07}%
\providecommand{\TrbEoneTcpCropADs}{}\renewcommand{\TrbEoneTcpCropADs}{86.2}%
\providecommand{\TrbEoneTcpCropADsSd}{}\renewcommand{\TrbEoneTcpCropADsSd}{0.2}%
\providecommand{\TrbEoneTcpCropARc}{}\renewcommand{\TrbEoneTcpCropARc}{90.3}%
\providecommand{\TrbEoneTcpCropACrr}{}\renewcommand{\TrbEoneTcpCropACrr}{0.20}%
\providecommand{\TrbEoneTcpCropBDs}{}\renewcommand{\TrbEoneTcpCropBDs}{83.6}%
\providecommand{\TrbEoneTcpCropBDsSd}{}\renewcommand{\TrbEoneTcpCropBDsSd}{0.6}%
\providecommand{\TrbEoneTcpCropBRc}{}\renewcommand{\TrbEoneTcpCropBRc}{92.7}%
\providecommand{\TrbEoneTcpCropBCrr}{}\renewcommand{\TrbEoneTcpCropBCrr}{0.20}%
\providecommand{\TrbEoneTcpCropCDs}{}\renewcommand{\TrbEoneTcpCropCDs}{64.0}%
\providecommand{\TrbEoneTcpCropCDsSd}{}\renewcommand{\TrbEoneTcpCropCDsSd}{7.2}%
\providecommand{\TrbEoneTcpCropCRc}{}\renewcommand{\TrbEoneTcpCropCRc}{85.7}%
\providecommand{\TrbEoneTcpCropCCrr}{}\renewcommand{\TrbEoneTcpCropCCrr}{0.47}%
\providecommand{\TrbEoneTcpCropCCpk}{}\renewcommand{\TrbEoneTcpCropCCpk}{8.84}%
\providecommand{\TrbEoneTcpDownADs}{}\renewcommand{\TrbEoneTcpDownADs}{85.0}%
\providecommand{\TrbEoneTcpDownADsSd}{}\renewcommand{\TrbEoneTcpDownADsSd}{1.3}%
\providecommand{\TrbEoneTcpDownARc}{}\renewcommand{\TrbEoneTcpDownARc}{86.9}%
\providecommand{\TrbEoneTcpDownACrr}{}\renewcommand{\TrbEoneTcpDownACrr}{0.13}%
\providecommand{\TrbEoneTcpDownBDs}{}\renewcommand{\TrbEoneTcpDownBDs}{86.5}%
\providecommand{\TrbEoneTcpDownBDsSd}{}\renewcommand{\TrbEoneTcpDownBDsSd}{0.0}%
\providecommand{\TrbEoneTcpDownBRc}{}\renewcommand{\TrbEoneTcpDownBRc}{86.5}%
\providecommand{\TrbEoneTcpDownBCrr}{}\renewcommand{\TrbEoneTcpDownBCrr}{0.00}%
\providecommand{\TrbEoneTcpDownCDs}{}\renewcommand{\TrbEoneTcpDownCDs}{82.7}%
\providecommand{\TrbEoneTcpDownCDsSd}{}\renewcommand{\TrbEoneTcpDownCDsSd}{4.8}%
\providecommand{\TrbEoneTcpDownCRc}{}\renewcommand{\TrbEoneTcpDownCRc}{86.9}%
\providecommand{\TrbEoneTcpDownCCrr}{}\renewcommand{\TrbEoneTcpDownCCrr}{0.20}%
\providecommand{\TrbEoneTcpDownCCpk}{}\renewcommand{\TrbEoneTcpDownCCpk}{2.17}%
\providecommand{\TrbEoneSlNatDs}{}\renewcommand{\TrbEoneSlNatDs}{94.0}%
\providecommand{\TrbEoneSlNatDsSd}{}\renewcommand{\TrbEoneSlNatDsSd}{0.0}%
\providecommand{\TrbEoneSlNatRc}{}\renewcommand{\TrbEoneSlNatRc}{100.0}%
\providecommand{\TrbEoneSlNatCrr}{}\renewcommand{\TrbEoneSlNatCrr}{0.00}%
\providecommand{\TrbEoneSlCropADs}{}\renewcommand{\TrbEoneSlCropADs}{94.0}%
\providecommand{\TrbEoneSlCropADsSd}{}\renewcommand{\TrbEoneSlCropADsSd}{0.0}%
\providecommand{\TrbEoneSlCropARc}{}\renewcommand{\TrbEoneSlCropARc}{100.0}%
\providecommand{\TrbEoneSlCropACrr}{}\renewcommand{\TrbEoneSlCropACrr}{0.00}%
\providecommand{\TrbEoneSlCropBDs}{}\renewcommand{\TrbEoneSlCropBDs}{84.0}%
\providecommand{\TrbEoneSlCropBDsSd}{}\renewcommand{\TrbEoneSlCropBDsSd}{0.2}%
\providecommand{\TrbEoneSlCropBRc}{}\renewcommand{\TrbEoneSlCropBRc}{94.3}%
\providecommand{\TrbEoneSlCropBCrr}{}\renewcommand{\TrbEoneSlCropBCrr}{0.00}%
\providecommand{\TrbEoneSlCropCDs}{}\renewcommand{\TrbEoneSlCropCDs}{71.4}%
\providecommand{\TrbEoneSlCropCDsSd}{}\renewcommand{\TrbEoneSlCropCDsSd}{5.3}%
\providecommand{\TrbEoneSlCropCRc}{}\renewcommand{\TrbEoneSlCropCRc}{93.6}%
\providecommand{\TrbEoneSlCropCCrr}{}\renewcommand{\TrbEoneSlCropCCrr}{0.33}%
\providecommand{\TrbEoneSlCropCCpk}{}\renewcommand{\TrbEoneSlCropCCpk}{3.33}%
\providecommand{\TrbEoneSlDownADs}{}\renewcommand{\TrbEoneSlDownADs}{94.0}%
\providecommand{\TrbEoneSlDownADsSd}{}\renewcommand{\TrbEoneSlDownADsSd}{0.0}%
\providecommand{\TrbEoneSlDownARc}{}\renewcommand{\TrbEoneSlDownARc}{100.0}%
\providecommand{\TrbEoneSlDownACrr}{}\renewcommand{\TrbEoneSlDownACrr}{0.00}%
\providecommand{\TrbEoneSlDownBDs}{}\renewcommand{\TrbEoneSlDownBDs}{94.0}%
\providecommand{\TrbEoneSlDownBDsSd}{}\renewcommand{\TrbEoneSlDownBDsSd}{0.0}%
\providecommand{\TrbEoneSlDownBRc}{}\renewcommand{\TrbEoneSlDownBRc}{100.0}%
\providecommand{\TrbEoneSlDownBCrr}{}\renewcommand{\TrbEoneSlDownBCrr}{0.00}%
\providecommand{\TrbEoneSlDownCDs}{}\renewcommand{\TrbEoneSlDownCDs}{94.0}%
\providecommand{\TrbEoneSlDownCDsSd}{}\renewcommand{\TrbEoneSlDownCDsSd}{0.0}%
\providecommand{\TrbEoneSlDownCRc}{}\renewcommand{\TrbEoneSlDownCRc}{100.0}%
\providecommand{\TrbEoneSlDownCCrr}{}\renewcommand{\TrbEoneSlDownCCrr}{0.00}%
\providecommand{\TrbEoneSlDownCCpk}{}\renewcommand{\TrbEoneSlDownCCpk}{0.00}%
\providecommand{\TrbEoneAgree}{}\renewcommand{\TrbEoneAgree}{eight}%
\providecommand{\TrbEoneTies}{}\renewcommand{\TrbEoneTies}{one}%
\providecommand{\TrbEoneSignP}{}\renewcommand{\TrbEoneSignP}{0.020}%
\providecommand{\TrbEoneRhoTurn}{}\renewcommand{\TrbEoneRhoTurn}{0.59}%
\providecommand{\TrbEoneRhoTurnLo}{}\renewcommand{\TrbEoneRhoTurnLo}{0.07}%
\providecommand{\TrbEoneRhoTurnHi}{}\renewcommand{\TrbEoneRhoTurnHi}{0.82}%
\providecommand{\TrbEoneUnNatDs}{}\renewcommand{\TrbEoneUnNatDs}{66.5}%
\providecommand{\TrbEoneUnNatDsSd}{}\renewcommand{\TrbEoneUnNatDsSd}{7.9}%
\providecommand{\TrbEoneUnNatRc}{}\renewcommand{\TrbEoneUnNatRc}{90.4}%
\providecommand{\TrbEoneUnNatCrr}{}\renewcommand{\TrbEoneUnNatCrr}{0.47}%
\providecommand{\TrbEoneUnCropBDs}{}\renewcommand{\TrbEoneUnCropBDs}{58.2}%
\providecommand{\TrbEoneUnCropBDsSd}{}\renewcommand{\TrbEoneUnCropBDsSd}{3.2}%
\providecommand{\TrbEoneUnCropBRc}{}\renewcommand{\TrbEoneUnCropBRc}{77.2}%
\providecommand{\TrbEoneUnCropBCrr}{}\renewcommand{\TrbEoneUnCropBCrr}{0.40}%
\providecommand{\TrbEoneUnCropCDs}{}\renewcommand{\TrbEoneUnCropCDs}{65.8}%
\providecommand{\TrbEoneUnCropCDsSd}{}\renewcommand{\TrbEoneUnCropCDsSd}{9.3}%
\providecommand{\TrbEoneUnCropCRc}{}\renewcommand{\TrbEoneUnCropCRc}{98.1}%
\providecommand{\TrbEoneUnCropCCrr}{}\renewcommand{\TrbEoneUnCropCCrr}{0.53}%
\providecommand{\TrbEoneUnDownBDs}{}\renewcommand{\TrbEoneUnDownBDs}{65.0}%
\providecommand{\TrbEoneUnDownBDsSd}{}\renewcommand{\TrbEoneUnDownBDsSd}{8.0}%
\providecommand{\TrbEoneUnDownBRc}{}\renewcommand{\TrbEoneUnDownBRc}{87.8}%
\providecommand{\TrbEoneUnDownBCrr}{}\renewcommand{\TrbEoneUnDownBCrr}{0.40}%
\providecommand{\TrbEoneUnDownCDs}{}\renewcommand{\TrbEoneUnDownCDs}{67.8}%
\providecommand{\TrbEoneUnDownCDsSd}{}\renewcommand{\TrbEoneUnDownCDsSd}{10.8}%
\providecommand{\TrbEoneUnDownCRc}{}\renewcommand{\TrbEoneUnDownCRc}{90.5}%
\providecommand{\TrbEoneUnDownCCrr}{}\renewcommand{\TrbEoneUnDownCCrr}{0.40}%
\providecommand{\TrbEoneVaNatDs}{}\renewcommand{\TrbEoneVaNatDs}{65.6}%
\providecommand{\TrbEoneVaNatDsSd}{}\renewcommand{\TrbEoneVaNatDsSd}{4.3}%
\providecommand{\TrbEoneVaNatRc}{}\renewcommand{\TrbEoneVaNatRc}{84.5}%
\providecommand{\TrbEoneVaNatCrr}{}\renewcommand{\TrbEoneVaNatCrr}{0.27}%
\providecommand{\TrbEoneVaCropBDs}{}\renewcommand{\TrbEoneVaCropBDs}{53.0}%
\providecommand{\TrbEoneVaCropBDsSd}{}\renewcommand{\TrbEoneVaCropBDsSd}{10.0}%
\providecommand{\TrbEoneVaCropBRc}{}\renewcommand{\TrbEoneVaCropBRc}{80.6}%
\providecommand{\TrbEoneVaCropBCrr}{}\renewcommand{\TrbEoneVaCropBCrr}{0.40}%
\providecommand{\TrbEoneVaCropCDs}{}\renewcommand{\TrbEoneVaCropCDs}{52.8}%
\providecommand{\TrbEoneVaCropCDsSd}{}\renewcommand{\TrbEoneVaCropCDsSd}{7.8}%
\providecommand{\TrbEoneVaCropCRc}{}\renewcommand{\TrbEoneVaCropCRc}{80.5}%
\providecommand{\TrbEoneVaCropCCrr}{}\renewcommand{\TrbEoneVaCropCCrr}{0.47}%
\providecommand{\TrbEoneVaCropCCpk}{}\renewcommand{\TrbEoneVaCropCCpk}{5.85}%
\providecommand{\TrbEoneVaDownBDs}{}\renewcommand{\TrbEoneVaDownBDs}{58.2}%
\providecommand{\TrbEoneVaDownBDsSd}{}\renewcommand{\TrbEoneVaDownBDsSd}{0.9}%
\providecommand{\TrbEoneVaDownBRc}{}\renewcommand{\TrbEoneVaDownBRc}{84.5}%
\providecommand{\TrbEoneVaDownBCrr}{}\renewcommand{\TrbEoneVaDownBCrr}{0.40}%
\providecommand{\TrbEoneVaDownCDs}{}\renewcommand{\TrbEoneVaDownCDs}{70.3}%
\providecommand{\TrbEoneVaDownCDsSd}{}\renewcommand{\TrbEoneVaDownCDsSd}{0.3}%
\providecommand{\TrbEoneVaDownCRc}{}\renewcommand{\TrbEoneVaDownCRc}{84.5}%
\providecommand{\TrbEoneVaDownCCrr}{}\renewcommand{\TrbEoneVaDownCCrr}{0.20}%
\providecommand{\TrbEoneVaDownCCpk}{}\renewcommand{\TrbEoneVaDownCCpk}{2.05}%
\providecommand{\TrbEoneUnDsSdMax}{}\renewcommand{\TrbEoneUnDsSdMax}{10.8}%
\providecommand{\TrbEoneRho}{}\renewcommand{\TrbEoneRho}{0.56}%
\providecommand{\TrbEoneRhoLo}{}\renewcommand{\TrbEoneRhoLo}{\ensuremath{-}0.03}%
\providecommand{\TrbEoneRhoHi}{}\renewcommand{\TrbEoneRhoHi}{0.69}%
\providecommand{\TrbEoneNPts}{}\renewcommand{\TrbEoneNPts}{20}%
\providecommand{\TrbEoneResolved}{}\renewcommand{\TrbEoneResolved}{three}%
\providecommand{\TrbEoneNPairs}{}\renewcommand{\TrbEoneNPairs}{ten}%
\providecommand{\TrbEoneCrrRho}{}\renewcommand{\TrbEoneCrrRho}{0.62}%
\providecommand{\TrbEoneCrrRhoLo}{}\renewcommand{\TrbEoneCrrRhoLo}{0.28}%
\providecommand{\TrbEoneCrrRhoHi}{}\renewcommand{\TrbEoneCrrRhoHi}{0.66}%
\providecommand{\TrbEtwoUnRatioMin}{}\renewcommand{\TrbEtwoUnRatioMin}{2.76}%
\providecommand{\TrbEtwoUnRatioMax}{}\renewcommand{\TrbEtwoUnRatioMax}{6.16}%
\providecommand{\TrbEtwoVaRatioMin}{}\renewcommand{\TrbEtwoVaRatioMin}{1.91}%
\providecommand{\TrbEtwoVaRatioMax}{}\renewcommand{\TrbEtwoVaRatioMax}{4.24}%
\providecommand{\TrbEtwoAllRatioMin}{}\renewcommand{\TrbEtwoAllRatioMin}{1.91}%
\providecommand{\TrbEtwoAllRatioMax}{}\renewcommand{\TrbEtwoAllRatioMax}{8.50}%
\providecommand{\TrbNovRecKm}{}\renewcommand{\TrbNovRecKm}{12.5}%
\providecommand{\TrbNovRecMin}{}\renewcommand{\TrbNovRecMin}{38}%
\providecommand{\TrbNovVmax}{}\renewcommand{\TrbNovVmax}{4.0}%
\providecommand{\TrbNovThold}{}\renewcommand{\TrbNovThold}{1.5}%
\providecommand{\TrbNovHorizon}{}\renewcommand{\TrbNovHorizon}{4.0}%
\providecommand{\TrbEsixWzOk}{}\renewcommand{\TrbEsixWzOk}{4}%
\providecommand{\TrbEsixWzN}{}\renewcommand{\TrbEsixWzN}{5}%
\providecommand{\TrbEsixWzAge}{}\renewcommand{\TrbEsixWzAge}{1.48}%
\providecommand{\TrbEsixWzRate}{}\renewcommand{\TrbEsixWzRate}{50}%
\providecommand{\TrbEsixZeroOk}{}\renewcommand{\TrbEsixZeroOk}{5}%
\providecommand{\TrbEsixZeroN}{}\renewcommand{\TrbEsixZeroN}{5}%
\providecommand{\TrbEsixZeroAge}{}\renewcommand{\TrbEsixZeroAge}{0.80}%
\providecommand{\TrbEsixZeroRate}{}\renewcommand{\TrbEsixZeroRate}{25}%
\providecommand{\TrbEsixHalfOk}{}\renewcommand{\TrbEsixHalfOk}{3}%
\providecommand{\TrbEsixHalfN}{}\renewcommand{\TrbEsixHalfN}{3}%
\providecommand{\TrbEsixHalfAge}{}\renewcommand{\TrbEsixHalfAge}{1.59}%
\providecommand{\TrbEsixHalfRate}{}\renewcommand{\TrbEsixHalfRate}{27}%
\providecommand{\TrbEsixOneOk}{}\renewcommand{\TrbEsixOneOk}{3}%
\providecommand{\TrbEsixOneN}{}\renewcommand{\TrbEsixOneN}{3}%
\providecommand{\TrbEsixOneAge}{}\renewcommand{\TrbEsixOneAge}{2.34}%
\providecommand{\TrbEsixOneGap}{}\renewcommand{\TrbEsixOneGap}{1.09}%
\providecommand{\TrbEsixOneRate}{}\renewcommand{\TrbEsixOneRate}{40}%
\providecommand{\TrbEsixTwoOk}{}\renewcommand{\TrbEsixTwoOk}{0}%
\providecommand{\TrbEsixTwoN}{}\renewcommand{\TrbEsixTwoN}{2}%
\providecommand{\TrbEsixTwoAge}{}\renewcommand{\TrbEsixTwoAge}{3.88}%
\providecommand{\TrbEsixTwoHold}{}\renewcommand{\TrbEsixTwoHold}{44}%
\providecommand{\TrbEsixTwoRate}{}\renewcommand{\TrbEsixTwoRate}{80}%
\providecommand{\TrbEsixTwoSteerMax}{}\renewcommand{\TrbEsixTwoSteerMax}{220}%
\providecommand{\TrbEsixFtOk}{}\renewcommand{\TrbEsixFtOk}{0}%
\providecommand{\TrbEsixFtN}{}\renewcommand{\TrbEsixFtN}{2}%
\providecommand{\TrbEsixFtAge}{}\renewcommand{\TrbEsixFtAge}{0.83}%
\providecommand{\TrbEsixFtRate}{}\renewcommand{\TrbEsixFtRate}{22}%
\providecommand{\TrbEsixSbOk}{}\renewcommand{\TrbEsixSbOk}{0}%
\providecommand{\TrbEsixSbN}{}\renewcommand{\TrbEsixSbN}{2}%
\providecommand{\TrbEsixSbAge}{}\renewcommand{\TrbEsixSbAge}{0.82}%
\providecommand{\TrbEsixSbRate}{}\renewcommand{\TrbEsixSbRate}{26}%
\providecommand{\TrbEsixRouteLen}{}\renewcommand{\TrbEsixRouteLen}{143}%
\providecommand{\TrbEsixNTrips}{}\renewcommand{\TrbEsixNTrips}{22}%
\providecommand{\TrbEsixDistKm}{}\renewcommand{\TrbEsixDistKm}{2.7}%
\providecommand{\TrbEsixMin}{}\renewcommand{\TrbEsixMin}{13}%
\providecommand{\TrbEsixNInt}{}\renewcommand{\TrbEsixNInt}{seven}%
\providecommand{\TrbEsixWzIv}{}\renewcommand{\TrbEsixWzIv}{133.9}%
\providecommand{\TrbEsixTwoIvA}{}\renewcommand{\TrbEsixTwoIvA}{116.4}%
\providecommand{\TrbEsixTwoIvB}{}\renewcommand{\TrbEsixTwoIvB}{59.3}%
\providecommand{\TrbEsixFtIvA}{}\renewcommand{\TrbEsixFtIvA}{70.8}%
\providecommand{\TrbEsixFtIvB}{}\renewcommand{\TrbEsixFtIvB}{73.2}%
\providecommand{\TrbEsixSbIvA}{}\renewcommand{\TrbEsixSbIvA}{46.2}%
\providecommand{\TrbEsixSbIvB}{}\renewcommand{\TrbEsixSbIvB}{48.0}%
\providecommand{\TrbEsixGateSb}{}\renewcommand{\TrbEsixGateSb}{0.77}%
\providecommand{\TrbEsixGateTv}{}\renewcommand{\TrbEsixGateTv}{0.73}%
\providecommand{\TrbEsixGatePninetySb}{}\renewcommand{\TrbEsixGatePninetySb}{1.27}%
\providecommand{\TrbEsixGatePninetyTv}{}\renewcommand{\TrbEsixGatePninetyTv}{2.55}%
\providecommand{\TrbEsixZyDvaOk}{}\renewcommand{\TrbEsixZyDvaOk}{0}%
\providecommand{\TrbEsixZyDvaN}{}\renewcommand{\TrbEsixZyDvaN}{5}%
\providecommand{\TrbEsixZyRcOk}{}\renewcommand{\TrbEsixZyRcOk}{0}%
\providecommand{\TrbEsixZyRcN}{}\renewcommand{\TrbEsixZyRcN}{5}%
\providecommand{\TrbEsixZyLwamOk}{}\renewcommand{\TrbEsixZyLwamOk}{5}%
\providecommand{\TrbEsixZyLwamN}{}\renewcommand{\TrbEsixZyLwamN}{5}%
\providecommand{\TrbEsixZyDvaIvMin}{}\renewcommand{\TrbEsixZyDvaIvMin}{19.5}%
\providecommand{\TrbEsixZyDvaIvMax}{}\renewcommand{\TrbEsixZyDvaIvMax}{124.0}%
\providecommand{\TrbEsixZyRcIvMin}{}\renewcommand{\TrbEsixZyRcIvMin}{41.5}%
\providecommand{\TrbEsixZyRcIvMax}{}\renewcommand{\TrbEsixZyRcIvMax}{82.0}%
\newcommand{\trbnew}[1]{#1}
\newcommand{\trbnewon}{}
\newcommand{\trbold}[1]{#1}

\title{Triage Before Fine-Tuning:\\
Feasibility-Gated Selection and Monitored Adaptation of Embodied Foundation Models for a Fixed Physical Platform}

\author{Zilin Huang, Zhengyang Wan, Zihao Sheng, Boyue Wang, Ruolin Yang, Yuhao Luo, Sikai Chen\thanks{Corresponding author: \texttt{sikai.chen@wisc.edu}}\\
University of Wisconsin--Madison\\
Project page: \url{https://zilin-huang.github.io/Triage/}}

\begin{document}
\maketitle
\lhead{Preprint.}

\begin{abstract}
\rv{Embodied foundation models are increasingly deployed by selection: a practitioner picks a public checkpoint by benchmark rank and fine-tunes it on local data. The platform is fixed before the model is chosen: its sensors, onboard compute and operating site are set by the task, and at the start there is little or no local data. We ask what such a checkpoint loses on a fixed physical platform and which part of the loss local data can buy back. Our answer is to split the gap by \emph{what can repair it}. A sensing gap, information the sensors do not capture, cannot be closed with their data; a latency gap, output arriving after the control loop needs it, belongs to the loop, not the weights; only a distribution gap is fine-tuning's to repair. We turn this into \textsc{Triage}, a procedure for \emph{feasibility-gated selection} and \emph{monitored adaptation} in five stages: gates on the sensor contract and on \trbnew{end-to-end} latency that no score can override, \trbnew{probes on ten minutes of local logs, a ranking by target-relevant score among the candidates that pass}, and fine-tuning that is monitored \trbnew{and checked in closed loop before release}. On driving benchmarks, a deadline below a leading closed-loop model's latency makes it collide on 80 to 90\% of routes, and losing field of view moves plans 2.2 to 5.7 times more than losing resolution at equal pixels (1.6 to 2.4 in manipulation) \trbnew{and costs more driving score in closed loop}. \trbnew{\trbold{Local frames close 75\% of a stability gap but raise perturbation sensitivity} at small budgets, fine-tuning removes more of a moderate crop's cost than of a strong crop's,} and picking by score violates a gate in two of four deployment profiles. On a full-scale vehicle, the factory camera failed the sensing gate and an added camera repaired it. \trbnew{The latency gate, applied to the end-to-end latency measured on board, removed the two highest-scoring candidates, neither of which completed a closed-loop trip. The selected model passed closed-loop acceptance in \TrbEsixWzOk{} of \TrbEsixWzN{} trips, and the closed-loop check withheld a fine-tuned model that had improved offline.} \textsc{Triage} lets organizations that cannot train foundation models from scratch find out, before buying data or hardware, whether a public checkpoint can run on the robot or vehicle they own, and what to fix if not.}
\end{abstract}

\section{Introduction}
\label{sec:intro}

Public checkpoints of embodied foundation models, released by research
groups and industry teams, now span end-to-end driving systems
\citep{chitta2022transfuser,liao2025diffusiondrive}, vision-language-action
policies \citep{renz2025simlingo,zhou2025autovla,recogdrive2026,kim2024openvla},
video world models \citep{driveva2026} and generalist robot policies
\citep{oxe2024,octo2024,black2024pi0}, and public benchmarks built on
recorded datasets \citep{caesar2021nuplan,dauner2024navsim,walke2023bridge}
and simulated environments
\citep{dosovitskiy2017carla,jia2024bench,li2024simpler} evaluate and
compare them. It is increasingly practical to begin a new deployment with an existing model: select a pretrained checkpoint and adapt it.

A benchmark evaluates candidates under its own observation, environment,
and execution conditions, whereas a deployed policy acts through a
particular platform's sensors, onboard compute, and controller, and its
actions change the state from which its next observations come. \rv{The
order of these decisions matters. A deployer, such as a campus shuttle
operator, a mining or construction fleet, or a hospital delivery robot,
chooses the platform for the task: payload, ruggedness, cost, and sensing
are fixed long before any model is considered and are not changed merely
to suit a checkpoint. Such a deployer does not train a foundation model
from scratch and starts with little or no local data, since collecting it
on the target platform is the expensive step the model is meant to make
worthwhile.} Benchmark scores are thus evidence of capability, not of
compatibility with the platform. In our experiments a leading closed-loop model of a standard driving benchmark (SimLingo; \citealp{renz2025simlingo}) consumes a single
$110^\circ$ camera that neither the $64^\circ$ front camera of a
nuPlan-style monocular camera setup nor its surround ring reproduces as
one view, and under an inference deadline shorter than its own latency it collides on 80 to 90\% of routes; the
failure belongs to the deployment loop, not to the model. Rank alone is
therefore not a sufficient criterion; the question is which model is a
viable starting point for the platform and what must change to make it
work: the model, the sensing hardware, the execution architecture, or the
training data.

\begin{figure}[t]
\centering
\includegraphics[width=\linewidth]{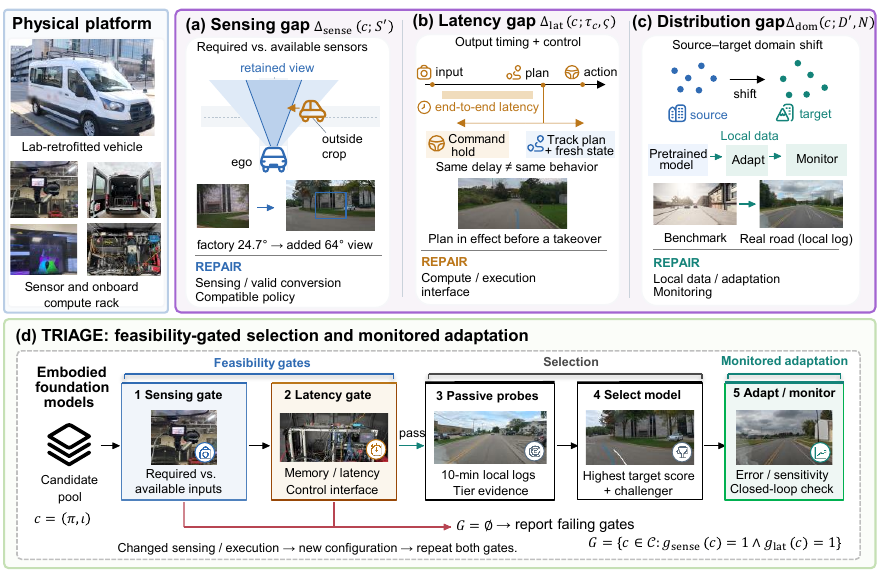}
\caption{\textbf{The repairability principle.} (a)~Sensing gap: no local data closes it. (b)~Latency gap: damage depends on how stale outputs are consumed. (c)~Distribution gap: local data repairs it. (d)~\textsc{Triage}: gates (Stages~1--2), selection (3--4), monitored adaptation (5).}
\label{fig:hero}
\end{figure}

Fine-tuning is the natural response to such differences, \rv{but the
first question is not how to fine-tune; it is which part of the gap
between a public checkpoint and the platform fine-tuning can reach at
all}: more data cannot recover
information the observations do not preserve, and lower open-loop
error does not resolve delayed execution. \textbf{Our insight} is
that deployment gaps should therefore be diagnosed by
\emph{repairability}, by whether their dominant cause is missing
information, late outputs, or distribution shift, since these call for
interventions that are not interchangeable (\cref{fig:hero}a--c). A \emph{sensing gap} arises when task-relevant distinctions cannot
be recovered from the deployed observations \trbnew{(Proposition~1, \cref{app:infogap})};
the repair is a different sensing configuration or a model whose
requirements match the platform. A \emph{latency gap} arises when model
outputs arrive too late for the control loop; its consequences depend
on whether the controller replays stale commands or tracks a delayed
plan with fresh state
\citep{katsikopoulos2003delay,ramstedt2019realtime}, and the repair
is faster computation or a different output-consumption architecture,
not more data. A \emph{distribution gap} arises from differences in
appearance, scene structure, or behavior, the mechanism target-domain
adaptation is designed to address \citep{bendavid2010theory}.

Building on this diagnosis, \textbf{we introduce \textsc{Triage}}
(\cref{fig:hero}d), a staged diagnostic framework for the
feasibility-gated selection and monitored adaptation of embodied
foundation models. It checks each candidate's sensing requirements
against the platform and measures its \trbnew{end-to-end latency, from camera
frame to plan,} on the deployment hardware against the \trbnew{deadline of the}
intended control interface. Candidates
that violate either requirement are excluded, and a higher score cannot
bring them back: that takes a change to sensing, compute, model
implementation, or the execution interface. \trbnew{Among the remaining
candidates the target-relevant benchmark score selects, while passive
\trbold{probes on short local logs} confirm each candidate's tier and check that
it responds to its images as it should}; adaptation is evaluated on tracking
performance, perturbation sensitivity \citep{kumar2022finetuning}, and
reliance on the image. When no candidate satisfies the
requirements, \textsc{Triage} reports the limiting constraints rather
than forcing a choice. \rv{\trbnew{Beyond a benchmark ranking, \textsc{Triage}
contributes three things: gates with operational definitions, and
evidence that selecting by score or by closed-loop rank violates them; a split of the gap by
what repairs it; and monitoring of adaptation for the robustness it can
cost. Within the feasible set it ranks as the benchmark does, since no
cheaper evidence we tested ranked better.}} Its purpose is to determine when benchmark
evidence applies to a deployment and where effort should go, not to
produce another universal ranking.

\textbf{Contributions.}
(i)~We formulate the deployment of pretrained embodied foundation models
on a fixed platform as a problem of \emph{repairability} rather than
benchmark ranking, distinguishing sensing, latency, and distribution gaps
by their causes and interventions, and turn it into \textsc{Triage}, a
staged framework that returns either a viable starting point or the
constraint preventing deployment.
(ii)~We evaluate \textsc{Triage} on 1{,}085 closed-loop runs in
CARLA/Bench2Drive, thirteen driving configurations on four-city nuPlan
logs (eleven scored under the official NAVSIM evaluator), and six
open-weight manipulation models on Bridge~V2. Restricting field of view
costs $2.2$--$5.7\times$ more than reducing resolution at a matched pixel
budget in driving and $1.6$--$2.4\times$ in manipulation, and whether
fine-tuning compensates a sensory restriction depends on the model\trbnew{, on whether its vision encoder adapts and on how much is cropped}.
(iii)~We run \trbnew{all five stages of} \textsc{Triage} on a full-scale vehicle
(\cref{sec:casestudy,app:onvehicle}): its factory camera fails the sensing gate\trbnew{; the latency gate, applied to the end-to-end latency measured on board, removes the two highest-scoring candidates, neither of which completed a closed-loop trip; Stage~4 returns a model that passes closed-loop acceptance; and the closed-loop check withholds a locally fine-tuned model that had improved offline (\cref{tab:e6}).}

\section{Related work}
\label{sec:related}

\textbf{Selecting and adapting checkpoints.}
Transferability estimators such as LEEP \citep{nguyen2020leep} and
LogME \citep{you2021logme} estimate transfer performance from labeled
target predictions or features, and offline policy selection
\citep{konyushkova2021active}, simulation-based evaluation
\citep{li2024simpler}, and runtime monitoring \citep{sentinel2024}
estimate performance or detect failures; none is designed to check
sensing or control compatibility with the platform before deployment.
Driving benchmarks \citep{jia2024bench,dauner2024navsim} address
limitations of open-loop evaluation \citep{li2024egostatus}, and
counterfactual image edits show that leaderboard scores do not reveal
whether end-to-end and vision-language-action driving policies respond
to safety-critical scene content under domain shift
\citep{yang2026beyond}; we ask the prior question of whether a platform
can run the checkpoint at all. Surgical fine-tuning
\citep{lee2023surgical} selects which weights to update for a given
shift; we extend intervention selection to sensing, compute, and
execution.

\textbf{Diagnosing the deployment gap.}
Sim-to-real surveys organize the reality gap by system component
\citep{aljalbout2025reality}; our taxonomy cuts across these components,
classifying mismatches by the interventions they need, and rests on
Blackwell's comparison of experiments \citep{blackwell1953}, delayed
decision processes \citep{katsikopoulos2003delay,ramstedt2019realtime},
and domain-adaptation theory \citep{bendavid2010theory}. Delay-aware
learning \citep{chen2020delayaware} and real-time chunking
\citep{black2025rtc} address latency through state augmentation or
asynchronous execution, interventions beyond ordinary target-domain
fine-tuning, and cross-embodiment studies \citep{oxe2024,gr1g12025} show
that data and interface design can accommodate some differences across
robots. We therefore diagnose individual gap components rather than
treating an entire embodiment as transferable or irreparable.

\section{The \textsc{Triage} framework}
\label{sec:framework}

\subsection{Problem formulation}
\label{sec:problem}
A deployer holds a platform whose sensors $S'$, compute $T'$ and
operating domain $D'$ are fixed before any model is chosen, and a
budget of $N$ local frames for adaptation; the candidates are a pool
$\mathcal{C}$ of public checkpoints, each trained under its own sensors,
platform and data $(S_i,T_i,D_i)$. What is deployed is a checkpoint
$\pol$ with the interface $\iota$ it runs through (cameras consumed, pre-
and post-processing, controller, hardware), a \emph{deployment
configuration} $c=(\pol,\iota)$, and what matters is its risk on $D'$,
$R(c)=\E_{\omega\sim P^{c}_{D'}}[L(\omega)]$, the expected task loss
(collisions, tracking error) over rollouts whose law depends on policy
and interface alike. Benchmark scores and local probes are evidence
about $R(c)$, not measurements of it. \emph{Post-training} is any
adaptation on observations from the target sensors and their labels;
changing architecture or interface creates a new configuration.

The framework rests on one structural claim: the gap between $R(c)$
and the checkpoint's risk on its own platform has three sources, and
the frame budget acts on only one,
\begin{equation}
\Delta(c)=\Bigl(
\underbrace{\Delta_{\mathrm{sense}}(c;S')}_{\substack{\text{sensing term:}\\ \text{invariant to }N}},\;
\underbrace{\Delta_{\mathrm{lat}}(c;\tau_c,\varsigma)}_{\substack{\text{latency term:}\\ \text{invariant to }N}},\;
\underbrace{\Delta_{\mathrm{dom}}(c;D',N)}_{\substack{\text{distribution term:}\\ \text{reduced by }N}}
\Bigr),
\label{eq:decomp}
\end{equation}
with $\tau_c$ the \trbnew{end-to-end latency, taken as fixed in the analysis and measured for the gate by the 95th percentile $\hat\tau_c$ of the end-to-end latency (\cref{sec:gates})} and $\varsigma$ the way the control
loop consumes stale outputs. Each term has a distinct repair (\cref{app:theory}), and
\textsc{Triage} determines which term a candidate's gap occupies before
any score is compared.

\subsection{Three gap classes}
\label{sec:gaps}
\label{sec:principle}
\textbf{Sensing gaps.} A sensing gap
$\Delta_{\mathrm{sense}}(c;S')$ exists when the decision depends on
observations the target sensors $S'$ do not carry, as when a
checkpoint trained with three cameras, LiDAR or six surround cameras
must decide from one. \trbnew{Proposition~1 (\cref{app:infogap}) bounds
what any policy can do with a garbled channel: no policy computed from
it, fine-tuned or not, reaches a lower risk than the best policy on the
original channel, and the gap is strict exactly when no optimal decision
can do without what was removed. Its proof uses only the garbling
direction of Blackwell's comparison of experiments \citep{blackwell1953},
extended to sequential decisions \citep{lecam1964}, which holds for every
decision rule and hence for imitation and reinforcement learning alike.
It orders two channels only when one is a garbling of the other at the
same delay, so a slower, richer sensor and a faster, coarser one are not
ordered; \textsc{Triage} therefore treats sensor, model and latency as
one configuration.} What matters is whether the missing
information is \emph{inferable} from what remains: monocular depth is
\citep{kwon2025memdistill}, cross-traffic outside the frustum is not.
The deficit concentrates where the decision needs the missing view: on turning frames, the position two seconds ahead lies outside a moderately cropped image $79.7\%$ of the time (\cref{fig:fovgeometry}).
The repair is a sensor or a model the platform's sensors can feed; how
much of a moderate deficit is non-inferable is empirical.

\textbf{Latency gaps.} A latency gap $\Delta_{\mathrm{lat}}(c;\tau_c,\varsigma)$
exists when the policy's output arrives later than the control loop
consumes it. With a fixed latency $\tau_c$, control period $\delta$ and
$k=\lceil\tau_c/\delta\rceil$ pending outputs, the closed loop is a delayed decision process \citep{katsikopoulos2003delay,ramstedt2019realtime} that combines the newest output, $k$ control steps old, with current proprioception $p_t$
through a consumption rule $\varsigma$. Fine-tuning changes neither $k$ nor $\varsigma$: it can learn to
extrapolate what is predictable over the delay, but the policy still
acts on $k$-step-old observations, so the least attainable risk is
unchanged (Prop.~2, \cref{app:timegap}). The consumption rule decides
the failure mode: holding the last command discards $p_t$ and accumulates
open-loop error, whereas tracking the last plan with fresh state loses,
if the tracked loop is contracting, at most a constant times the plan's
drift during the delay. The repair is faster compute or an architecture
that tolerates stale plans.

\textbf{Distribution gaps.} A distribution gap
$\Delta_{\mathrm{dom}}(c;D',N)$ exists when the appearance, layout or
behavior the checkpoint was trained on differs from the target domain
$D'$ with sensors and control loop unchanged, as when a checkpoint
trained in a simulator or another city meets unfamiliar streets:
covariate shift, or concept shift where local driving behavior differs.
It is the only class local data can reach and repair: in the source-and-target bound of \citet{bendavid2010theory}
(\cref{app:distgap}), a hypothesis trained on $m$ source and $N$ target
samples pays the divergence $d_{\mathcal{H}\Delta\mathcal{H}}(D,D')$
with a weight the optimal mixture drives to zero as $N$ grows, at the price of an estimation term.

\subsection{Gates before ranking}
\label{sec:gates}
Sensing and latency compatibility are checked by operational \emph{gates}, not by estimating the theoretical risk gaps of \cref{eq:decomp}. A score cannot override a failed requirement; selection is restricted to configurations passing both checks and ranks them by a proxy $\widehat{\Delta}_{\mathrm{dom}}$\trbnew{, the target-relevant benchmark score, with probe readings confirming each candidate's tier} (\cref{sec:procedure}):
\begin{equation}
G=\{\,c\in\mathcal{C}:\ g_{\mathrm{sense}}(c)=1
\ \wedge\ g_{\mathrm{lat}}(c)=1\,\},
\qquad
\hat{c}=\arg\min_{c\in G}\ \widehat{\Delta}_{\mathrm{dom}}(c;D',N).
\label{eq:lexico-main}
\end{equation}
$g_{\mathrm{sense}}=1$ means the configured inputs meet the sensor contract; $g_{\mathrm{lat}}=1$ means the model fits the hardware and its \trbnew{measured latency $\hat\tau_c$} meets a preset, interface-specific deadline $d'$: the control period $\delta$ under command hold, and under plan tracking a separate replanning deadline whose configuration must be validated in closed loop, not inferred from stale-image tolerance. \trbnew{$\hat\tau_c$ is the 95th percentile of the end-to-end latency, from the acquisition of a plan's newest frame to the finished plan, \trbold{measured on the deployment hardware in its operating state;} inference time alone, as timed in \cref{sec:exp-time}, is a lower bound on this end-to-end latency that can exclude a candidate but not admit one.} Remark~1 (\cref{app:lexico}) gives a sufficient condition for gating first under a common risk measure. An empty $G$ reports the failing gates, and changed sensing or execution interfaces re-enter both gates.

\subsection{Workflow}
\label{sec:procedure}
\cref{fig:hero}d shows the five stages; \cref{alg:triage} (\cref{app:procedure}) states them formally with their costs.
\emph{Stage~1} checks each sensor contract against $S'$ on paper; a structural
deficit excludes, a convertible one re-enters as its own
configuration. \emph{Stage~2} checks memory fit and measures \trbnew{$\hat\tau_c$} on the deployment hardware in its operating state; the candidates that pass form $G$.
\emph{Stage~3} runs each remaining candidate offline on ten minutes of local logs and records its \emph{probe readings}: temporal self-consistency and paired responses to
gamma, JPEG, field-of-view and resolution degradations, with two sanity checks (planned turns agree in sign with the driver;
blind anchors, the image-free baselines of \cref{tab:pool} (the rest are \emph{sighted}), respond exactly zero). \emph{Stage~4} \trbnew{is a gate-restricted benchmark ranking: the candidate in $G$ with the highest \trbold{target-relevant score is selected} and the \trbold{runner-up kept as a challenger.} The probe readings confirm each candidate's architectural tier \trbold{(traditional stacks; VLAs and world models)} or veto it; they do not rank within a tier (RQ7).} \emph{Stage~5} spends $N$ on fine-tuning the selection while watching tracking error, perturbation sensitivity and the plan's reliance on the image, and tests \trbnew{the adapted model and} any converted interface in closed loop \trbnew{before release}.

\section{Experiments}
\label{sec:experiments}

The subsections cover the two gates, the distribution gap and its repair, selection, and transfer to manipulation; full results are in \cref{app:extended} and metric definitions in \cref{app:instruments}.

\label{sec:exp-setup}
\begin{wraptable}{r}{0.61\linewidth}
\vspace{-1.0\baselineskip}
\centering\scriptsize
\caption{\textbf{The driving pool.} Official PDMS (per-city best in bold), sensors as consumed, and p50 latency on an idle RTX A6000.}
\label{tab:pool}
\setlength{\tabcolsep}{1.6pt}
\begin{tabular}{@{}lllrrrrrr@{}}
\toprule
Configuration & Venue & Sensors & ms & SIN & BOS & LVG & PIT & Glob.\\
\midrule
\multicolumn{9}{@{}l}{\emph{Blind anchors}}\\
Const.-velocity & anchor & none & 0.01 & 17.6 & 20.6 & 21.1 & 22.3 & 20.7\\
EgoStatusMLP & NeurIPS'24 & state & 0.12 & 52.9 & 63.8 & 78.9 & 55.8 & 65.5\\
\midrule
\multicolumn{9}{@{}l}{\emph{Traditional stacks}}\\
TCP & NeurIPS'22 & 3 cam, stitched & 63 & -- & -- & -- & -- & --\\
TransFuser & TPAMI'23 & 3 cam + LiDAR & 69 & 76.8 & 80.8 & 91.0 & 79.9 & 83.4\\
LTF & TPAMI'23 & 3 cam & 69 & 74.8 & 82.6 & 91.2 & 82.1 & 84.1\\
DiffusionDrive & CVPR'25 & 3 cam + LiDAR & 85 & 82.0 & 87.0 & 93.0 & 86.6 & 88.1\\
\midrule
\multicolumn{9}{@{}l}{\emph{VLAs}}\\
ReCogDrive-2B & ICLR'26 & mono & 513 & 85.7 & 89.3 & \textbf{94.7} & 88.8 & 90.4\\
AutoVLA-3B & NeurIPS'25 & 3 cam & 1{,}249 & 82.5 & 90.1 & 92.4 & 88.7 & 89.4\\
Curious-VLA-3B & CVPR'26 & mono & 4{,}067 & 85.9 & 88.5 & 93.6 & 88.2 & 89.8\\
Alpamayo-1.5-10B & release'26 & flexible & 959 & -- & -- & -- & -- & --\\
\midrule
\multicolumn{9}{@{}l}{\emph{World models}}\\
DriveVLA-W0-8B & ICLR'26 & mono & 485 & 86.2 & 89.5 & 93.7 & 89.0 & 90.3\\
WAM-Flow-1.5B & CVPR'26 & mono & 781 & 85.5 & 90.1 & 93.7 & 88.8 & 90.3\\
DriveVA-5B & ECCV'26 & mono video & 947 & \textbf{87.6} & \textbf{91.1} & 94.0 & \textbf{89.6} & \textbf{91.2}\\
\bottomrule
\end{tabular}
\vspace{-1.2\baselineskip}
\end{wraptable}
\textbf{Setup.} Closed loop: CARLA/\allowbreak Bench2Drive
\citep{dosovitskiy2017carla,jia2024bench} at fixed versions,
1{,}085 accepted runs of TCP, SimLingo, UniAD and VAD, with ORION added, over six town--weather environments plus a deadline axis and the occlusion and frame-drop axes of Bench2Drive-Robust \citep{zhang2026b2drobust}. Open loop: the official NAVSIM evaluator at a fixed version \citep{dauner2024navsim} (PDMS, a 0 to 100 composite of safety, comfort and progress terms) over nuPlan logs \citep{caesar2021nuplan} from Singapore,
Boston, Las Vegas and Pittsburgh, 12{,}146 test tokens. The pool (\cref{tab:pool}) holds thirteen
driving configurations in four tiers; every configuration with a published
score entered after reproducing it under the same evaluator (one exception, ORION, in \cref{app:protocols}). A candidate's sensor contract is the set of sensors its inference code consumes (\cref{tab:pool}).

\subsection{The sensing gate}
\label{sec:exp-info}

\begin{figure}[t]
\centering
\includegraphics[width=\linewidth]{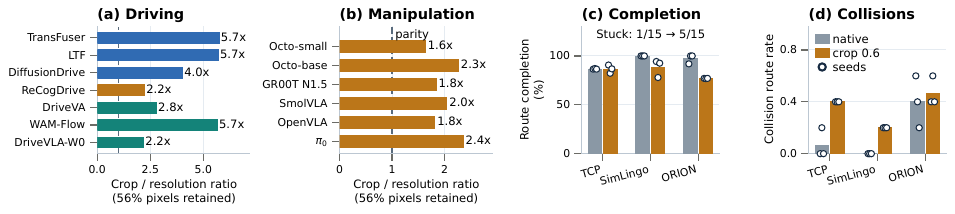}
\caption{\textbf{The sensing gate (RQ1, RQ3, RQ9).} (a,\,b)~At equal pixels a crop costs more than a resolution drop in every model. (c,\,d)~A moderate crop raises TCP's and SimLingo's collisions and costs ORION completion.}
\label{fig:info}\label{fig:fovratios}\label{fig:cropcl}
\end{figure}

\RQ{1}\textbf{: At equal pixel budgets, does structure loss damage
more than detail loss?} Open loop on nuPlan frames from four cities
and seven configurations, central crops keeping $90$, $75$ and $60\%$ of
each image dimension remove structure, and downscaling each dimension
to $75$, $50$ and $33\%$ removes detail, both read as mean plan displacement, the average distance between the plans on the degraded and original frames. Every sighted model displaces further as the view narrows; resolution curves are shallower, and blind
anchors sit at exactly zero on both axes (curves in \cref{app:ext-info}). At a matched $56\%$ of pixels kept
(\cref{fig:info}a), losing them at the border costs $2.2$--$5.7\times$
more than losing them everywhere, and six of seven
respond less to $1/9$ of native resolution than to a crop that
removes $19\%$ of pixels. The ratio does not track the sensor suite, though the largest absolute response is monocular (DriveVLA-W0, $1.2$\,m). Losing view costs several times more than losing sharpness, which is why a narrower view is gated and a
lower resolution treated as a format conversion.

\begin{figure}[t]
\centering
\includegraphics[width=\linewidth]{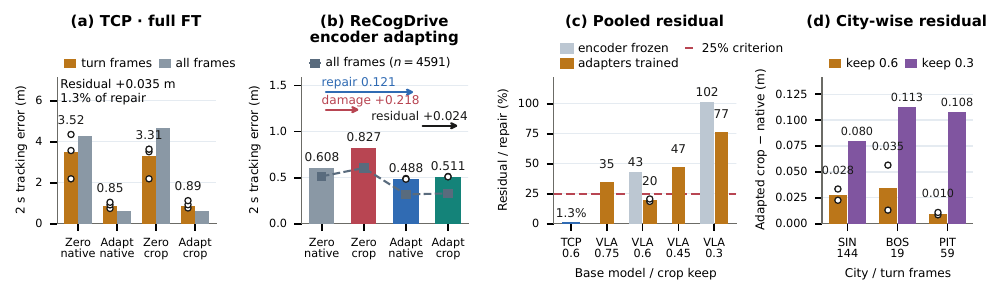}
\caption{\textbf{Fine-tuning under a moderate crop (RQ2).} Turn-frame tracking error, native/crop $\times$ zero-shot/adapted: (a)~TCP; (b)~ReCogDrive\trbnew{, encoder adapting (bars: two-seed mean; dots: seeds; dashed: all frames)}. (c)~Residual as a share of repair. (d)~ReCogDrive's residual by city\trbnew{, encoder adapting}.}
\label{fig:repair}\label{fig:repairinfo}\label{fig:infogatevla}
\end{figure}

\RQ{2}\textbf{: Can fine-tuning compensate a moderate sensory
deficit?} \trbnew{Yes, provided the vision encoder adapts; a strong crop leaves a residual that fine-tuning does not remove.} Open loop on nuPlan frames, we fine-tune TCP under a crop
keeping $60\%$ of each dimension and
evaluate under it against a matched-budget uncropped run, with turning frames reported separately; ReCogDrive repeats it with a recipe that leaves its vision encoder frozen (LoRA(LM)+head, \cref{app:ext-info})\trbnew{, and again with the encoder adapting (two seeds at this crop, one at each of three other crop sizes)}. We set the criterion that a sensing gap leaves at least $25\%$ of the
uncropped repair as residual. TCP repairs $73$--$76\%$ of its turn error with and without the crop and leaves $1.3\%$, so the criterion is not met
(\cref{fig:repair}a); ReCogDrive leaves $43\%$ and meets it
\trbnew{with its encoder frozen (\cref{fig:repair}c)}, repairing straight driving to within a
millimeter but not turns\trbnew{. With the encoder adapting, the residual falls to \TrbEfourShareSixA{} and \TrbEfourShareSixB\% of the repair in the two seeds (\cref{fig:repair}b,\,c), below the criterion. Both values exceed the seed spread, but each seed's 95\% interval includes zero: the frozen encoder, not the crop, accounts for most of the $43\%$}. With a crop keeping $30\%$ the residual is $101.8\%$ of the
same repair (\cref{fig:repair}c)\trbnew{, and \TrbEfourShareThree\% with the encoder adapting, its interval above zero}, and the compensated fraction (the
share of the crop's zero-shot cost that adaptation removes) falls
from $0.746$ to $0.642$ \trbnew{(from \TrbEfourCfSixB--\TrbEfourCfSixA{} to \TrbEfourCfThree{} with the encoder adapting)}: the larger the gap, the less local data buys
back. City-wise residuals are positive in all three cities with
turns (\cref{fig:repair}d)\trbnew{, with either recipe}. \trbnew{A moderate deficit therefore re-enters as its own configuration, to be adapted and then checked in closed loop (\cref{alg:triage}).}

\RQ{3}\textbf{: Does a moderate sensory deficit harm safety even
when capability metrics look intact?} Closed loop on Bench2Drive, two
probes run on released checkpoints: a random rectangle blacked out on
every camera at severities $50$ and $80$, and the crop keeping $60\%$ applied
zero-shot to TCP and SimLingo over $30$ runs. At severity $50$ the
forward-facing configurations' collision route rates rise by $+0.3$
while the six-camera configurations hold or fall; at $80$ every configuration
fails, so redundant cameras help only up to moderate severity. The crop
(\cref{fig:info}c,\,d) separates capability from safety: route
completion stays within seed noise, while
collision route rate rises beyond it
($0.067$ to $0.400$ for TCP, $0.000$ to $0.200$ for SimLingo). ORION is damaged in a different channel: completion falls
from $0.973$ to $0.770$ and stuck routes rise from $1/15$ to $5/15$.
RQ2 (fine-tuned, open loop) and RQ3 (zero-shot, closed loop) show that open-loop repair does not certify closed-loop safety.

\subsection{The latency gate}
\label{sec:exp-time}

\begin{figure}[t]
\centering
\includegraphics[width=\linewidth]{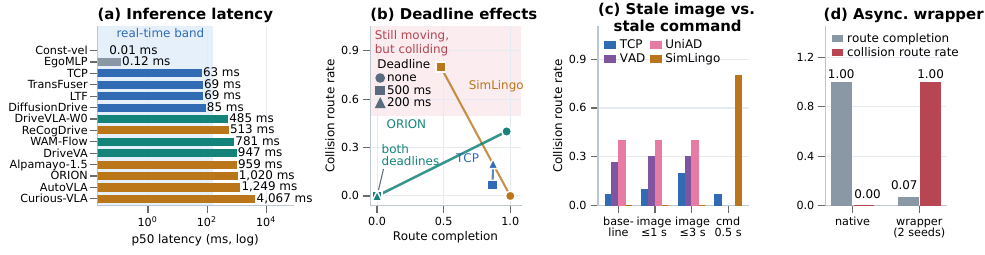}
\caption{\textbf{The latency gate (RQ4).} (a)~Latency spans five decades. (b)~At a $500$\,ms deadline, SimLingo collides and ORION stalls. (c)~Stale images are tolerated, stale commands not. (d)~The wrapper always collides.}
\label{fig:timegate}
\end{figure}

\RQ{4}\textbf{: Is a latency violation a safety event, what decides
its failure mode, and can wrappers or fine-tuning repair it?} We time every model per frame with no other process on the GPU, then enforce deadlines closed loop on Bench2Drive, holding the last command whenever a plan is late, over three towns,
five routes and two or three seeds. Latency spans five decades (\cref{fig:timegate}a): $63$--$85$\,ms for the traditional stacks of \cref{tab:pool},
$0.49$--$4.1$\,s for every VLA and world model. At a $500$\,ms deadline SimLingo still covers $48$ to $58\%$ of route length but collides on
$0.8$--$0.9$ of routes in every town (\cref{fig:timegate}b). Freezing the camera input in random bursts of up to 1 or 3\,s, while the model keeps replanning on fresh speed and position, keeps collision rates near baseline (\cref{fig:timegate}c): a stale image is tolerated when the loop acts on fresh state, a stale command is not. Of three
asynchronous-planning wrappers two failed before completing a route,
and the filtered-pose one completes $0.07$ of route length and
collides on every route (\cref{fig:timegate}d). Fine-tuning does not
move the clock: the training architecture's latency changed by
$+0.18\%$, and ReCogDrive's adapted model, its adapters mounted rather than
merged, is $13.3\%$ slower. ORION, whose $1{,}020$\,ms latency exceeds both deadlines tested, misses every scheduled call and never leaves the start; latency therefore belongs in a gate, and its repair lies in the consumer, not in fine-tuning or a wrapper.

\subsection{The distribution gap}
\label{sec:exp-dist}

\begin{figure}[t]
\centering
\includegraphics[width=\linewidth]{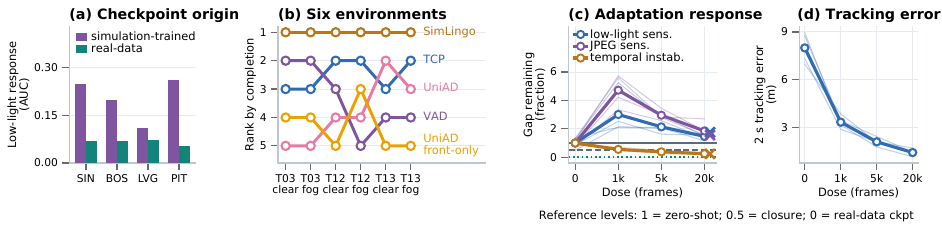}
\caption{\textbf{The distribution gap (RQ5, RQ6).} (a)~The simulation-trained checkpoint is $1.5$--$4.9\times$ more sensitive. (b)~Environments reorder the pool. (c)~Stability closes; sensitivity spikes at 1k. (d)~Tracking error falls.}
\label{fig:dist}\label{fig:repairdist}
\end{figure}

\RQ{5}\textbf{: How large and how structured is the gap that training data and environment create?} We contrast public checkpoints of the same TransFuser architecture (\cref{app:twins}), one trained in CARLA and one on real data, open loop under an identical probe code on nuPlan frames.
The simulation-trained checkpoint reacts $1.5$--$4.9\times$ more
strongly to darkened and compressed frames in all four cities
(\cref{fig:dist}a), and its plans disagree with their own previous
frame $8.6$--$10.3\times$ more (\cref{app:ext-dist}). Closed loop,
six environments reorder the pool: 13 of 15 environment pairs change
pairwise order among the native configurations
(\cref{fig:dist}b), with each configuration's spread across
environments dwarfing seed noise. A UniAD pair narrows the claim
to $1.13\times$; UniAD holds state across calls, and run-to-run noise exceeds its perturbation readings on the simulation-trained side.

\RQ{6}\textbf{: What does fine-tuning repair, and at what cost?} We behavior-clone the simulation-trained
TransFuser on target-city frames at 1k, 5k and 20k, measuring each gap against the real-data checkpoint, with a second seed at 20k and an LP-FT variant \citep{kumar2022finetuning}. Stability closes $75\%$ of the zero-shot gap by 20k (\cref{fig:dist}c), but after 1k frames sensitivity
to darkened and compressed inputs rises $2.4\times$ and $3.4\times$,
leaving a gap $3.0$ to $4.7\times$ its zero-shot size, so these gaps do not close; the spike appears in every city, and LP-FT
raises the 1k response 1.6 to 2.0 times. Tracking error still falls from
$8.0$ to $1.4$\,m (\cref{fig:dist}d). On ReCogDrive (LoRA(LM)+head) no spike appears (\cref{fig:dose}): the three responses fall to $0.36$, $0.33$
and $0.55$ of zero-shot by 20k, though the smallest dose is worse
than zero-shot on turns ($1.15$). Since falling sensitivity also fits
a model that attends less to the image, we measure the plan's reliance on the image as its response to a constant gray frame (\cref{app:instruments}): it falls below the range seen when only the training seed changes, at every dose, so part of the fall is exactly that.
Fine-tuning repairs its assigned class, but the spike depends on the recipe and the base model, so \cref{alg:triage}'s monitored step must watch the plan's reliance on the image beside sensitivity.

\begin{figure}[t]
\centering
\includegraphics[width=\linewidth]{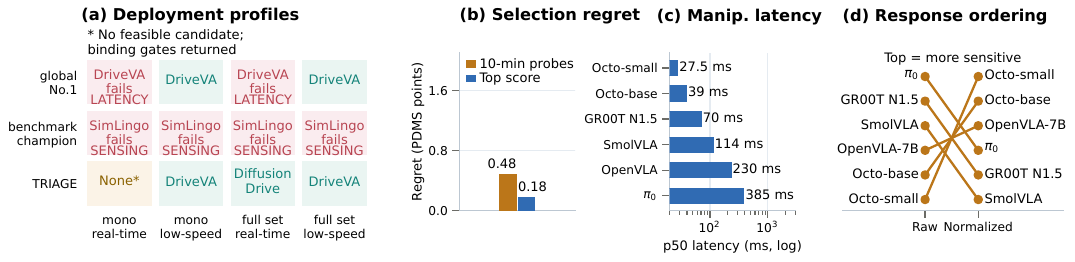}
\caption{\textbf{Selection (RQ7; c, d from RQ9).} (a)~Picks per profile (red: fails a gate; amber: empty set). (b)~Probes lose to the top score. (c)~Manip.\ latency. (d)~Raw vs.\ normalized ranks.}
\label{fig:select}\label{fig:deploymentprofiles}\label{fig:selection}\label{fig:portability}
\end{figure}

\subsection{Selection with the full framework}
\label{sec:exp-select}

\RQ{7}\textbf{: Does \textsc{Triage} select better than standard
practice, and where does cheap evidence stop?}
\begin{wraptable}{r}{0.49\linewidth}
\vspace{-4pt}
\centering\scriptsize
\caption{\textbf{Newly admitted candidates (RQ8).} Green: admitted, score reproduced; red: does not fit this platform.}
\label{tab:admissions}
\setlength{\tabcolsep}{3pt}
\begin{tabular}{@{}llll@{}}
\toprule
Candidate & Sensor contract & p50 (ms) & PDMS (pub.\ / ours)\\
\midrule
Alpamayo~2 Super & 6 cams, 70\,GiB & -- & \multicolumn{1}{>{\columncolor{failred}[\tabcolsep][0pt]}l@{}}{does not fit}\\
Alpamayo-1.5 & flexible & 959 & probe level only\\
WAM-Flow & mono & 781 & \multicolumn{1}{>{\columncolor{passgreen}[\tabcolsep][0pt]}l@{}}{90.3 / 90.32}\\
DriveVLA-W0 & mono & 485 & \multicolumn{1}{>{\columncolor{passgreen}[\tabcolsep][0pt]}l@{}}{90.2 / 90.30}\\
\bottomrule
\end{tabular}
\vspace{-6pt}
\end{wraptable}
Open loop under the
NAVSIM evaluator, across four deployment profiles (\cref{fig:select}a),
\trbnew{picking the top benchmark score without gates (\emph{score-first}) \trbold{violates the latency gate in \TrbScoreFirstViolProfiles/\TrbNProfiles{}} profiles and picking the closed-loop champion (\emph{closed-loop-first}) the
\trbold{sensing gate in \TrbChampionViolProfiles/\TrbNProfiles.} In the full-camera real-time profile the infeasible score-first pick outscores the best feasible candidate by \TrbInfeasibleAdvantage{} points, so the score cannot reveal the violation.} \textsc{Triage} agrees with the scoreboard
wherever all gates pass and, for monocular real-time deployment,
returns an empty set with the gates that fail. \trbnew{These profiles use the inference times of \cref{tab:pool}.} \cref{fig:select}b \trbnew{asks whether probes could replace the score}: nine configurations in four
cities give \trbnew{36 rows \trbold{with ten fixed features and
four predictor families,} and \TrbLocoSelections{} leave-one-city-out selections made without gates}; regret is the PDMS the pick gives up
against the best candidate in the held-out city. \trbnew{A predictor fitted to the probe readings \trbold{never picks below the top tier, but within it} loses to taking the top score (\TrbUngatedProbe{} against \TrbUngatedScore{} \trbold{points of mean regret}), \trbold{and adding latency as a feature worsens it to \TrbUngatedProbeLat.} With the gates applied (\cref{tab:trb-selection}), the gate-restricted benchmark ranking of Stage~4 gives up \TrbGatedScoreMin{} to \TrbGatedScoreMax{} points in the three profiles with feasible candidates, against \TrbRandomMin{} to \TrbRandomMax{} for a uniform pick, and no probe-based rule beats it in any profile.} Raw sensitivities are not
comparable across models (\cref{fig:engagement}a): normalizing each
by the model's response to a constant gray frame leaves no stateless
stack in its rank. Cheap probes place tiers but cannot separate models of the same generation; latency belongs in a gate, not a selection feature.

\RQ{8}\textbf{: Does the framework absorb new state-of-the-art
candidates?} Four NAVSIM candidates joined after the pool was assembled
(\cref{tab:admissions}; full results in \cref{app:ext-select}).
Alpamayo~2 Super does not fit this platform on paper: its documented $70$\,GiB footprint (for an $80$\,GB H100) and six cameras exceed a monocular $48$\,GB platform. Alpamayo-1.5 was admitted ($959$\,ms); its front-only ablation leaves plan self-consistency intact but lowers agreement with the driver's turn direction ($0.915$ against $0.954$), placing the model card's caution in turn correctness rather than plan stability. WAM-Flow and DriveVLA-W0 reproduced their published scores and RQ1's asymmetry.

\subsection{Transfer to manipulation}
\label{sec:exp-gen}

\RQ{9}\textbf{: Do the instruments and the sensory asymmetry
transfer to a second embodiment?} The same sensor-contract and latency checks and the same probes run unmodified on six manipulation models (Octo-small, Octo-base,
GR00T~N1.5, SmolVLA, OpenVLA-7B and $\pi_0$; sources in \cref{tab:manipcontracts})
in offline inference on recorded episodes of an official Bridge~V2 shard
\citep{walke2023bridge}. Latency spans 27.5\,ms (Octo-small) to 385\,ms ($\pi_0$), a $14\times$ spread (\cref{fig:select}c), narrower than the five decades of driving. The field-of-view asymmetry holds in all six: cropping the view moves the actions $1.6$--$2.4\times$ more than downscaling it at the same pixel budget (\cref{fig:info}b), with the constant-action anchor at exactly zero. The probe's limitation transfers too (\cref{fig:select}d): responses to losing the image
span $31.8\times$, and ranking by raw versus normalized response gives
a near-reversal; as in driving, the probe resolves several-fold differences but not small ones.

\section{Real-world field study on a laboratory vehicle}
\label{sec:casestudy}

\begin{wrapfigure}{r}{0.7\linewidth}
\vspace{-10pt}
\centering
\resizebox{\linewidth}{!}{\begin{tikzpicture}[x=1bp,y=1bp,every node/.style={inner sep=0,anchor=south west}]
\node at (0,197.693) {\includegraphics[trim=0 197.693bp 400.949bp 0,clip]{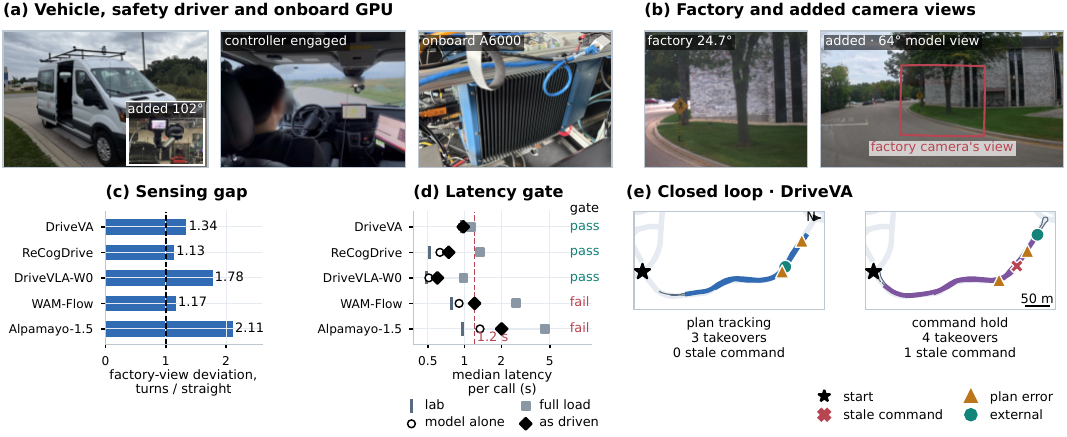}};
\node at (0,125.693) {\includegraphics[trim=0 125.693bp 315.949bp 12bp,clip]{vehicle_main_v2}};
\node at (203.051,125.693) {\includegraphics[trim=302bp 125.693bp 0 0,clip]{vehicle_main_v2}};
\node at (0,0) {\includegraphics{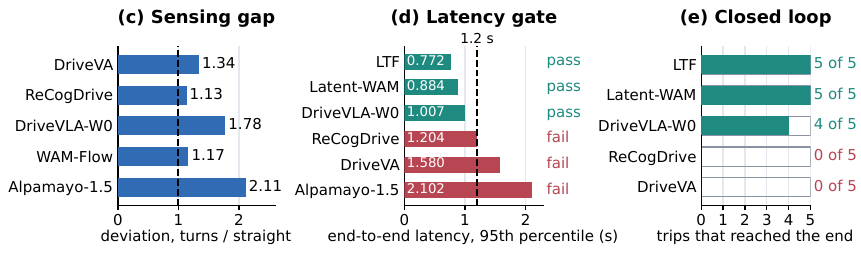}};
\end{tikzpicture}}
\caption{\textbf{The field study.} (a)~Inset: added camera. (b)~One turn in both views. (c)~Factory-view deviation, turns over straight. \trbnew{(d)~Onboard end-to-end latency, 95th percentile, against the $1.2$\,s deadline. (e)~Trips that reached the end of the $143$\,m route.}}
\label{fig:vehicle-gate}
\vspace{-8pt}
\end{wrapfigure}

We run \textsc{Triage} on a full-scale laboratory vehicle (\cref{fig:vehicle-gate}a; \cref{app:onvehicle}) to test the offline checks on a real platform; its onboard GPU is the RTX A6000 model behind every latency in \cref{tab:pool}.

\textbf{The sensing gate fails, and the prescribed repair works.} The $24.7^\circ$ factory front camera is narrower than every sighted sensor contract ($64$ to $110^\circ$), so no sighted candidate passes Stage~1, a physical instance of the crop of RQ1--RQ3. Following \cref{alg:triage}, we add a wide-angle front camera ($102^\circ$) and reproject its frames into the $64^\circ$ view the monocular candidates consume\trbnew{; two more added cameras supply the side views of the three-view candidates (\cref{fig:vehicle-rig})}. On a hand-driven lap with the models planning but not in control \trbnew{(Field experiment~III, \cref{app:veh-v3})}, all five candidates' plans from the factory view departed from those from the $64^\circ$ view more on turns than on straight road, by $1.13$ to $2.11\times$ (\cref{fig:vehicle-gate}b,\,c), and four of five erred more against the driver on turns, the turn-specific damage of RQ2; on the added view a crop again moved plans $2.4$--$6.4\times$ more than an equal-pixel resolution cut. \trbnew{In closed loop on a \TrbEsixRouteLen\,m route with a right turn (Field experiment~V, \cref{app:veh-e6}), the three-view LTF reached the end in all \TrbEsixZeroN{} trips; with its two side views blacked out it left its lane in both trips (\cref{tab:e6}), although on a hand-driven trip the blacked-out plans stayed within $1$\,m of the driven path at the median (\TrbEsixGateSb\,m).}

\trbnew{\textbf{The latency gate must use the end-to-end latency measured on board.} No sighted candidate plans within the vehicle's $10$\,ms control step, \trbold{so the vehicle tracks plans under a $1.2$\,s deadline.} Inference time alone already removes \trbold{Curious-VLA ($4.1$\,s), WAM-Flow (by $7$\,ms) and} the 10B-parameter Alpamayo-1.5. \trbold{With the same GPU model, weights and inputs, onboard} inference time \trbold{differed from the laboratory by over $10\%$ for three of five candidates, and under standard load three of five crossed the deadline} (Field experiment~I, \cref{app:veh-v1}). The controller, however, waits for the end-to-end latency, from the camera read of a plan's newest frame to the finished plan. Of six candidates measured on board, five meet the deadline on inference time and \TrbEfivePkNPass{} at the 95th percentile of the end-to-end latency; DriveVA and ReCogDrive, the two highest-scoring, fail (\cref{fig:vehicle-gate}d; \cref{tab:e5-parked}). Over \TrbNRuns{} closed-loop runs, DriveVA's end-to-end latency was \TrbCyclePfifty\,s at the median and \TrbCyclePninetyfiveLo{} to \TrbCyclePninetyfiveHi\,s at the 95th percentile per run, against \TrbInferPfifty\,s of inference: about \TrbOutsideModel\,s went to reprojecting, encoding and decoding frames, preprocessing and publishing, and each inference waited for the next frame of a \TrbSamplerPeriod\,s sampler. The end-to-end latency overran the deadline on every plan (shortest \TrbCycleMin\,s), \trbold{and the image behind the plan in use was $2.26$\,s old at the median.}}

\trbnew{\textbf{The two candidates the latency gate removes do not complete the route.} \trbold{Driven at $5.4$\,km/h with a safety driver at the controls} (Field experiment~IV, \cref{app:veh-v4}), DriveVA \trbold{drew three takeovers in $6.1$\,min under plan tracking and four in $7.4$\,min under command hold} (\cref{fig:vehicle-v4-map}); \trbold{a rater blind to the controller attributed one, under command hold, to stale commands.} On the \TrbEsixRouteLen\,m route of Field experiment~V, DriveVA and ReCogDrive reached the end in \TrbEsixZyDvaOk{} of \TrbEsixZyDvaN{} and \TrbEsixZyRcOk{} of \TrbEsixZyRcN{} trips, while the three candidates that pass the gate, DriveVLA-W0, LTF and Latent-WAM~\citep{wang2026latentwam}, a three-view model without a score in our pool, reached it in \TrbEsixWzOk{} of \TrbEsixWzN, \TrbEsixZeroOk{} of \TrbEsixZeroN{} and \TrbEsixZyLwamOk{} of \TrbEsixZyLwamN{} (\cref{fig:vehicle-gate}e; \cref{tab:e6}). For these five candidates as deployed, the gate's verdict matched the closed-loop outcome. ReCogDrive is a borderline case: it misses the deadline by \TrbEfivePkRcMarginMs\,ms, a margin too small to account for its trips. DriveVA, which overran the deadline on every plan, is the clear one. LTF, whose own end-to-end latency is \TrbEfiveLtfEteNewPninetyfive\,s, still completed every trip with its plans held for an extra $0.5$ or $1.0$\,s; at $2.0$\,s, a plan age of \TrbEsixTwoAge\,s at the controller, both trips ended in a takeover. At $4$\,m/s on this route the $1.2$\,s deadline is therefore conservative: driving failed between $1.0$ and $2.0$\,s of added delay.}

\begin{wraptable}{r}{0.6\linewidth}
\vspace{-10pt}
{\centering\scriptsize\trbnewon\setlength{\tabcolsep}{2pt}
\caption{\trbnew{\textbf{Closed loop on the \TrbEsixRouteLen\,m local route.} Plan age: median; steering-wheel rate: 95th percentile.}}
\label{tab:e6}
\begin{tabular*}{\linewidth}{@{\extracolsep{\fill}}lcccc@{}}
\toprule
Configuration & Reached & Takeover (m) & Plan age (s) & Steer (deg/s) \\
\midrule
DriveVLA-W0, front (selected) & \reachcell{\TrbEsixWzOk}{\TrbEsixWzN} & \TrbEsixWzIv & \TrbEsixWzAge & \TrbEsixWzRate \\
LTF, three views (challenger) & \reachcell{\TrbEsixZeroOk}{\TrbEsixZeroN} & -- & \TrbEsixZeroAge & \TrbEsixZeroRate \\
LTF, $+0.5$\,s & \reachcell{\TrbEsixHalfOk}{\TrbEsixHalfN} & -- & \TrbEsixHalfAge & \TrbEsixHalfRate \\
LTF, $+1.0$\,s & \reachcell{\TrbEsixOneOk}{\TrbEsixOneN} & -- & \TrbEsixOneAge & \TrbEsixOneRate \\
LTF, $+2.0$\,s & \reachcell{\TrbEsixTwoOk}{\TrbEsixTwoN} & \TrbEsixTwoIvA, \TrbEsixTwoIvB & \TrbEsixTwoAge & \TrbEsixTwoRate \\
LTF, side views blacked out & \reachcell{\TrbEsixSbOk}{\TrbEsixSbN} & \TrbEsixSbIvA, \TrbEsixSbIvB & \TrbEsixSbAge & \TrbEsixSbRate \\
LTF fine-tuned locally (Stage~5) & \reachcell{\TrbEsixFtOk}{\TrbEsixFtN} & \TrbEsixFtIvA, \TrbEsixFtIvB & \TrbEsixFtAge & \TrbEsixFtRate \\
\midrule
DriveVA, front & \reachcell{\TrbEsixZyDvaOk}{\TrbEsixZyDvaN} & \TrbEsixZyDvaIvMin{} to \TrbEsixZyDvaIvMax & -- & -- \\
ReCogDrive, front & \reachcell{\TrbEsixZyRcOk}{\TrbEsixZyRcN} & \TrbEsixZyRcIvMin{} to \TrbEsixZyRcIvMax & -- & -- \\
Latent-WAM, three views & \reachcell{\TrbEsixZyLwamOk}{\TrbEsixZyLwamN} & -- & -- & -- \\
\bottomrule
\end{tabular*}\par}
\vspace{-8pt}
\end{wraptable}

\trbnew{\textbf{Stage~4 selects a model that passes closed-loop acceptance.} \trbold{On local driving the probes separated every sighted candidate from the blind anchor} (Field experiment~II, \cref{app:veh-v2}). The four-city reference ranges did not carry over: in Field experiment~V temporal instability lay above its range for all four candidates that have one, so on this platform the range is reported and does not exclude a candidate (\cref{app:veh-nov-select}). Among the candidates that pass both gates, the ranking by benchmark score returned DriveVLA-W0, with LTF as challenger (\cref{tab:e5-stage3}); both passed closed-loop acceptance (\cref{tab:e6}, \cref{app:veh-e6}).}

\trbnew{\textbf{Stage~5's closed-loop check withholds a fine-tuned model.} Fine-tuned on \TrbEsevenNTrain{} local samples, LTF, which had completed every closed-loop trip, lowered its held-out displacement error with neither monitor firing (\cref{app:veh-stage5}), then drifted out of its lane in both closed-loop trips, and the closed-loop check withheld it. Neither the open-loop plan comparison nor the offline monitors anticipated the closed-loop failures of this section, which is why Stage~5 releases a model only after a closed-loop check.}

\section{Conclusion and limitations}
\label{sec:limits}
\label{sec:conclusion}
\leavevmode\rv{Choosing an embodied foundation model for a fixed platform is not a leaderboard lookup: the checkpoint--platform gap splits by what can repair it, and only its distribution part is fine-tuning's problem. \textsc{Triage} turns this into feasibility-gated selection and monitored adaptation\trbnew{: gates that no score overrides, each naming the repair its failure calls for; a ranking by the target-relevant score among the candidates that pass; and fine-tuning monitored for the robustness it can cost}. \trbnew{In a field study on a full-scale laboratory vehicle that ran all five stages}, the sensing gate and its turn-specific damage transferred, while laboratory latency, even on the same GPU model, and \trbnew{the probes' reference ranges, calibrated on benchmark logs,} did not; both must be measured on-platform\trbnew{, latency as the end-to-end latency}. \trbnew{In closed loop, the gates' verdicts held, and a model that fine-tuning had improved offline failed.} A benchmark score informs deployment only with its sensor contract and latency.

\textbf{Limitations.} \trbnew{\textsc{Triage} targets teams starting from public checkpoints with little local data. For a team training a reused architecture on a large in-house corpus, Stages~1 and~2 still apply, since they concern the platform and its control loop, not the weights; Stages~3 to~5 are untested at that scale, including whether public weights still help. \textsc{Triage} names the class of repair a failing gate calls for, not production hardware.} Closed-loop evidence is simulated and real-data evidence open loop, apart from \trbnew{a short field study with one vehicle, two local routes and one rater}. Repair covers behavior cloning at modest budgets on three architectures; manipulation is offline, and RQ5 uses two checkpoint pairs.}

\clearpage
\subsubsection*{Acknowledgments}
This research was supported by the U.S. Department of Transportation (No.~69A3552348305) and by grants from NVIDIA utilizing NVIDIA RTX PRO 6000 Blackwell Max-Q Workstation Edition GPUs.

\subsubsection*{Reproducibility statement}
\rv{All evaluations use public checkpoints and public benchmarks at fixed versions: CARLA~0.9.15 with Bench2Drive, and the NAVSIM~v1.1 evaluator on the \texttt{navtest} split. A configuration with a published score entered the pool only after that score was reproduced under the same evaluator, with one documented exception; closed-loop and open-loop results are never pooled, missing measurements are never imputed, and fine-tuning data are log-level disjoint from every evaluation window (\cref{app:protocols}). Every reported quantity is defined in \cref{app:instruments}, the formal statements with their assumptions and proof sketches are in \cref{app:theory}, and \cref{alg:triage} with its costs is in \cref{app:procedure}. \cref{app:extended} gives the complete results of every research question, including per-city and per-seed values; \cref{app:twins} documents the search for checkpoint pairs; and \cref{app:onvehicle} describes the vehicle, its calibration and the procedure and results of each field experiment. Code, probe definitions and per-run outputs will be released under an open-source license.}

\subsubsection*{Ethics statement}
This work evaluates publicly released driving and manipulation checkpoints on public benchmark logs, in simulation, and in a field study on a laboratory vehicle. The field study is not a study of human participants: it was carried out by members of our laboratory, a safety driver who had completed the institution's Institutional Review Board (IRB) certification training and an operator, and its recordings of public roads are used only for the measurements reported here. \rv{Closed-loop safety outcomes
were measured in simulation, apart from the closed loop of
the field study, driven with speed commands capped at $1.5$\,m/s \trbnew{for the controller contrast and $4.0$\,m/s on the \TrbEsixRouteLen\,m route,} and a safety driver
ready to take over. The field study ran on low-traffic business-park
and residential roads posted at $25$\,mph or less, under safety abort rules fixed before the drive.}
Redistribution of third-party weights follows their licenses; \rv{version identifiers and download scripts} are released, not the weights.

\subsubsection*{AI use statement}
\rv{We used generative AI tools (large language model assistants) to implement methods, by writing, debugging and running evaluation, analysis and figure scripts on archived outputs; to discuss design options for the methodology and experiments; and to check the formal statements and proofs of \cref{app:theory}. We have not used generative AI tools to generate or label data: every takeover in the field study was rated by a human rater. Additionally, we used generative AI tools to draft and edit paper text, including the abstract, the conclusion and figure captions; to search for and summarize related work and public checkpoints; and to check the numbers in the text against the archived outputs. We have reviewed all AI-assisted work: every cited work was checked against its source and every reported number against the archived outputs.} We take responsibility for the final content of this work, including text, claims or artifacts produced with the aid of generative AI.

\bibliography{main}
\bibliographystyle{iclr2027_conference}

\clearpage
\appendix
\makeatletter\setlength{\@fptop}{0pt}\setlength{\@fpsep}{12pt}\makeatother
\FloatBarrier
\FloatBarrier
\section{Details of the real-world field study}
\label{app:onvehicle}
\providecommand{\vrfmt}[1]{#1}
\providecommand{\vr}[1]{\ifcsname vr@#1\endcsname\vrfmt{\csname vr@#1\endcsname}\else\pend{}\fi}
\expandafter\def\csname vr@v1.alpamayo.iso50\endcsname{1337}
\expandafter\def\csname vr@v1.alpamayo.iso95\endcsname{2289}
\expandafter\def\csname vr@v1.alpamayo.load50\endcsname{4531}
\expandafter\def\csname vr@v1.alpamayo.load95\endcsname{7490}
\expandafter\def\csname vr@v1.alpamayo.nd50\endcsname{2014}
\expandafter\def\csname vr@v1.alpamayo.nd95\endcsname{3452}
\expandafter\def\csname vr@v1.alpamayo.ndover\endcsname{100.0}
\expandafter\def\csname vr@v1.driveva.iso50\endcsname{947}
\expandafter\def\csname vr@v1.driveva.iso95\endcsname{949}
\expandafter\def\csname vr@v1.driveva.load50\endcsname{1133}
\expandafter\def\csname vr@v1.driveva.load95\endcsname{1174}
\expandafter\def\csname vr@v1.driveva.nd50\endcsname{976}
\expandafter\def\csname vr@v1.driveva.nd95\endcsname{981}
\expandafter\def\csname vr@v1.driveva.ndover\endcsname{0.0}
\expandafter\def\csname vr@v1.drivevlaw0.iso50\endcsname{509}
\expandafter\def\csname vr@v1.drivevlaw0.iso95\endcsname{515}
\expandafter\def\csname vr@v1.drivevlaw0.load50\endcsname{980}
\expandafter\def\csname vr@v1.drivevlaw0.load95\endcsname{1051}
\expandafter\def\csname vr@v1.drivevlaw0.nd50\endcsname{598}
\expandafter\def\csname vr@v1.drivevlaw0.nd95\endcsname{606}
\expandafter\def\csname vr@v1.drivevlaw0.ndover\endcsname{0.0}
\expandafter\def\csname vr@v1.recogdrive.iso50\endcsname{627}
\expandafter\def\csname vr@v1.recogdrive.iso95\endcsname{643}
\expandafter\def\csname vr@v1.recogdrive.load50\endcsname{1334}
\expandafter\def\csname vr@v1.recogdrive.load95\endcsname{1497}
\expandafter\def\csname vr@v1.recogdrive.nd50\endcsname{741}
\expandafter\def\csname vr@v1.recogdrive.nd95\endcsname{840}
\expandafter\def\csname vr@v1.recogdrive.ndover\endcsname{0.0}
\expandafter\def\csname vr@v1.wamflow.iso50\endcsname{900}
\expandafter\def\csname vr@v1.wamflow.iso95\endcsname{935}
\expandafter\def\csname vr@v1.wamflow.load50\endcsname{2657}
\expandafter\def\csname vr@v1.wamflow.load95\endcsname{3010}
\expandafter\def\csname vr@v1.wamflow.nd50\endcsname{1207}
\expandafter\def\csname vr@v1.wamflow.nd95\endcsname{1262}
\expandafter\def\csname vr@v1.wamflow.ndover\endcsname{59.5}
\expandafter\def\csname vr@v2.driveva.ag\endcsname{0.245}
\expandafter\def\csname vr@v2.driveva.aj\endcsname{0.182}
\expandafter\def\csname vr@v2.driveva.dgray\endcsname{3.15}
\expandafter\def\csname vr@v2.driveva.ti\endcsname{0.70}
\expandafter\def\csname vr@v2.drivevlaw0.ag\endcsname{0.891}
\expandafter\def\csname vr@v2.drivevlaw0.aj\endcsname{0.680}
\expandafter\def\csname vr@v2.drivevlaw0.dgray\endcsname{5.72}
\expandafter\def\csname vr@v2.drivevlaw0.ti\endcsname{1.39}
\expandafter\def\csname vr@v2.recogdrive.ag\endcsname{0.179}
\expandafter\def\csname vr@v2.recogdrive.aj\endcsname{0.135}
\expandafter\def\csname vr@v2.recogdrive.dgray\endcsname{2.11}
\expandafter\def\csname vr@v2.recogdrive.ti\endcsname{0.84}
\expandafter\def\csname vr@v2.wamflow.ag\endcsname{0.280}
\expandafter\def\csname vr@v2.wamflow.aj\endcsname{0.139}
\expandafter\def\csname vr@v2.wamflow.dgray\endcsname{1.34}
\expandafter\def\csname vr@v2.wamflow.ti\endcsname{0.71}
\expandafter\def\csname vr@v3.alpamayo.crop\endcsname{0.73}
\expandafter\def\csname vr@v3.alpamayo.kappa\endcsname{3.31}
\expandafter\def\csname vr@v3.alpamayo.res\endcsname{0.22}
\expandafter\def\csname vr@v3.driveva.crop\endcsname{0.94}
\expandafter\def\csname vr@v3.driveva.kappa\endcsname{4.26}
\expandafter\def\csname vr@v3.driveva.res\endcsname{0.22}
\expandafter\def\csname vr@v3.drivevlaw0.crop\endcsname{1.61}
\expandafter\def\csname vr@v3.drivevlaw0.kappa\endcsname{2.37}
\expandafter\def\csname vr@v3.drivevlaw0.res\endcsname{0.68}
\expandafter\def\csname vr@v3.recogdrive.crop\endcsname{0.42}
\expandafter\def\csname vr@v3.recogdrive.kappa\endcsname{5.00}
\expandafter\def\csname vr@v3.recogdrive.res\endcsname{0.08}
\expandafter\def\csname vr@v3.wamflow.crop\endcsname{0.61}
\expandafter\def\csname vr@v3.wamflow.kappa\endcsname{6.36}
\expandafter\def\csname vr@v3.wamflow.res\endcsname{0.10}

This appendix details the real-world field study of
\cref{sec:casestudy}, which applies \textsc{Triage} to a \textbf{fixed physical platform} (a full-scale vehicle retrofitted for autonomous driving). We
ran \trbnew{five field experiments}, each testing one part of the \textsc{Triage} workflow (\cref{fig:hero}d) against its results on Bench2Drive \citep{jia2024bench} and NAVSIM \citep{dauner2024navsim}.
\cref{app:veh-platform} describes the vehicle and its cameras, lists which candidates pass the sensing and latency gates, and gives the route and how the data were collected. Field
experiment~I (\cref{app:veh-v1}) re-measures the latency of candidates on the onboard GPU (the \textbf{latency gate}); Field experiment~II
(\cref{app:veh-v2}) runs the probes on a local recording\trbnew{ (Stage~3)}; Field experiment~III (\cref{app:veh-v3}) measures the
\textbf{sensing gate} on vehicle frames, comparing the factory camera with
the added one; and Field experiment~IV (\cref{app:veh-v4}) drives DriveVA \citep{driveva2026} in closed loop under command hold and plan tracking (the latency gate's consumption rule)\trbnew{. Field experiment~V uses all three added cameras; \cref{app:veh-nov} describes them, the routes and the procedure (\cref{fig:vehicle-rig}), the end-to-end latency of its six candidates is set beside their inference times in \cref{tab:e5-parked}, \cref{app:veh-nov-select} runs Stages~3 and~4 on its recordings, \cref{app:veh-stage5} runs Stage~5 offline on them, and Field experiment~V (\cref{app:veh-e6}) drives the resulting choice and tests both gates and Stage~5 in closed loop}.

\begin{figure}[!ht]
\centering
\includegraphics[width=\linewidth]{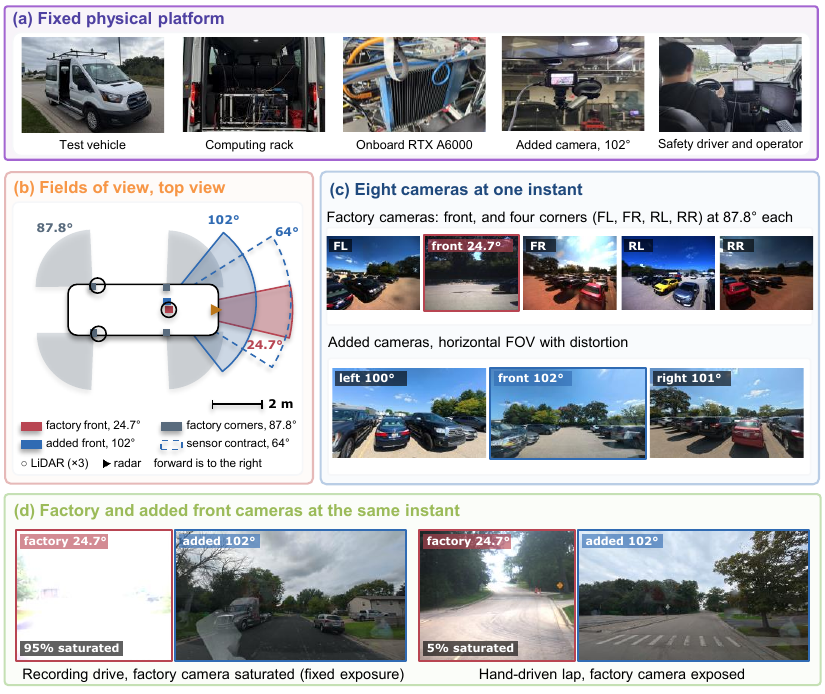}
\caption{\textbf{The fixed physical platform.} (a)~Vehicle, compute rack, added camera, safety driver and operator. (b)~Fields of view from above: factory front ($24.7^\circ$, red) and corner ($87.8^\circ$) cameras, added front camera ($102^\circ$, blue) and the $64^\circ$ sensor contract (dashed). (c)~All eight cameras at one instant. (d)~Factory and added front cameras on the recording drive and on a hand-driven lap.}
\label{fig:vehicle-platform}
\end{figure}

\FloatBarrier

\subsection{Platform, candidates, route and data collection}
\label{app:veh-platform}

\paragraph{Fixed physical platform.} The test vehicle is a full-size electric van
retrofitted with drive-by-wire, an inertial navigation system and an onboard RTX A6000, the same GPU model as the laboratory workstation on which every latency in \cref{tab:pool} was measured (\cref{fig:vehicle-platform}a). Its factory sensors are a forward
camera ($24.7^\circ$ horizontal, $2448\times2048$, mounted $2.45$\,m
high), four corner cameras ($87.8^\circ$ each, yawed about $\pm44^\circ$
and $\pm135^\circ$), three lidars and a front radar
(\cref{fig:vehicle-platform}b,\,c); these factory cameras are the fixed
camera setup that every candidate is first checked against. Because
they fail the sensing gate (below), we added three DJI wide-angle
cameras, front, left and right ($1920\times1080$, $100$ to $102^\circ$
horizontal measured with lens distortion; \cref{fig:vehicle-platform}c),
of which only the front one, mounted $1.83$\,m high and pitched
$8.9^\circ$ upward, was used in \trbnew{Field experiments~I to~IV; Field experiment~V used all three views (\cref{app:veh-nov})}. The front added camera was
read at $29.9$\,frames per second over USB by the inference process and
at $10$\,Hz as republished on the vehicle network for recording; its
frames are published without undistortion, which our reprojection
performs. \cref{fig:vehicle-platform}d shows the factory and added front cameras at the same instant.

\begin{table}[!ht]
\centering\small
\caption{\textbf{Sensing and latency gates for the candidates considered on the vehicle.} \emph{Factory}, \emph{Added}: the sensing gate without and with the added front camera\trbnew{; LTF and Latent-WAM need three views and pass with the three added cameras of Field experiment~V}; \trbnew{\emph{Lab p50}: laboratory inference time against the $1.2$\,s deadline, a first screen.} Blue: passes sensing; green: passes \trbnew{the screen}; red: reason for exclusion; --: not reached. \trbnew{On board, the end-to-end latency removes DriveVA and ReCogDrive (\cref{tab:e5-parked}) and inference time alone WAM-Flow and Alpamayo-1.5 (\cref{app:veh-v1}).} Sources: constant velocity \citep{dauner2024navsim}, EgoStatusMLP \citep{li2024egostatus}, DriveVA \citep{driveva2026}, ReCogDrive \citep{recogdrive2026}, DriveVLA-W0 \citep{li2026drivevlaw0}, WAM-Flow \citep{xu2026wamflow}, Alpamayo-1.5 \citep{alpamayo15}, Curious-VLA \citep{curious2026}, AutoVLA \citep{zhou2025autovla}, TransFuser \citep{chitta2022transfuser}, DiffusionDrive \citep{liao2025diffusiondrive}, SimLingo \citep{renz2025simlingo}, UniAD \citep{hu2023uniad}, VAD \citep{jiang2023vad}, ORION \citep{fu2025orion}, TCP \citep{wu2022tcp}.}
\label{tab:veh-gates}
\begin{tabularx}{\linewidth}{@{}l>{\raggedright\arraybackslash}p{0.165\linewidth}ccrc>{\raggedright\arraybackslash}X@{}}
\toprule
Candidate & Sensor contract & Factory & Added & \begin{tabular}[b]{@{}r@{}}Lab p50\\(ms)\end{tabular} & \begin{tabular}[b]{@{}c@{}}\trbnew{Lab}\\\trbnew{screen}\end{tabular} & \trbnew{Outcome} \\
\midrule
\multicolumn{7}{@{}l}{\emph{Blind anchors}}\\
Constant velocity & none & pass & pass & $0.01$ & pass & reference \\
EgoStatusMLP & ego state only & pass & pass & $0.12$ & pass & reference \\
\midrule
\multicolumn{7}{@{}l}{\emph{NAVSIM candidates}}\\
DriveVA & front $64^\circ$, video & fail & \cellcolor{passblue}pass & $947$ & \cellcolor{passgreen}pass & \trbnew{highest PDMS; fails the gate on the end-to-end latency} \\
ReCogDrive & front $64^\circ$ & fail & \cellcolor{passblue}pass & $513$ & \cellcolor{passgreen}pass & \trbnew{second-highest PDMS; fails the gate on the end-to-end latency} \\
DriveVLA-W0 & front $64^\circ$ & fail & \cellcolor{passblue}pass & $485$ & \cellcolor{passgreen}pass & \trbnew{selected (\cref{tab:e5-stage3})} \\
WAM-Flow & front $64^\circ$ & fail & \cellcolor{passblue}pass & $781$ & \cellcolor{passgreen}pass & \trbnew{fails on onboard inference time} \\
Alpamayo-1.5 & flexible (fed $64^\circ$) & fail & \cellcolor{passblue}pass & $959$ & \cellcolor{passgreen}pass & no NAVSIM score; latency and probes only \\
Curious-VLA & front $64^\circ$ & fail & \cellcolor{passblue}pass & \cellcolor{failred}$4{,}067$ & fail & \\
AutoVLA & \trbnew{three front cameras} & fail & fail & \cellcolor{failred}$1{,}249$ & \trbnew{fail} & \trbnew{inference time over the $1.2$\,s deadline} \\
\trbnew{LTF} & \trbnew{three cameras} & \trbnew{fail} & \trbnew{fail} & \trbnew{$69$} & \trbnew{--} & \trbnew{passes with three added cameras; challenger (\cref{tab:e5-stage3})} \\
\trbnew{Latent-WAM} & \trbnew{three cameras} & \trbnew{fail} & \trbnew{fail} & \trbnew{--} & \trbnew{--} & \trbnew{passes with three added cameras; no score in the pool} \\
TransFuser & three cameras + LiDAR & \multicolumn{2}{c}{not evaluated} & $69$ & -- & LiDAR differs in model and mounting \\
DiffusionDrive & three cameras + LiDAR & \multicolumn{2}{c}{not evaluated} & $85$ & -- & as TransFuser \\
\midrule
\multicolumn{7}{@{}l}{\emph{Bench2Drive candidates}}\\
SimLingo & front $110^\circ$ & fail & fail & $598$ & -- & $102^\circ < 110^\circ$ \\
UniAD front-only & front $70^\circ$ & fail & \cellcolor{passblue}pass & $722$ & pass & \cellcolor{failred}trained in simulation, no real-data score; \trbnew{not ranked} \\
UniAD & six cameras ($5\times70^\circ$, rear $110^\circ$) & fail & fail & $722$ & -- & no rear camera \\
VAD & six cameras & fail & fail & $567$ & -- & no rear camera \\
ORION & six cameras & fail & fail & $1{,}020$ & -- & no rear camera \\
TCP & three $70^\circ$ front cameras, stitched & fail & fail & $63$ & -- & \cellcolor{failred}\trbnew{trained in simulation, no real-data score} \\
\bottomrule
\end{tabularx}
\end{table}

\paragraph{Sensing gate: factory versus added cameras.} On the factory cameras every sighted sensor contract fails (\cref{tab:veh-gates}): the front view is narrower than every monocular contract, and the multi-camera contracts need views the vehicle lacks. With the added camera, the six monocular NAVSIM candidates pass through the $64^\circ$ virtual view.

\begin{figure}[!ht]
\centering
\includegraphics[width=\linewidth]{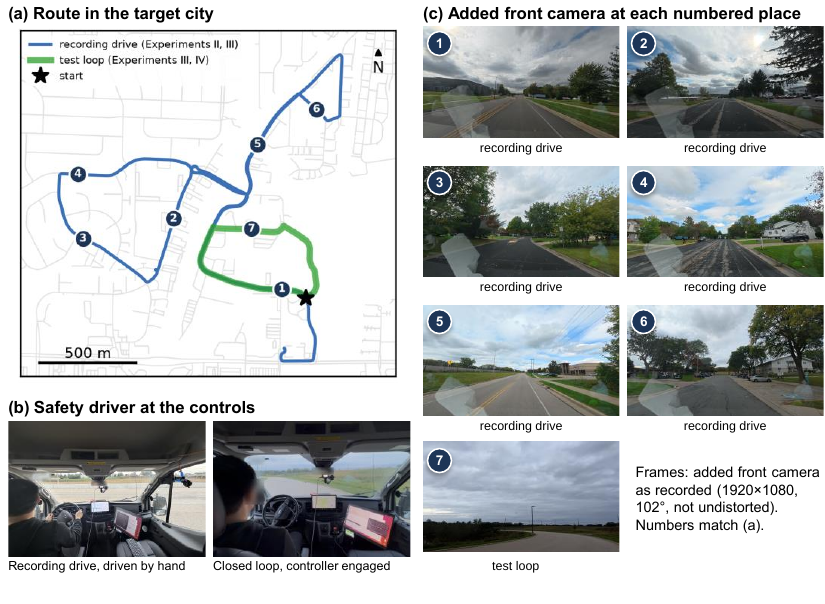}
\caption{\textbf{Route and data collection.} (a)~Recording drive (blue; Field experiments~II and~III) and test loop (green; Field experiments~III and~IV); numbers mark the places in (c), the star the start of the loop. (b)~The safety driver. (c)~The added front camera at each place.}
\label{fig:vehicle-route}
\end{figure}

\paragraph{Camera calibration and virtual views.} The added front
camera's intrinsics and distortion come from a checkerboard calibration
with the same field-of-view mode and stabilization setting, and its
extrinsics from a lidar-to-camera calibration of the vehicle. The virtual view
reprojects its frames into the NAVSIM front camera's geometry
($f_x=1545$\,px at $1920\times1080$ with nuPlan distortion) at the
median orientation of the nuPlan front camera. Fields of view are horizontal. The added
camera is mounted higher than the nuPlan capture vehicle's front camera
($1.83$ against $1.52$\,m).

\paragraph{Route.} All driving took place on business-park and
residential roads in the target city, posted at $25$\,mph or less (\cref{fig:vehicle-route}a, with the added camera's view at seven places in \cref{fig:vehicle-route}c). Field experiments~II and~III use a $24$-minute
recording driven over $7.7$\,km through places 1 to 6 of \cref{fig:vehicle-route}. The hand-driven laps of Field experiment~III and the closed-loop runs of Field experiment~IV use a $1.66$\,km test loop with six turns (place 7 in \cref{fig:vehicle-route}), driven counterclockwise from a fixed start at an intersection.

\paragraph{Data collection and safety.} A safety driver who had completed the institution's Institutional Review Board (IRB) certification training drove
the recording and the hand-driven laps in normal traffic, with an
operator on board running the recording and inference processes (\cref{fig:vehicle-route}b). In closed loop the safety driver stayed
at the controls: the controller was engaged only with both feet off the
pedals, \trbnew{in Field experiment~IV speed commands were capped at $1.5$\,m/s and} the controller stopped the
vehicle whenever no new plan arrived within $2.2$\,s \trbnew{(the settings of Field experiment~V are in \cref{app:veh-nov})}, and any brake, throttle or steering input by the driver disengaged it at once.
Such a disengagement, read from the controller's own enable flag together with the drive-by-wire override flags, counts as a takeover; each takeover was classified afterwards by a rater blind to the controller (\cref{app:veh-v4}).

\FloatBarrier

\subsection{Field experiment I: latency on the vehicle}
\label{app:veh-v1}

\paragraph{Setup.} Five candidates were timed on the onboard
A6000 with no other process on the GPU: DriveVA, ReCogDrive,
DriveVLA-W0, WAM-Flow and Alpamayo-1.5. Alpamayo-1.5 has no NAVSIM score to rank by, so it is timed here but takes no part in the selection. Each ran 200 timed calls after 20 warm-up calls in each of three conditions: the
model process alone (\emph{alone}); the vehicle's standard start-up with cameras,
localization, logging and its three display tools running, the model not controlling the vehicle (\emph{full load}); and the same load with the
display tools and the added camera's network node closed, the state in which Field experiment IV was driven (\emph{deployed}). \trbnew{Inference time is compared with the deadline in the deployed condition.}

\begin{table}[!ht]
\centering\small
\caption{\textbf{Field experiment I: latency on the vehicle} (ms). \emph{Alone}: model process only; \emph{full load}: standard start-up with display tools; \emph{deployed}: without display tools, as in Field experiment IV. \emph{Over $1.2$\,s}: share of deployed calls above the deadline. Green: deployed median within the deadline; red: over it.}
\label{tab:v1}
\setlength{\tabcolsep}{3.5pt}\begin{tabularx}{\linewidth}{@{}>{\raggedright\arraybackslash}p{0.15\linewidth}*{8}{Y}@{}}
\toprule
Candidate & Lab p50 & Alone p50 & Alone p95 & Full load p50 & Full load p95 & Deployed p50 & Deployed p95 & Over $1.2$\,s (\%) \\
\midrule
DriveVA & 947 & \vr{v1.driveva.iso50} & \vr{v1.driveva.iso95} & \vr{v1.driveva.load50} & \vr{v1.driveva.load95} & \cellcolor{passgreen}\vr{v1.driveva.nd50} & \vr{v1.driveva.nd95} & \vr{v1.driveva.ndover} \\
ReCogDrive-2B & 513 & \vr{v1.recogdrive.iso50} & \vr{v1.recogdrive.iso95} & \vr{v1.recogdrive.load50} & \vr{v1.recogdrive.load95} & \cellcolor{passgreen}\vr{v1.recogdrive.nd50} & \vr{v1.recogdrive.nd95} & \vr{v1.recogdrive.ndover} \\
DriveVLA-W0 & 485 & \vr{v1.drivevlaw0.iso50} & \vr{v1.drivevlaw0.iso95} & \vr{v1.drivevlaw0.load50} & \vr{v1.drivevlaw0.load95} & \cellcolor{passgreen}\vr{v1.drivevlaw0.nd50} & \vr{v1.drivevlaw0.nd95} & \vr{v1.drivevlaw0.ndover} \\
WAM-Flow & 781 & \vr{v1.wamflow.iso50} & \vr{v1.wamflow.iso95} & \vr{v1.wamflow.load50} & \vr{v1.wamflow.load95} & \cellcolor{failred}\vr{v1.wamflow.nd50} & \vr{v1.wamflow.nd95} & \vr{v1.wamflow.ndover} \\
Alpamayo-1.5 & 959 & \vr{v1.alpamayo.iso50} & \vr{v1.alpamayo.iso95} & \vr{v1.alpamayo.load50} & \vr{v1.alpamayo.load95} & \cellcolor{failred}\vr{v1.alpamayo.nd50} & \vr{v1.alpamayo.nd95} & \vr{v1.alpamayo.ndover} \\
\bottomrule
\end{tabularx}
\end{table}

\paragraph{Laboratory latency does not carry over to the vehicle.} On fixed test inputs the models on the vehicle reproduced their laboratory outputs. Alpamayo-1.5, whose reasoning length depends on the image, was timed on the 200 windows of its laboratory measurement with the same settings. With the same GPU model, weights
and inputs, DriveVA kept its laboratory latency ($947$\,ms, \cref{tab:v1}) and
DriveVLA-W0 rose by $5\%$, but WAM-Flow rose by $15\%$, ReCogDrive by
$22\%$ and Alpamayo-1.5 by $39\%$ ($959$ to $1337$\,ms, $174$ of $200$
calls over $1.2$\,s). For three of five candidates, a latency measured on one computer did not carry over to another with the same GPU model, which is why Field experiment I re-measures before any later stage.

\paragraph{Under full load three of five candidates cross the deadline.} With the vehicle's standard start-up running, the three display tools and the camera compression took most of the computer's processor time. Median latency rose to $1133$\,ms for DriveVA,
$980$\,ms for DriveVLA-W0, $1334$\,ms for ReCogDrive, $2657$\,ms for
WAM-Flow and $4531$\,ms for Alpamayo-1.5.

\paragraph{The display tools account for most of the slowdown.} A second measurement under load closed the display tools and the added cameras' nodes, the state in which Field experiment IV was driven: factory cameras, localization, logging and control services running, vehicle parked (\cref{tab:v1}, \emph{deployed}). Median latency was then $976$\,ms for DriveVA,
$598$\,ms for DriveVLA-W0, $741$\,ms for ReCogDrive, $1207$\,ms for
WAM-Flow and $2014$\,ms for Alpamayo-1.5, between $3\%$ and $51\%$ above the readings with the model alone,
so the display tools, not the sensors, account for most of the slowdown under full load. \trbnew{Inference time is compared with the deadline on this measurement because it matches the deployment: on inference time} DriveVA, ReCogDrive and DriveVLA-W0 meet the
$1.2$\,s deadline; WAM-Flow misses it by $7$\,ms and is excluded; Alpamayo-1.5 misses it.

\paragraph{Summary.} Laboratory latency did not carry over to the vehicle: with the model alone, three of five candidates ran more than $10\%$ slower than in the laboratory on the same GPU model, and under full load three of five crossed the $1.2$\,s deadline, mostly because of the display tools. \trbnew{In the deployed condition, inference time alone removes WAM-Flow and Alpamayo-1.5. Inference time is a lower bound on the end-to-end latency that the gate of \cref{sec:gates} uses: on the end-to-end latency DriveVA and ReCogDrive also fail (\cref{tab:e5-parked}, which sets the three measures side by side).}

\FloatBarrier

\subsection{Field experiment II: probes on a local recording}
\label{app:veh-v2}

\paragraph{Setup.} The safety driver drove the recording route (\cref{fig:vehicle-route}) for $24$ minutes with the factory and added front cameras recording.
On the virtual view of this recording, the probe readings of
\cref{app:instruments} (temporal instability; the darkening and
compression responses $A_\gamma$ and $A_{\mathrm{jpeg}}$; the response
to a gray frame, $D_{\mathrm{gray}}$; and run-to-run noise, from six
repeated calls on one frame) were computed after the drive for DriveVA,
ReCogDrive, DriveVLA-W0, WAM-Flow and the constant-velocity anchor. A candidate's readings count only if its sign diagnostic, the share of turning frames on which the plan turns the way the driver did, is at least $0.8$. The probes should separate every sighted candidate from the blind anchor: the anchor responds exactly zero, and each sighted candidate's $D_{\mathrm{gray}}$ exceeds twice its run-to-run noise. \trbnew{Each reading is compared with its four-city range through a 95\% interval from \TrbRrB{} resamples of the recording's windows, drawn independently and in blocks of \TrbRrBlock{} windows; a reading leaves its range only if its whole interval does.}

\paragraph{Every sighted candidate is separated from the blind anchor.} The
recording lasted $23.9$\,min ($18.3$\,min moving, $7.7$\,km, mean
moving speed $25$\,km/h) and yields $2842$ five-frame windows; both
front cameras were recorded at $10$\,Hz and the navigation system at
$100$\,Hz. Replacing every frame with gray moved DriveVA's plan by $3.15$\,m and ReCogDrive's by $2.11$\,m, far above their run-to-run noise ($0.208$\,m and $0$), and it moved
DriveVLA-W0 by $5.72$\,m and WAM-Flow by $1.34$\,m (run-to-run noise $0$ and $0.068$\,m), so all four candidates are separated from the anchor (\cref{tab:e5-stage3}, upper block).
Sign diagnostics were healthy (DriveVA $0.968$ and ReCogDrive $0.974$, on over $1000$ turning windows each).

{\trbnewon\paragraph{No candidate's readings lie within its four-city ranges.} With block resampling, the temporal-instability intervals lie entirely above the ranges for DriveVA (\TrbRrTiDva{} against \TrbRrRgDva\,m), for DriveVLA-W0 (\TrbRrTiWz{} against \TrbRrRgWz\,m) and for WAM-Flow (\TrbRrTiWf{} against \TrbRrRgWf\,m); ReCogDrive's (\TrbRrTiRc{} against \TrbRrRgRc\,m) touches its range only under block resampling, while \trbold{its darkening and compression responses ($0.179$ and $0.135$}; intervals \TrbRrAgRc{} and \TrbRrAjRc) lie entirely below their ranges \trbold{($0.206$ to $0.235$ and $0.173$ to $0.195$)} under both schemes. Standardized over the four cities, the six readings of DriveVA and ReCogDrive lie nearest Pittsburgh, but \trbold{farther from it ($6.88$ in standardized units) than any two cities lie from each other ($2.0$ to $4.9$).}} A shift shared by four architectures points to a common input path, such as the reprojected virtual camera, rather than to one model.\par

\paragraph{Summary.} On the local recording the probes separated every sighted candidate from the blind anchor. \trbnew{No candidate's readings lay within its four-city ranges, a shift shared by all four architectures, so on this platform the range is reported beside the ranking and does not exclude a candidate (\cref{app:veh-nov-select}).}

\FloatBarrier

\subsection{Field experiment III: the sensing gate on vehicle frames}
\label{app:veh-v3}

\paragraph{Setup.} On the virtual view of every tenth window of the local recording,
the response curves of RQ1 (center crops keeping $90$, $75$ and $60\%$
of each image dimension; resolution reduced to $75$, $50$ and $33\%$)
were repeated for the four candidates with an offline ratio (DriveVA,
ReCogDrive, DriveVLA-W0, WAM-Flow), and for Alpamayo-1.5, which has no offline ratio. Offline, the crop-over-resolution ratio at the matched budget exceeded one in every driving and manipulation model. The turn/straight comparison measures the factory camera's sensing gap on synchronous frames: each of the
five models plans on the factory view, fed unchanged as its front
camera, and on the $64^\circ$ virtual view, frames paired within
$0.1$\,s, and windows are split into turning (heading change of at
least $15^\circ$ over the next $2$\,s) and straight. A sensing gap shows as the $2$\,s plan points of the two views lying farther apart on turning than on straight windows, the state-conditional signature RQ2 measured under
a synthetic crop. We also compare the factory view's error against the driver's path at $2$\,s on turning and straight windows, and its departure from the virtual view with the offline displacement at the strongest crop.

\paragraph{A crop moves plans more than a resolution cut.} At the matched budget of $56\%$ of the pixels the crop moved plans by $0.42$ to $1.61$\,m and the resolution cut by $0.08$ to $0.68$\,m (\cref{tab:r7-detail}), so the ratio was $4.26$ for DriveVA, $5.00$ for ReCogDrive, $2.37$ for DriveVLA-W0 and $6.36$ for WAM-Flow: above one for every candidate and at or above each offline value; Alpamayo-1.5 gives $3.31$.

\paragraph{The factory camera's plans depart most on turns.} The factory camera, set to a fixed exposure, was overexposed throughout the local recording (median brightness $255$ of $255$; $97\%$ of frames more than half saturated, in one frame per second), while the added camera was not (median $114$; \cref{fig:vehicle-platform}d), so the turn/straight comparison uses the second hand-driven lap of the test loop ($32\%$ of frames more than half saturated, against $56\%$ on the first), where $22$ of $889$ usable windows were turning. On this lap all five models show the signature: the factory camera's plans departed from the virtual view's more on turning than on straight windows, by $1.13$ to $2.11$ times (\cref{tab:r7-detail}). Against the driver's path at $2$\,s, the factory view's plans erred more than the virtual view's on turns for four of the five models (DriveVA $2.60$ against $0.97$\,m), ReCogDrive being the exception; and over all windows they departed from the virtual view's by more than the offline displacement at the strongest crop for each of the four models with an offline result (DriveVA $1.64$ against $0.91$\,m), consistent with a $24.7^\circ$ view spanning about a third of the image. \cref{fig:vehicle-gate}b shows one turning window of the hand-driven lap.

\begin{table}[!ht]
\centering\small\setlength{\tabcolsep}{2pt}
\caption{\textbf{Field experiment III: the sensing gate on vehicle frames} (m). \emph{Crop against resolution}: plan displacement at the matched budget ($56\%$ of pixels) on every tenth window of the local recording ($285$ windows), with the offline ratio from RQ1. \emph{Factory against virtual view}: second hand-driven lap, 22 turning and 867 straight windows. \emph{Deviation}: distance between the two views' $2$\,s plan points; \emph{error to driver}: virtual / factory view; \emph{offline crop}: offline displacement at the $60\%$ crop. Green: ratio above one.}
\label{tab:r7-detail}
\fontsize{8}{9.5}\selectfont\begin{tabular*}{\linewidth}{@{\extracolsep{\fill}}lrrrrrrrrrrr@{}}
\toprule
 & \multicolumn{4}{c}{Crop against resolution} & \multicolumn{3}{c}{Factory vs virtual} & \multicolumn{2}{c}{Error to driver} & \multicolumn{2}{c}{All windows} \\
\cmidrule(lr){2-5}\cmidrule(lr){6-8}\cmidrule(lr){9-10}\cmidrule(lr){11-12}
Model & Crop & Res. & Ratio & Offline & Turning & Straight & Ratio & Turning & Straight & Deviation & \begin{tabular}[b]{@{}r@{}}Offline\\crop\end{tabular} \\
\midrule
DriveVA & \vr{v3.driveva.crop} & \vr{v3.driveva.res} & \cellcolor{passgreen}\vr{v3.driveva.kappa} & 2.8 & 2.18 & 1.63 & \cellcolor{passgreen}1.34 & 0.97 / 2.60 & 0.73 / 1.67 & 1.64 & 0.91 \\
ReCogDrive-2B & \vr{v3.recogdrive.crop} & \vr{v3.recogdrive.res} & \cellcolor{passgreen}\vr{v3.recogdrive.kappa} & 2.2 & 0.88 & 0.78 & \cellcolor{passgreen}1.13 & 0.84 / 0.56 & 0.58 / 0.75 & 0.78 & 0.59 \\
DriveVLA-W0 & \vr{v3.drivevlaw0.crop} & \vr{v3.drivevlaw0.res} & \cellcolor{passgreen}\vr{v3.drivevlaw0.kappa} & 2.2 & 5.40 & 3.04 & \cellcolor{passgreen}1.78 & 1.93 / 4.19 & 2.07 / 2.22 & 3.09 & 1.71 \\
WAM-Flow & \vr{v3.wamflow.crop} & \vr{v3.wamflow.res} & \cellcolor{passgreen}\vr{v3.wamflow.kappa} & 5.7 & 0.92 & 0.79 & \cellcolor{passgreen}1.17 & 1.00 / 1.28 & 0.59 / 1.09 & 0.79 & 0.54 \\
Alpamayo-1.5 & \vr{v3.alpamayo.crop} & \vr{v3.alpamayo.res} & \cellcolor{passgreen}\vr{v3.alpamayo.kappa} & -- & 1.47 & 0.70 & \cellcolor{passgreen}2.11 & 1.36 / 1.87 & 1.21 / 1.29 & 0.72 & -- \\
\bottomrule
\end{tabular*}
\end{table}

\paragraph{Summary.} On vehicle frames a field-of-view crop again moved plans $2.4$ to $6.4$ times more than a resolution cut at the same pixel budget, at or above each offline value. On the $22$ turning windows of the hand-driven lap, the plans all five models made from the factory view departed from those made from the virtual view more on turns than on straight road ($1.13$ to $2.11$ times).

\FloatBarrier

\subsection{Field experiment IV: the latency gate in closed loop}
\label{app:veh-v4}

\paragraph{Setup.} Field experiment IV drives DriveVA, \trbnew{the highest-scoring candidate, which meets the $1.2$\,s deadline on inference time and overruns it on the end-to-end latency,} on the test loop under two controllers fed identical plans. \emph{Plan tracking}, the vehicle's existing controller, tracks the newest plan with the current pose at every $10$\,ms control step; \emph{command hold}, the same controller changed in one place, computes speed and curvature once per plan and holds them until the next, as in RQ4. Each controller drove two runs, plan tracking first, with speed capped at $1.5$\,m/s ($5.4$\,km/h). Plan age is the time from the plan's newest frame, as received, to the control step that uses the plan; a deadline miss is an engaged control step on which no new plan has arrived for more than $1.2$\,s. Takeovers are counted after the first stretch of each run, which was spent adjusting the equipment.

\begin{figure}[!ht]
\centering
\includegraphics[width=\linewidth]{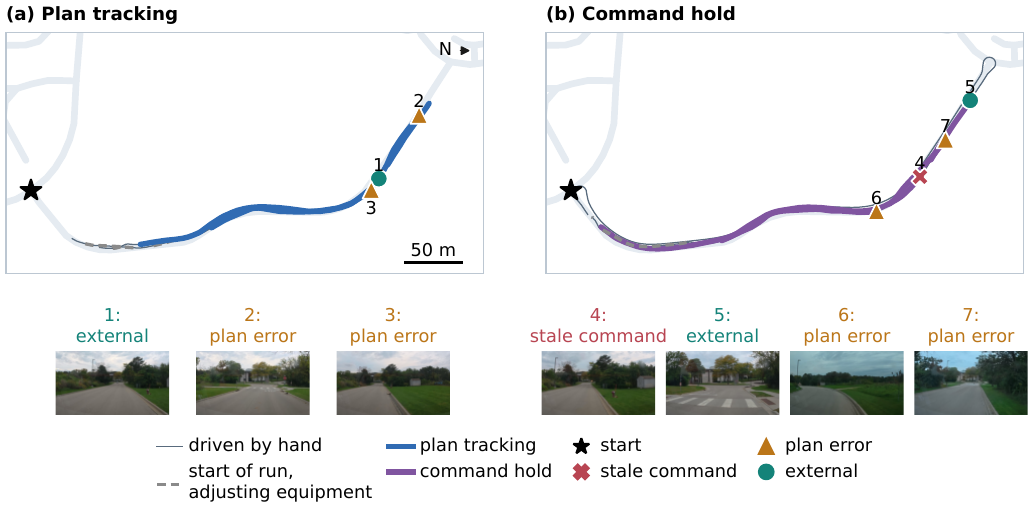}
\caption{\textbf{Field experiment IV: closed-loop runs and takeovers.} Top: track of each controller's two runs, north to the right, by hand (gray), during equipment adjustment (dashed) and under control (color), with takeovers numbered as in \cref{tab:v4-takeovers}. Bottom: the image the model received at each takeover; two building signs are blurred.}
\label{fig:vehicle-v4-map}
\end{figure}

\paragraph{Both controllers miss the deadline in a fifth of control steps.} The inference process took $1.03$\,s per plan at the median in every run \trbnew{(\TrbInferPninetyfiveLo{} to \TrbInferPninetyfiveHi\,s at the 95th percentile)}, and plans arrived every $1.5$\,s at the median. Plan tracking was engaged for $6.1$\,min over $511$\,m with three takeovers, command hold for $7.4$\,min over $595$\,m with four (\cref{fig:vehicle-v4-map}, \cref{tab:v4-takeovers}). Both controllers consumed the same plans: $20\%$ of engaged control steps missed the deadline under each, and the image behind the plan in effect was $2.26$\,s old at the median ($2.93$\,s at the 95th percentile).

{\trbnewon\paragraph{End to end, DriveVA fails the deadline.} Each plan's end-to-end latency can be split ($\TrbNPlans$ plans received while engaged, \TrbNRuns{} runs). From the camera read of the newest frame, \TrbPreInferPfifty\,s passed before inference started (reprojection to the virtual view, JPEG encoding and the hand-off to the inference loop), inference took \TrbInferPfifty\,s, decoding, preprocessing, resampling and publishing \TrbPostInferPfifty\,s, and the plan was recorded on the trajectory topic the controller reads \TrbTransportPfifty\,ms after the process finished, all at the median. The end-to-end latency was \TrbCyclePfifty\,s at the median and \TrbCyclePninetyfiveLo{} to \TrbCyclePninetyfiveHi\,s at the 95th percentile per run, and no plan took less than \TrbCycleMin\,s. Because the inference process starts a plan only on a frame sampled after the previous inference ended, and samples every \TrbSamplerPeriod\,s, consecutive plans were \TrbTicksPerCycle{} sampling periods apart in \TrbTickShare\% of cases. DriveVA meets the deadline on median inference time with the vehicle parked (\cref{app:veh-v1}); on the 95th percentile of this end-to-end latency, which the gate of \cref{sec:gates} uses, it fails the $1.2$\,s deadline, consistent with the deadline misses above. \trbnew{\Cref{tab:e5-parked} gives the end-to-end latency of all six candidates with the vehicle parked.}\par}

\begin{table}[!ht]
\centering\small\setlength{\tabcolsep}{2pt}
\caption{\trbnew{\textbf{Three measures of latency for the candidates on the vehicle} (s). \emph{Lab}: median inference time on an idle RTX A6000 (\cref{tab:pool}). \emph{Onboard}: vehicle parked, with its full input path, recording and tracker running; the end-to-end latency runs from the exposure of the newest frame a plan used to the finished plan, \emph{older frame} from the oldest frame of the input history. Gate: the $1.2$\,s deadline on the median onboard inference time and on the 95th percentile of the end-to-end latency (\cref{sec:gates}); under driving load LTF's is \TrbEfiveLtfEteNewPninetyfive\,s (\cref{app:veh-nov}). Green: passes; red: fails.}}
\label{tab:e5-parked}
\fontsize{8}{9.5}\selectfont\begin{tabular*}{\linewidth}{@{\extracolsep{\fill}}llrrrrrrrrcc@{}}
\toprule
 & & & \multicolumn{2}{c}{\trbnew{Inference p50}} & \multicolumn{5}{c}{\trbnew{Onboard end-to-end latency}} & \multicolumn{2}{c}{\trbnew{$1.2$\,s gate on}}\\
\cmidrule(lr){4-5}\cmidrule(lr){6-10}\cmidrule(l){11-12}
\trbnew{Candidate} & \trbnew{Input} & \trbnew{PDMS} & \trbnew{Lab} & \trbnew{Onboard} & \trbnew{p50} & \trbnew{p95} & \trbnew{\begin{tabular}[b]{@{}r@{}}Over\\$1.2$\,s\end{tabular}} & \trbnew{\begin{tabular}[b]{@{}r@{}}Older frame\\p95\end{tabular}} & \trbnew{Plans} & \trbnew{inference} & \trbnew{end-to-end} \\
\midrule
\trbnew{LTF} & \trbnew{three views} & \TrbEfivePkLtfPdms & \TrbEfivePkLtfLab & \trbnew{\TrbEfivePkLtfInfer} & \trbnew{\TrbEfivePkLtfPfifty} & \trbnew{\TrbEfivePkLtfPninetyfive} & \trbnew{\TrbEfivePkLtfOver} & \trbnew{\TrbEfivePkLtfOld} & \trbnew{\TrbEfivePkLtfN} & \gatecell{\TrbEfivePkLtfGateInf} & \gatecellend{\TrbEfivePkLtfGate} \\
\trbnew{Latent-WAM} & \trbnew{three views} & \trbnew{\TrbEfivePkLwamPdms} & \trbnew{\TrbEfivePkLwamLab} & \trbnew{\TrbEfivePkLwamInfer} & \trbnew{\TrbEfivePkLwamPfifty} & \trbnew{\TrbEfivePkLwamPninetyfive} & \trbnew{\TrbEfivePkLwamOver} & \trbnew{\TrbEfivePkLwamOld} & \trbnew{\TrbEfivePkLwamN} & \gatecell{\TrbEfivePkLwamGateInf} & \gatecellend{\TrbEfivePkLwamGate} \\
\trbnew{DriveVLA-W0} & \trbnew{front} & \TrbEfivePkWzPdms & \TrbEfivePkWzLab & \trbnew{\TrbEfivePkWzInfer} & \trbnew{\TrbEfivePkWzPfifty} & \trbnew{\TrbEfivePkWzPninetyfive} & \trbnew{\TrbEfivePkWzOver} & \trbnew{\TrbEfivePkWzOld} & \trbnew{\TrbEfivePkWzN} & \gatecell{\TrbEfivePkWzGateInf} & \gatecellend{\TrbEfivePkWzGate} \\
\trbnew{ReCogDrive} & \trbnew{front} & \TrbEfivePkRcPdms & \TrbEfivePkRcLab & \trbnew{\TrbEfivePkRcInfer} & \trbnew{\TrbEfivePkRcPfifty} & \trbnew{\TrbEfivePkRcPninetyfive} & \trbnew{\TrbEfivePkRcOver} & \trbnew{\TrbEfivePkRcOld} & \trbnew{\TrbEfivePkRcN} & \gatecell{\TrbEfivePkRcGateInf} & \gatecellend{\TrbEfivePkRcGate} \\
\trbnew{DriveVA} & \trbnew{front} & \TrbEfivePkDvaPdms & \TrbEfivePkDvaLab & \trbnew{\TrbEfivePkDvaInfer} & \trbnew{\TrbEfivePkDvaPfifty} & \trbnew{\TrbEfivePkDvaPninetyfive} & \trbnew{\TrbEfivePkDvaOver} & \trbnew{\TrbEfivePkDvaOld} & \trbnew{\TrbEfivePkDvaN} & \gatecell{\TrbEfivePkDvaGateInf} & \gatecellend{\TrbEfivePkDvaGate} \\
\trbnew{Alpamayo-1.5} & \trbnew{front} & \TrbEfivePkAlpPdms & \TrbEfivePkAlpLab & \trbnew{\TrbEfivePkAlpInfer} & \trbnew{\TrbEfivePkAlpPfifty} & \trbnew{\TrbEfivePkAlpPninetyfive} & \trbnew{\TrbEfivePkAlpOver} & \trbnew{\TrbEfivePkAlpOld} & \trbnew{\TrbEfivePkAlpN} & \gatecell{\TrbEfivePkAlpGateInf} & \gatecellend{\TrbEfivePkAlpGate} \\
\bottomrule
\end{tabular*}
\end{table}

{\trbnewon\paragraph{Inference time understates the end-to-end latency.} \Cref{tab:e5-parked} sets three measures of each on-vehicle candidate's latency side by side. The laboratory and onboard inference medians cover the model call alone; the end-to-end latency adds the input path before inference and the publishing after it, so inference time is a lower bound on the end-to-end latency: it can exclude a candidate but cannot admit one (\cref{sec:gates}). Every candidate with a laboratory reading meets the $1.2$\,s deadline on it, and on onboard inference time every candidate except Alpamayo-1.5 does, DriveVA and ReCogDrive among them; these two have the highest PDMS of the candidates. Measured as the end-to-end latency, they fail: ReCogDrive misses the deadline by \TrbEfivePkRcMarginMs\,ms, and DriveVA overruns it with every plan, as does Alpamayo-1.5. The other \TrbEfivePkNPass{} candidates pass, LTF, Latent-WAM \citep{wang2026latentwam} and DriveVLA-W0, and the ranking of \cref{app:veh-nov-select} therefore selects DriveVLA-W0, with LTF as challenger. The end-to-end latency is measured from the newest input frame to the finished plan (\cref{sec:gates}); from the older of DriveVLA-W0's \TrbEfivePkWzOldFrames{} input frames it is \TrbEfivePkWzOld\,s at the 95th percentile.\par}

\begin{figure}[!ht]
\centering
\includegraphics[width=\linewidth]{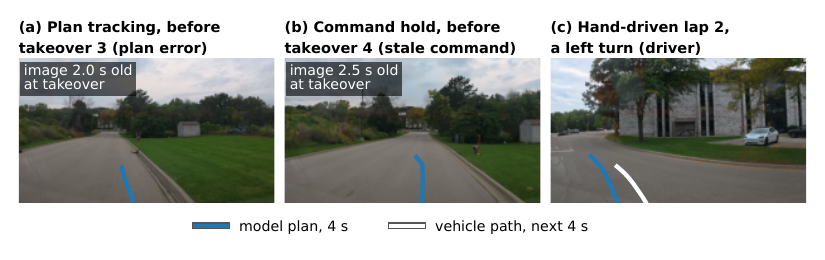}
\caption{\textbf{Field experiment IV: plans in the image.} Plan in effect (blue) before takeovers 3 and 4 (a, b) and on a hand-driven left turn (c), with the path driven over the next $4$\,s (white), projected on flat ground. Before the takeovers the vehicle moved less than $5$\,m, all below the image. Labels: age of the plan's image.}
\label{fig:vehicle-plan-overlay}
\end{figure}

\paragraph{One takeover is attributed to a stale command.} A rater blind to the controller, the run and the clock time classified each takeover (\cref{tab:v4-takeovers}): one as stale-command, under command hold (\cref{fig:vehicle-plan-overlay}b), four as plan errors, two under each controller, and two as external.

\begin{table}[!ht]
\centering\small
\caption{{\textbf{Field experiment IV: every takeover while engaged}, in time order, with the blinded rating. \emph{By}: how the driver took over; \emph{engaged before}: length of the engaged segment; \emph{plan age}: age of the image behind the plan; \emph{offset}: distance from the newest plan's path, left positive.}}
\label{tab:v4-takeovers}
\begin{tabular*}{\linewidth}{@{\extracolsep{\fill}}rrlrlrrrl@{}}
\toprule
No. & Run & Controller & \begin{tabular}[b]{@{}r@{}}Engaged\\before (s)\end{tabular} & By & \begin{tabular}[b]{@{}r@{}}Speed\\(m/s)\end{tabular} & \begin{tabular}[b]{@{}r@{}}Plan\\age (s)\end{tabular} & \begin{tabular}[b]{@{}r@{}}Offset\\(m)\end{tabular} & Rating \\
\midrule
1 & 1 & plan tracking & 108 & brake & 1.37 & 2.62 & -0.36 & external \\
2 & 1 & plan tracking & 39 & brake, steering & 1.45 & 1.63 & -0.29 & plan error \\
3 & 2 & plan tracking & 176 & steering & 0.33 & 1.96 & -0.11 & plan error \\
4 & 3 & command hold & 238 & brake & 1.57 & 2.45 & -0.35 & stale command \\
5 & 3 & command hold & 43 & brake & 1.01 & 2.38 & +0.75 & external \\
6 & 4 & command hold & 103 & steering & 1.60 & 1.65 & -1.08 & plan error \\
7 & 4 & command hold & 62 & steering & 0.65 & 2.47 & -2.23 & plan error \\
\bottomrule
\end{tabular*}
\end{table}

\FloatBarrier

\paragraph{Summary.} The rater attributed none of the three takeovers under plan tracking and one of the four under command hold to stale commands, the direction RQ4 found in simulation.

\FloatBarrier

{\trbnewon
\subsection{The three-camera configuration: cameras, routes and procedure}
\label{app:veh-nov}

\begin{figure}[!ht]
\centering
\includegraphics[width=\linewidth]{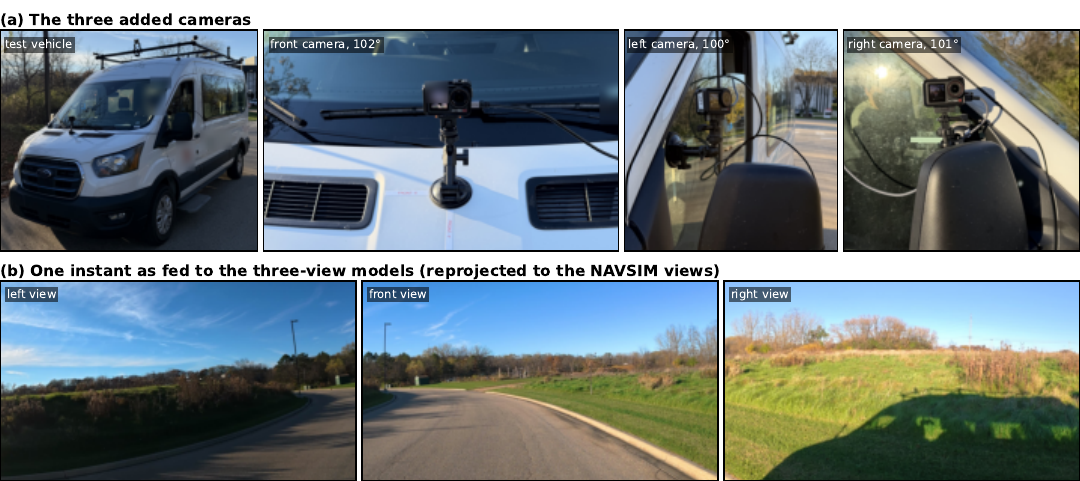}
\caption{\trbnew{\textbf{The three added cameras.} (a)~The test vehicle with the front, left and right cameras, each recording at $10$\,Hz. (b)~One instant of a hand-driven trip as the three-view candidates receive it, each camera reprojected into the corresponding NAVSIM view. People, signs and vehicle markings are blurred.}}
\label{fig:vehicle-rig}
\end{figure}

\paragraph{Cameras and vehicle.} Field experiment~V used the vehicle of \cref{app:veh-platform} with three added action cameras, one on the hood and one on each side mirror (\cref{fig:vehicle-rig}a), recorded at $10$\,Hz and $1920\times1080$; their horizontal fields of view are $102^\circ$ (front), $100^\circ$ (left) and $101^\circ$ (right). The factory cameras were not used. Each camera is reprojected into the matching NAVSIM view (\cref{fig:vehicle-rig}b), so the three-view candidates (LTF, Latent-WAM) receive all three views and the single-view candidates the front one. Localization came from the inertial navigation system with an RTK-fixed solution throughout, and every model ran on the onboard RTX A6000.

\paragraph{Schedule, place and conditions.} The safety driver drove \TrbEfiveLtfNTrips{} trips by hand, \TrbNovRecMin{} minutes and \TrbNovRecKm\,km in all, on the business-park and residential roads of \cref{app:veh-platform}: a loop, driven twice in one trip, two further legs, one of them on streets that no other trip covers, and one more lap of the loop. LTF ran in shadow on every trip, which gives its end-to-end latency under driving load and the recordings of \cref{app:veh-nov-select,app:veh-stage5}. With the three added cameras, LTF running in shadow on \TrbEfiveLtfNTrips{} hand-driven trips (\TrbEfiveLtfNPlans{} plans) took \TrbEfiveLtfEtePfifty\,s at the median and \TrbEfiveLtfEtePninetyfive\,s at the 95th percentile from the estimated exposure of the oldest of its three frames to the finished plan (\TrbEfiveLtfEteNewPninetyfive\,s at the 95th percentile from the newest), with inference only \TrbEfiveLtfInferPfifty\,s: \TrbEfiveLtfPrePfifty\,s passed between the exposure and the start of inference, \TrbEfiveLtfDeliveryPfifty\,s of them until the last of the three frames arrived (a side camera's), and plans followed each other every \TrbEfiveLtfGapPfifty\,s at the median, with \TrbEfiveLtfNOver{} plans over $1.2$\,s, so it passes the latency gate under driving load. With the vehicle parked, each of the six candidates ran one shadow trip for the latency comparison of \cref{tab:e5-parked}. The closed-loop trips of Field experiment~V (\cref{app:veh-e6}) were driven on a short access road.

\paragraph{Software and controller.} The candidates ran inside a copy of the vehicle's test software, extended with three settings that the operator chooses per trip: an added delay between the finished plan and its publication ($0$, $0.5$, $1.0$ or $2.0$\,s, the plan's time stamp unchanged), blacking out the two side views, and the checkpoint (original or fine-tuned). The operator's screen showed only model labels, so the safety driver did not know the configuration of a trip. The controller tracks each plan from the time of its newest frame; a plan covers \TrbNovHorizon\,s, after which the controller holds its last command, and it stops the vehicle if no new plan has arrived \TrbNovThold\,s later. Speed was capped at \TrbNovVmax\,m/s.

\paragraph{Procedure and data.} Each closed-loop trip started from rest at the start of the route with an RTK-fixed position. The safety driver stayed at the controls with a foot over the brake, and any brake, throttle or steering input ended the trip as a takeover, except braking within the last $5$\,m to stop at the end of the route, which has no stop line. The order of LTF's trips was drawn at random. For every trip the recording keeps the three camera streams, the inertial navigation solution at $100$\,Hz, the drive-by-wire reports, every plan with its time stamps and the controller's log.
\par}

\FloatBarrier

{\trbnewon
\subsection{Stages~3 and~4 with the three-camera configuration}
\label{app:veh-nov-select}\par}

\begin{table}[!ht]
\centering\small\setlength{\tabcolsep}{2pt}
\caption{\textbf{Probe readings on the local recordings} \trbnew{\textbf{and the Stage~4 ranking.} Upper block: Field experiment~II, added front camera, $2842$ windows. Lower block: the three-camera configuration, \TrbEsthreeNWin{} windows per candidate. TI: temporal instability (m) with its 95\% interval; $A_\gamma$, $A_\mathrm{jpeg}$: darkening and compression responses; gray: displacement under a gray frame (m); sign: agreement of the planned turn direction with the driver's; gate: the latency gate of \cref{tab:e5-parked}. Rank: by PDMS among candidates passing both gates and the sanity checks. Green: passes; red: fails.}}
\label{tab:e5-stage3}
\begin{tabular*}{\linewidth}{@{\extracolsep{\fill}}lccccccccc@{}}
\toprule
\trbnew{Candidate} & \trbnew{TI (95\% interval)} & \trbnew{Four-city TI} & \trbnew{$A_\gamma$} & \trbnew{$A_\mathrm{jpeg}$} & \trbnew{Gray} & \trbnew{Sign} & \trbnew{Gate} & \trbnew{PDMS} & \trbnew{Rank} \\
\midrule
\multicolumn{10}{@{}l}{\emph{Field experiment~II}}\\
DriveVA & \vr{v2.driveva.ti} \trbnew{(\TrbRrTiDva)} & \trbnew{0.37--0.51} & \vr{v2.driveva.ag} & \vr{v2.driveva.aj} & \vr{v2.driveva.dgray} & \trbnew{0.968} & -- & -- & -- \\
ReCogDrive & \vr{v2.recogdrive.ti} \trbnew{(\TrbRrTiRc)} & \trbnew{0.63--0.82} & \vr{v2.recogdrive.ag} & \vr{v2.recogdrive.aj} & \vr{v2.recogdrive.dgray} & \trbnew{0.974} & -- & -- & -- \\
DriveVLA-W0 & \vr{v2.drivevlaw0.ti} \trbnew{(\TrbRrTiWz)} & \trbnew{1.08--1.28} & \vr{v2.drivevlaw0.ag} & \vr{v2.drivevlaw0.aj} & \vr{v2.drivevlaw0.dgray} & -- & -- & -- & -- \\
WAM-Flow & \vr{v2.wamflow.ti} \trbnew{(\TrbRrTiWf)} & \trbnew{0.45--0.55} & \vr{v2.wamflow.ag} & \vr{v2.wamflow.aj} & \vr{v2.wamflow.dgray} & -- & -- & -- & -- \\
\midrule
\multicolumn{10}{@{}l}{\trbnew{\emph{Three-camera configuration}}}\\
\trbnew{DriveVLA-W0} & \trbnew{\TrbEsthreeWzTi{} (\TrbEsthreeWzTiLo, \TrbEsthreeWzTiHi)} & \TrbEsthreeWzRg & \trbnew{\TrbEsthreeWzAg} & \trbnew{\TrbEsthreeWzAj} & \trbnew{\TrbEsthreeWzGray} & \trbnew{\TrbEsthreeWzSign} & \gatecell{\TrbEsthreeWzGate} & \TrbEsthreeWzPdms & \trbnew{\TrbEsthreeWzRank} \\
\trbnew{LTF} & \trbnew{\TrbEsthreeLtfTi{} (\TrbEsthreeLtfTiLo, \TrbEsthreeLtfTiHi)} & \trbnew{\TrbEsthreeLtfRg} & \trbnew{\TrbEsthreeLtfAg} & \trbnew{\TrbEsthreeLtfAj} & \trbnew{\TrbEsthreeLtfGray} & \trbnew{\TrbEsthreeLtfSign} & \gatecell{\TrbEsthreeLtfGate} & \TrbEsthreeLtfPdms & \trbnew{\TrbEsthreeLtfRank} \\
\trbnew{Latent-WAM} & \trbnew{\TrbEsthreeLwamTi{} (\TrbEsthreeLwamTiLo, \TrbEsthreeLwamTiHi)} & \trbnew{\TrbEsthreeLwamRg} & \trbnew{\TrbEsthreeLwamAg} & \trbnew{\TrbEsthreeLwamAj} & \trbnew{\TrbEsthreeLwamGray} & \trbnew{\TrbEsthreeLwamSign} & \gatecell{\TrbEsthreeLwamGate} & \trbnew{\TrbEsthreeLwamPdms} & \trbnew{\TrbEsthreeLwamRank} \\
\trbnew{ReCogDrive} & \trbnew{\TrbEsthreeRcTi{} (\TrbEsthreeRcTiLo, \TrbEsthreeRcTiHi)} & \TrbEsthreeRcRg & \trbnew{\TrbEsthreeRcAg} & \trbnew{\TrbEsthreeRcAj} & \trbnew{\TrbEsthreeRcGray} & \trbnew{\TrbEsthreeRcSign} & \gatecell{\TrbEsthreeRcGate} & \TrbEsthreeRcPdms & \trbnew{\TrbEsthreeRcRank} \\
\trbnew{DriveVA} & \trbnew{\TrbEsthreeDvaTi{} (\TrbEsthreeDvaTiLo, \TrbEsthreeDvaTiHi)} & \TrbEsthreeDvaRg & \trbnew{\TrbEsthreeDvaAg} & \trbnew{\TrbEsthreeDvaAj} & \trbnew{\TrbEsthreeDvaGray} & \trbnew{\TrbEsthreeDvaSign} & \gatecell{\TrbEsthreeDvaGate} & \TrbEsthreeDvaPdms & \trbnew{\TrbEsthreeDvaRank} \\
\bottomrule
\end{tabular*}
\end{table}

{\trbnewon\paragraph{Setup.} We ran Stage~3 on the \TrbEsthreeNSamples{} usable samples ($2$\,Hz, about \TrbEsthreeMin{} minutes) of the four hand-driven trips of \cref{app:veh-nov}, which give \TrbEsthreeNWin{} windows of four consecutive samples, against \TrbEsthreeNWinSept{} windows in Field experiment~II, on the vehicle's own RTX A6000, with the turn signal as the command. The Stage~4 rule is as follows. Since in Field experiment~II no candidate's readings lay within its four-city ranges (\cref{app:veh-v2}), the range is reported but does not exclude; candidates that pass the sensing gate, the latency gate and the sanity checks are ranked by global PDMS, the first being the selected model and the second the challenger, and local closed-loop acceptance decides.

\paragraph{DriveVLA-W0 is selected, with LTF as challenger.} Every candidate passes the sanity checks: the blind anchor responds exactly zero, every sighted candidate's gray response exceeds twice its run-to-run noise, and planned turn directions agree with the driver's in \TrbEsthreeSignMin{} to \TrbEsthreeSignMax{} of turning frames (\cref{tab:e5-stage3}). The TI interval of each of the \TrbEsthreeNOutside{} candidates with a four-city range lies entirely above that range, so the range does not separate them. Among the candidates that pass both gates, DriveVLA-W0 (PDMS \TrbEsthreeWzPdms) is the selected model and LTF (\TrbEsthreeLtfPdms) the challenger; Latent-WAM passes both gates and the sanity checks but has no score in the pool, and ReCogDrive and DriveVA, the two highest-scoring candidates, fail the latency gate. Whether the selected model or the challenger is accepted is decided by the closed-loop trips of Field experiment~V (\cref{app:veh-e6}).\par}

\FloatBarrier

{\trbnewon
\subsection{Stage~5 offline: fine-tuning on the local recordings}
\label{app:veh-stage5}

We applied Stage~5 to LTF, the challenger of \cref{app:veh-nov-select}, which completed every closed-loop trip (\cref{app:veh-e6}), so that a closed-loop change after fine-tuning can be attributed to the fine-tuning. Its monitoring step ran offline on the four hand-driven trips of \cref{app:veh-nov}; Field experiment~V then applies Stage~5's closed-loop check to the fine-tuned checkpoint. One trip, the only one on streets that no other trip covers, is held out (\TrbEsevenNHeld{} samples at $2$\,Hz, \TrbEsevenNTurn{} of them in turns); the other three give \TrbEsevenNTrain{} training samples, reprojected from the three added cameras into the nuPlan front views, with the straight-ahead command the vehicle feeds. From the checkpoint the vehicle runs, the official LTF agent was fine-tuned on its trajectory loss alone (the recordings have no agent or map labels) for five epochs with \TrbEsevenNSeeds{} seeds, about \TrbEsevenTrainMin{} minutes each on one GPU. Fine-tuning counts as an improvement only if, for every seed, the 95\% interval of the change in average displacement error on the held-out trip lies below zero (block bootstrap over consecutive $10$\,s blocks); a monitor flags the dose if its sensitivity to darkening or compression exceeds $1.5$ times the zero-shot level, or if its response to a blank image falls below half of it.

Fine-tuning reduced the held-out average displacement error from \TrbEsevenAdeZs\,m to \TrbEsevenAdeFtLo{} to \TrbEsevenAdeFtHi\,m, a change of \TrbEsevenDeltaLo{} to \TrbEsevenDeltaHi\,m per seed; every interval ends at or below \TrbEsevenDeltaUpMax\,m, and the seeds differ by \TrbEsevenSpread\,m. The change holds on the \TrbEsevenNNovel{} held-out samples more than $30$\,m from any training position (\TrbEsevenNovelDeltaLo{} to \TrbEsevenNovelDeltaHi\,m, intervals ending at or below \TrbEsevenNovelUpMax\,m) and in turns (\TrbEsevenTurnDeltaLo{} to \TrbEsevenTurnDeltaHi\,m, although the interval of \TrbEsevenTurnNIncl{} seed includes zero). Neither monitor fired. Sensitivity to darkening and to compression fell to \TrbEsevenGammaRatio{} and \TrbEsevenJpegRatio{} times its zero-shot level, below the level at which \cref{alg:triage} withholds a dose, and the response to a blank image stayed at \TrbEsevenBlindRatio{} times, so the fine-tuned model still uses the images. The closed-loop check that Stage~5 requires before release, run on the vehicle afterwards, withheld the fine-tuned model (\cref{app:veh-e6}).\par}

\FloatBarrier

{\trbnewon
\subsection{Field experiment V: the selection, both gates and Stage~5 in closed loop}
\label{app:veh-e6}

\begin{figure}[!ht]
\centering
\includegraphics[width=\linewidth]{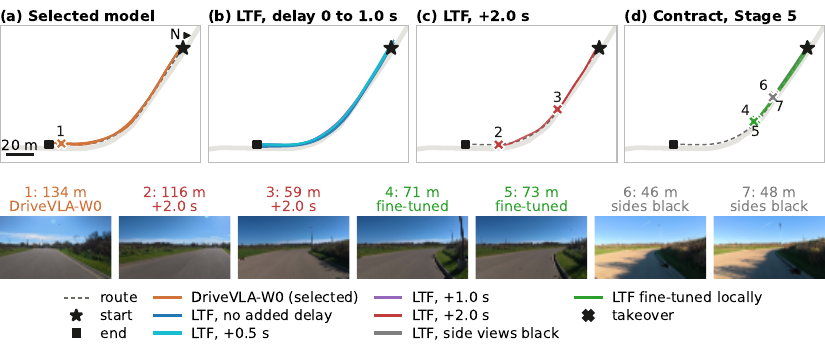}
\caption{\trbnew{\textbf{Field experiment V: closed-loop trips and takeovers.} Top: the route (dashed; start at the star, end at the square) and the autonomous stretch of every counted DriveVLA-W0 and LTF trip: (a)~the selected model, (b)~LTF with no added delay and with $0.5$ and $1.0$\,s, (c)~LTF with $2.0$\,s, (d)~LTF with its side views blacked out and LTF fine-tuned locally. Takeovers are numbered in order. Bottom: the front view at each takeover, with the distance along the route.}}
\label{fig:vehicle-e6-map}
\end{figure}

\paragraph{Setup.} We drove the rule of \cref{app:veh-nov-select} on a local route of \TrbEsixRouteLen\,m, about $65$\,m of straight road followed by a right turn, at up to $4.0$\,m/s, with the procedure of \cref{app:veh-nov}. The selected model is accepted if at least four of five trips reach the end without a takeover, and a test configuration of LTF is stopped after two consecutive takeovers. We ran three tests on LTF: the latency gate, by holding each finished plan for an extra $0.5$, $1.0$ or $2.0$\,s before publishing it, which leaves the plan's time stamp and therefore its true age at the controller unchanged; the sensing gate, by blacking out the two side views LTF's contract requires, admitted to closed loop only if, in one human-driven shadow trip, the median distance of its $4$\,s plan point from the driven path in the turn stayed within $1$\,m; and Stage~5, with the checkpoint fine-tuned on the local recordings (\cref{app:veh-stage5}). LTF's trips ran in one randomized block in which the safety driver saw only a model label; the $0.5$ and $1.0$\,s configurations ran three trips each. DriveVA, ReCogDrive and Latent-WAM were each driven \TrbEsixZyDvaN{} trips on the same route under the same speed cap, the first two on the front view and Latent-WAM on all three views. \Cref{tab:e6} gives the outcome of every configuration, and \cref{fig:vehicle-e6-map} shows where every trip drove and where each takeover happened. The DriveVLA-W0 and LTF trips, \TrbEsixNTrips{} in all, covered \TrbEsixDistKm\,km in \TrbEsixMin{} minutes of autonomous driving, with \TrbEsixNInt{} takeovers.

\paragraph{Stage~4 returns a model.} The selected model, DriveVLA-W0, reached the end in \TrbEsixWzOk{} of its \TrbEsixWzN{} trips; the safety driver took over once, at \TrbEsixWzIv\,m in the turn, for leaving the lane, so the model is accepted. The challenger LTF reached the end in all \TrbEsixZeroN{} trips. The safety driver noted that DriveVLA-W0 ran close to the road edge in its completed trips, while LTF kept a wider margin. The two highest-scoring candidates, DriveVA and ReCogDrive, fail the latency gate (\cref{tab:e5-parked}). On the same route DriveVA reached the end in \TrbEsixZyDvaOk{} of \TrbEsixZyDvaN{} trips and ReCogDrive in \TrbEsixZyRcOk{} of \TrbEsixZyRcN{}, with takeovers after \TrbEsixZyDvaIvMin{} to \TrbEsixZyDvaIvMax\,m and \TrbEsixZyRcIvMin{} to \TrbEsixZyRcIvMax\,m, while LTF and Latent-WAM reached the end in all their trips (\cref{tab:e6}).

\paragraph{The latency gate in closed loop.} Holding LTF's plans longer degraded driving step by step. LTF's own end-to-end latency is \TrbEfiveLtfEteNewPninetyfive\,s at the 95th percentile under driving load (\cref{app:veh-nov}), so every added delay puts it beyond the $1.2$\,s deadline. With $0.5$\,s added (plan age at the controller \TrbEsixHalfAge\,s at the median, against \TrbEsixZeroAge\,s undelayed) every trip was still completed and smooth. With $1.0$\,s (\TrbEsixOneAge\,s) every trip was completed, but plans arrived only every \TrbEsixOneGap\,s and the steering wheel moved in steps, at \TrbEsixOneRate\,deg/s at the 95th percentile against \TrbEsixZeroRate\,deg/s undelayed, which the safety driver felt as jerky. With $2.0$\,s (\TrbEsixTwoAge\,s) the controller spent \TrbEsixTwoHold\% of the time holding an expired plan, the steering wheel swung with growing amplitude up to \TrbEsixTwoSteerMax\,deg, and both trips ended in a takeover for leaving the lane, after \TrbEsixTwoIvA{} and \TrbEsixTwoIvB\,m. Driving thus degraded with the end-to-end latency, as the gate assumes, and failed between $1.0$ and $2.0$\,s of added delay; at $4$\,m/s on this route the $1.2$\,s deadline is conservative.

\paragraph{The sensing gate in closed loop.} In the shadow trip the blacked-out configuration's plans stayed within the admission threshold (median distance \TrbEsixGateSb\,m against \TrbEsixGateTv\,m with all three views; 90th percentiles \TrbEsixGatePninetySb{} and \TrbEsixGatePninetyTv\,m), so it was driven. Both closed-loop trips ended on the straight, after \TrbEsixSbIvA{} and \TrbEsixSbIvB\,m, with the vehicle drifting out of its lane, whereas LTF with its full contract completed all \TrbEsixZeroN. An open-loop comparison of the plans underestimated what violating the contract does in closed loop.

\paragraph{Stage~5's closed-loop check.} The fine-tuned LTF, which had lowered the held-out displacement error offline, drifted steadily to the right on the straight in both trips and was taken over on entering the turn, after \TrbEsixFtIvA{} and \TrbEsixFtIvB\,m, while the original checkpoint completed every trip. Under the rule of Stage~5 it is not released; the offline improvement did not carry over to closed loop.\par}

\FloatBarrier

\FloatBarrier
\section{Formal statements and proofs}
\label{app:theory}

\subsection{Setting}
This appendix states formally what \cref{sec:framework} states in words, with the same symbols: configurations $c=(\pol,\iota)$, candidates trained under $(S_i,T_i,D_i)$, the target $(S',T',D')$ with $N$ local frames, and the risk $R(c)$ of \cref{sec:problem}; lower is better. When $L$ is a sum of per-step losses, $R(c)$ is proportional to $\E_{x\sim d^{c}}[\ell(x,u_c(x))]$, with $d^{c}$ the normalized closed-loop state distribution. We write
$R^\ast(\cdot)$ for the least risk attainable by any policy under
the constraint in parentheses. \emph{Post-training} is the operator
\begin{equation}
\mathcal{T}_N:\ \pol_\theta \longmapsto \pol_{\theta'},
\qquad
\theta' = \mathcal{A}\bigl(\theta,\ \widehat{D}'_N\bigr),
\label{eq:posttrain}
\end{equation}
where $\mathcal{A}$ is any training algorithm initialized at $\theta$
whose data $\widehat{D}'_N$ are $N$ observations from the target sensors and their
labels; architecture and interface are unchanged by definition, and
altering either creates a new configuration.

The three terms of \cref{eq:decomp} are defined below as differences
of least attainable risks: $\Delta_{\mathrm{sense}}$ in
\cref{app:infogap}, $\Delta_{\mathrm{lat}}$ in \cref{app:timegap},
and $\Delta_{\mathrm{dom}}$ in \cref{app:distgap}. The one property
the framework uses is that $\mathcal{T}_N$ cannot reduce the first
two for any $N$ (Propositions~\ref{prop:info} and~\ref{prop:time}),
while only the third carries a coefficient that $N$ drives down.
They are listed, not summed.

\subsection{Sensing gaps}
\label{app:infogap}
Write $S_c$ for the observations the sensor contract of $c$ consumes
and $S'$ for those the platform provides, and suppose the platform's observations can be generated from the sensor contract's by a stochastic map
$M$, $S' = M\circ S_c$ (a lossy transformation: a cropped, lower-resolution
or missing view). The sensing term is
\begin{equation}
\Delta_{\mathrm{sense}}(c;S') \;=\; R^\ast(S') - R^\ast(S_c) \;\ge\; 0,
\label{eq:infoterm}
\end{equation}
the risk that no policy observing $S'$ can avoid relative to one
observing $S_c$. The operational gate $g_{\mathrm{sense}}$ in
\cref{eq:lexico-main} checks whether the configured inputs satisfy
the checkpoint's sensor contract.

\begin{proposition}[Post-training cannot undo a lossy transformation]
\label{prop:info}
For every policy $\pol'$ that is a function of observations from the target sensors, including any post-processing of $S'$ and any
$\mathcal{T}_N$-fine-tune of $\pol$ on data collected through $S'$,
$R(\pol') \ge R^\ast(S') \ge R^\ast(S_c)$. The second inequality is
strict if and only if no optimal decision rule under $S_c$ can be
implemented through $M$. Hence no post-training budget closes
$\Delta_{\mathrm{sense}}$; only changing the sensors or the requirement can.
\end{proposition}

\begin{proof}[Proof sketch]
Any rule $\phi'(a\mid o')$ for $S'$ is reproduced from $S_c$ by
\begin{equation}
\phi(a \mid o) \;=\; \int \phi'(a \mid o')\, M(do' \mid o),
\label{eq:garble}
\end{equation}
with identical risk in every decision problem (the easy direction of
Blackwell's comparison of experiments \citep{blackwell1953}), so the
risks attainable under $S'$ are a subset of those attainable under
$S_c$. Applying $M$ independently at every step extends this to
history-dependent policies and hence to closed-loop risk
\citep{lecam1964}. Equality holds exactly when some optimal rule under
$S_c$ can be written in the form of \cref{eq:garble}.
\end{proof}

\emph{Where the deficit sits.} In a one-step decision problem with
state prior $p$, let $\phi^\ast_c$ and $\phi^\ast_{S'}$ be optimal
rules under $S_c$ and $S'$. Then
\begin{equation}
\Delta_{\mathrm{sense}}(c;S') = \E_{x\sim p}\bigl[r(x)\bigr],
\qquad
r(x) = \E\bigl[\ell(x,\phi^\ast_{S'}(o'))\mid x\bigr]
- \E\bigl[\ell(x,\phi^\ast_c(o))\mid x\bigr],
\label{eq:statecond}
\end{equation}
and $r(x)=0$ on states where the two rules take the same action;
$r(x)$ may be negative on some states, but its average is not.
In the sequential case the two optimal policies visit different
states, so \cref{eq:statecond} locates the deficit rather than
decomposing it; the experiments read it as a prediction about which
states carry the residual (turns, for a narrowed view), not as a
quantity to be summed.

\begin{figure}[!ht]
\centering
\includegraphics[width=0.99\linewidth]{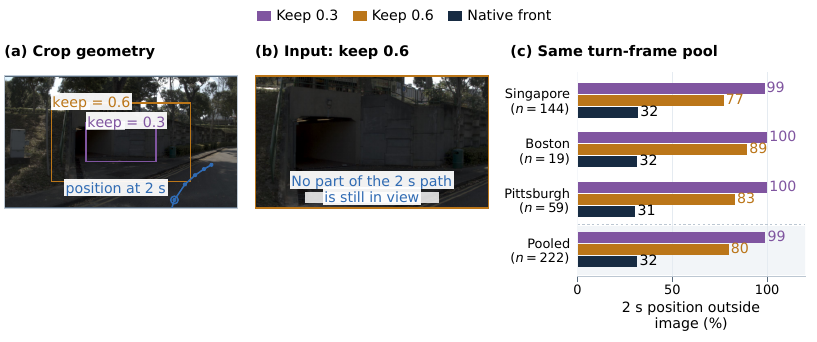}
\caption{\textbf{On turns, the crop removes the evidence the decision needs.} \textbf{(a)}~A turning frame with the logged $2$\,s path and both crop boundaries. \textbf{(b)}~The frame as delivered at \texttt{keep}~$=0.6$. \textbf{(c)}~Share of the $222$ turning frames whose $2$\,s position leaves the image.}
\label{fig:fovgeometry}
\end{figure}

This is visible directly in the data: on turning frames the position
the vehicle occupies two seconds later leaves a moderately cropped
image $79.7\%$ of the time (\cref{fig:fovgeometry}c; one such frame
before and after the crop in a and~b).

Two refinements are used in the main text. The criterion is
measured-\emph{or-inferable}: statistically redundant information,
such as monocular depth, is recoverable by data
\citep{kwon2025memdistill}, which does not contradict
\cref{prop:info}, since the proposition only bars recovering what the
observations never carried. And a policy with temporal memory may
recover in one frame what another frame lost, which shrinks the
practical deficit; the deficit must therefore be established by
experiment, not by comparing sensor lists.

\subsection{Latency gaps}
\label{app:timegap}
The control period $\delta$ is the interval at which the controller
needs a new command. The deadline $d'$ of the gate is set by the
controller and fixed before any \trbnew{$\hat\tau_c$} is measured: $d'=\delta$ when
a missed control step replays the last command, and $d'$ is the replanning
deadline when the controller tracks the last plan with fresh state.

\trbnew{$\hat\tau_c$ is the 95th percentile of the end-to-end latency of one plan,
from the acquisition of its newest frame to the plan being available to
the controller, measured on the deployment hardware in its operating
state; it includes preprocessing, inference, postprocessing and
transport. Inference time alone, timed on an idle GPU as in
\cref{sec:exp-time}, is a lower bound that can exclude a candidate but
not admit one. The analysis below treats the latency of $c$ as a fixed $\tau_c$.} Write
$R(c;\tau,\varsigma)$ for the risk of $c$ when its outputs arrive
with latency $\tau$ and are consumed under rule $\varsigma$, and
$R^\ast(S';\tau,\varsigma)$ for the least such risk over all policies
observing $S'$. The latency term is
\begin{equation}
\Delta_{\mathrm{lat}}(c;\tau_c,\varsigma) \;=\;
R^\ast(S';\tau_c,\varsigma) - R^\ast(S';0,\varsigma) \;\ge\; 0.
\label{eq:timeterm}
\end{equation}

\begin{proposition}[Latency-gap structure]
\label{prop:time}
With $k=\lceil\tau_c/\delta\rceil$ pending outputs, the closed loop is
the delayed decision process
\begin{equation}
\tilde{x}_t = \bigl(x_t,\ \xi_{t-1}, \dots, \xi_{t-k}\bigr),
\qquad
u_t = \varsigma\bigl(\xi_{t-k},\ p_t\bigr),
\label{eq:augstate}
\end{equation}
with $\xi$ the model outputs, $p_t$ fresh proprioception and $\varsigma$ the rule by which the loop consumes stale outputs
\citep{katsikopoulos2003delay,ramstedt2019realtime}. (i)~$R(c;\tau,
\varsigma)$ is a property of the loop, not of $\pol$ alone.
(ii)~$\mathcal{T}_N$ updates $\theta$ but neither $k$, which
architecture and compute fix, nor $\varsigma$, so every post-trained
policy still computes $\xi_{t-k}$ from observations at least $k$ control steps old, and for every $N$
\begin{equation}
R(\mathcal{T}_N c;\tau_c,\varsigma)\;\ge\;R^\ast(S';\tau_c,\varsigma)\;\ge\;R^\ast(S';0,\varsigma).
\label{eq:delaybound}
\end{equation}
The second inequality is \cref{prop:info} applied to the delayed
history, a lossy transformation of the fresh one, and is strict whenever the
optimal action depends on disturbances realized within the delay
window. Post-training can therefore recover the part of $\pol$'s own
excess risk that is predictable from delayed observations, but not
$\Delta_{\mathrm{lat}}$, and it does not change the failure mode
that $k$ and $\varsigma$ determine. (iii)~Under command hold,
$\varsigma_{\mathrm{cmd}}(\xi_{t-k},p_t)=u(\xi_{t-k})$ ignores $p_t$ and
the error grows open loop over the delay. Under plan tracking,
$\varsigma_{\mathrm{plan}}(\xi_{t-k},p_t)=\Gamma(\xi_{t-k},p_t)$ with a
feedback tracker $\Gamma$; if the tracked loop is contracting, then
\begin{equation}
R(c;\tau_c,\varsigma_{\mathrm{plan}}) - R(c;0,\varsigma_{\mathrm{plan}})
\;\le\; K \cdot
\E\bigl[\, \lVert \xi_{t} - \xi_{t-k} \rVert_{\infty} \bigr],
\label{eq:plandrift}
\end{equation}
with $K$ a constant that absorbs the loss's Lipschitz constant and
the loop's contraction rate.
\end{proposition}

Two remarks on (ii) and (iii). The recoverable part in (ii) is real:
if the state drifts at a known constant rate, learning to extrapolate
removes the delay's error entirely, which is why the latency term is
defined through the best delayed policy rather than through $\pol$.
In (iii), under command hold a double integrator whose held
acceleration command is off by $\Delta a$ is displaced by
$\tfrac12\lVert\Delta a\rVert(k\delta)^2$ after the delay; without a
contracting tracker, errors compound across steps and
\cref{eq:plandrift} does not hold. The drift on the right of
\cref{eq:plandrift} is measurable on recorded plans without running anything: on the turning frames of the model being fine-tuned it is
$1.36$\,m at a staleness of $500$\,ms (maximum over the eight plan
waypoints of the per-waypoint Euclidean distance, frame-weighted over
$144$, $19$ and $59$ turning frames in the three cities that have
them). The repairs for this class are therefore a faster compute tier or an architecture on the consuming side; a wrapper
changes $\varsigma$ only if the downstream consumer accepts stale
plans (\cref{fig:timegate}d shows one failing where the consumer rejects them).
Delay-aware retraining \citep{chen2020delayaware} changes the
policy's input interface, so in this taxonomy it is an architectural repair, not post-training \cref{eq:posttrain}.

\subsection{Distribution gaps}
\label{app:distgap}
This is the term $\Delta_{\mathrm{dom}}(c;D',N)$ of \cref{eq:decomp}:
appearance, layout and behavior shift between the training domain $D$
and the target $D'$, covariate shift and, where local behavior
differs, concept shift. We use the domain-adaptation bounds of \citet{bendavid2010theory} as a structural analogy. For a hypothesis $\pol$ trained on the source alone,
\begin{equation}
R_{D'}(\pol) \;\le\; R_{D}(\pol) \;+\;
\tfrac{1}{2}\, d_{\mathcal{H}\Delta\mathcal{H}}(D, D') \;+\;
\lambda,
\label{eq:bendavid}
\end{equation}
with $\lambda$ the combined error of the best hypothesis on both
domains, which absorbs concept shift. Post-training does not shrink
$d_{\mathcal{H}\Delta\mathcal{H}}(D,D')$, a property of the two
domains; it changes what the model is trained on. The relevant
statement is the paper's Theorem~3: if $\hat{\pol}$ minimizes an
$\alpha$-weighted empirical risk over $m$ source and $N$ target
\trbnew{labeled samples, and} $\pol^\ast_{D'}$ is the best hypothesis on the
target, then with high probability
\begin{equation}
R_{D'}(\hat{\pol}) \;\le\; R_{D'}(\pol^\ast_{D'})
\;+\; \epsilon_{\mathrm{est}}(\alpha,N,m)
\;+\; 2(1-\alpha)\bigl(\tfrac12 d_{\mathcal{H}\Delta\mathcal{H}}(D,D') + \lambda\bigr),
\label{eq:bendavid3}
\end{equation}
where the estimation term $\epsilon_{\mathrm{est}}$ shrinks as $N$
grows and grows with the weight $\alpha$ placed on the target. As $N$
grows the best $\alpha$ moves toward $1$, the divergence enters with a
vanishing coefficient, and what remains is target estimation error;
this is the sense in which $N$ reduces $\Delta_{\mathrm{dom}}$, while the divergence itself stays as it is. No choice of $\alpha$ or
$N$ enters \cref{eq:infoterm} or \cref{eq:timeterm}, so this is the
only gap class at whose cause post-training is aimed;
\cref{fig:repairdist} measures the rate, and the robustness cost
\citep{kumar2022finetuning} that accompanies it.

\subsection{Why the gates come before ranking}
\label{app:lexico}
The two gates of \cref{eq:lexico-main} are operational checks, not measurements of the theoretical risk gaps (\cref{sec:gates}).

Why this order? A deployer who picks by benchmark score takes the
top-scoring candidate $c^{\mathrm{score}}$ whether or not it fits the sensors. \textsc{Triage} can lose to that deployer only if skipping the
gates costs less than a ranking mistake.

\begin{remark}[When gating first is better]
\label{remark:lexico}
Suppose any candidate that fails a gate has risk at least
$R_{\mathrm{gate}}$, and the pick $\hat{c}$ has risk at most
$\epsilon_{\mathrm{rank}}$ above the best feasible candidate,
$R(\hat{c}) \le \min_{c\in G} R(c) + \epsilon_{\mathrm{rank}}$. If
$c^{\mathrm{score}}$ fails a gate and
$R_{\mathrm{gate}} > \min_{c\in G} R(c) + \epsilon_{\mathrm{rank}}$,
then
\begin{equation}
R(\hat{c}) \;\le\; \min_{c \in G} R(c) + \epsilon_{\mathrm{rank}}
\;<\; R_{\mathrm{gate}} \;\le\; R\bigl(c^{\mathrm{score}}\bigr).
\label{eq:dominance}
\end{equation}
If no candidate fails a gate, $G=\mathcal{C}$, and $\hat{c}$ is the
score-first pick whenever $\widehat{\Delta}_{\mathrm{dom}}$ orders
candidates as the score does.
\end{remark}

In words: when the scoreboard's favorite does not fit the sensors,
gating first is safer as long as running an unfit candidate costs
more than a ranking mistake among fit ones; when every candidate
fits, gating first changes nothing. Its content is the premise
$R_{\mathrm{gate}} > \min_{c\in G} R(c) + \epsilon_{\mathrm{rank}}$,
the one value judgment the framework makes. Its two sides are in different units: a
gate violation shows up in closed loop as collisions on $80$ to $90\%$ of routes (\cref{fig:timegate}), a ranking mistake in open
loop as $0.18$ to $0.48$ PDMS points of regret
(\cref{fig:selection}b; regret is the PDMS a pick gives up against
the best candidate in the held-out city).

\FloatBarrier

\FloatBarrier
\section{Algorithm and costs}
\label{app:procedure}

This appendix collects the formal statement of the procedure and its
cost accounting. Stages~1 and~2 evaluate the operational checks
$g_{\mathrm{sense}}$ and $g_{\mathrm{lat}}$, not the theoretical
risk gaps of \cref{eq:decomp};
Stages~3 and~4 estimate the ranked term $\Delta_{\mathrm{dom}}$;
Stage~5 spends the budget $N$ on it (\cref{fig:hero}d).

\begin{algorithm}[t]
\caption{\textsc{Triage}: repairability-ordered selection}
\label{alg:triage}
\begin{algorithmic}[1]
\Require pool $\mathcal{C}$; locked target $(S',T',D')$; deadline
$d'$; local logs ($\sim$10 min); data budget $N$
\State \textbf{Read on paper:} for each $c\in\mathcal{C}$, record the sensor contract and the documented memory footprint
\State \textbf{Sensing gate ($g_{\mathrm{sense}}$):} drop $c$ if its required
channels are neither contained in $S'$ nor closable by a
semantics-preserving conversion; log conversions as named
interface interventions that re-enter as their own configurations
\State \textbf{Latency gate ($g_{\mathrm{lat}}$):} for each candidate that passes the sensing gate, measure \trbnew{\trbold{on the deployment hardware in its operating state} the end-to-end latency from the acquisition of the newest frame to the finished plan, and take its 95th percentile as $\hat\tau_c$ (an inference time above $d'$ already excludes)}; drop $c$ if it does not fit
in the GPU's memory or if \trbnew{$\hat\tau_c$} $> d'$, where $d'$ is the control period under command hold and a preset replanning deadline when
the controller tracks the last plan with fresh state. Validate that
execution configuration in closed loop before release; stale-image
tolerance does not validate stale-plan tolerance
\State \textbf{If no candidate survives:} \Return the failing gate per candidate (recorded as pass, fail or unevaluated); no ranking is
defined. Either commission the repair that the failing gate's class names, or
adopt plan tracking as a new configuration with a deadline specified
independently of measured candidate latencies and validated before
release; re-enter both gates. Never widen a
gate to obtain a selection
\State \textbf{Probe:} run remaining candidates offline on recorded local logs; verify
sign diagnostics and blind anchors; normalize each response by the model's response to a constant gray frame; report readings below the model's run-to-run noise as not attributable; compute probe readings
\State \textbf{Rank (\trbnew{gate-restricted benchmark ranking;} estimate $\Delta_{\mathrm{dom}}$):} \trbnew{among the
remaining candidates, select the one with the highest target-relevant score,}
preferring models trained on target-domain data at equal scores; designate the
runner-up as challenger. \trbnew{The probe readings confirm or veto each candidate's architectural tier (traditional stacks; VLAs and world models) and supply the checks of line~5; they do not rank within a tier}
\State \textbf{Allocate and verify:} spend $N$ on post-training the
selection while monitoring tracking error, perturbation sensitivity and reliance on the image together; a dose whose sensitivity exceeds \trbnew{a set multiple of} its
zero-shot level is not deployable until it recovers; test \trbnew{every adapted model and} any converted interface or moderate sensory deficit in closed loop
before release; buy hardware only for sensing gaps; commission
architecture only for latency gaps
\State \Return feasible set, selection + challenger, budget
allocation
\end{algorithmic}
\end{algorithm}

\textbf{What the procedure costs.} Cheap evidence runs first and
cuts the pool before expensive evidence is owed. On the
thirteen-configuration pool of this paper (two of them blind
anchors), Stages~1--2 cost one person-afternoon of sensor-contract reading
plus a few minutes of timing per candidate with no other process on the GPU \trbnew{(\TrbTimingGpuMin{} GPU minutes of model calls for the \TrbTimingNModels{} configurations we timed together)}; on a monocular camera setup they excluded the six-camera and LiDAR sensor contracts, and
one candidate on its own documented $72{,}115$\,MiB memory footprint, before
a single frame was processed. \trbnew{Stage~3 runs each remaining candidate offline on ten minutes of recorded logs per city; for the \TrbProbeNModels{} world models we timed, this took \TrbProbeGpuhLo{} to \TrbProbeGpuhHi{} GPU hours per city on one RTX A6000, model loading included.} Only Stage~4 requires the expensive artifact,
a reproduction of each remaining candidate's published score under the benchmark's evaluator, up to a GPU-day per model (RQ8). 

The probe readings, their sanity checks and the two checks that bound what they license are defined in \cref{app:instruments}.

\clearpage

\section{Extended results}
\label{app:extended}

This appendix gives the complete account of every research question, with the setup, all results and the per-city readings that the main text summarizes. Example frames are medians under a stated ranking. Driving models are cited in \cref{tab:veh-gates} and manipulation models in \cref{tab:manipcontracts}.

\subsection{The sensing gate: measure, attempt repair, test safety}
\label{app:ext-info}

This section validates Stage~1 of \cref{fig:hero}d and the closed-loop step of Stage~5 by following one deficit through the
framework's arc. We measure how much a sensory loss damages behavior
(RQ1), test whether fine-tuning can compensate it (RQ2), and ask
what open-loop compensation is worth in closed loop (RQ3).

\RQ{1}\textbf{: At equal pixel budgets, does structure loss damage
more than detail loss?}

\emph{Setup.} Open loop, real frames, seven configurations, four
cities. Center crops that keep $90$, $75$ and $60\%$ of each image
dimension, resampled to native size, remove structure; downscaling
each dimension to $75$, $50$ and $33\%$ and back removes detail. Both are read as mean plan displacement on identical
frames; blind anchors are included as sanity checks that must respond exactly zero.

\begin{figure}[!ht]
\centering
\includegraphics[width=0.95\linewidth]{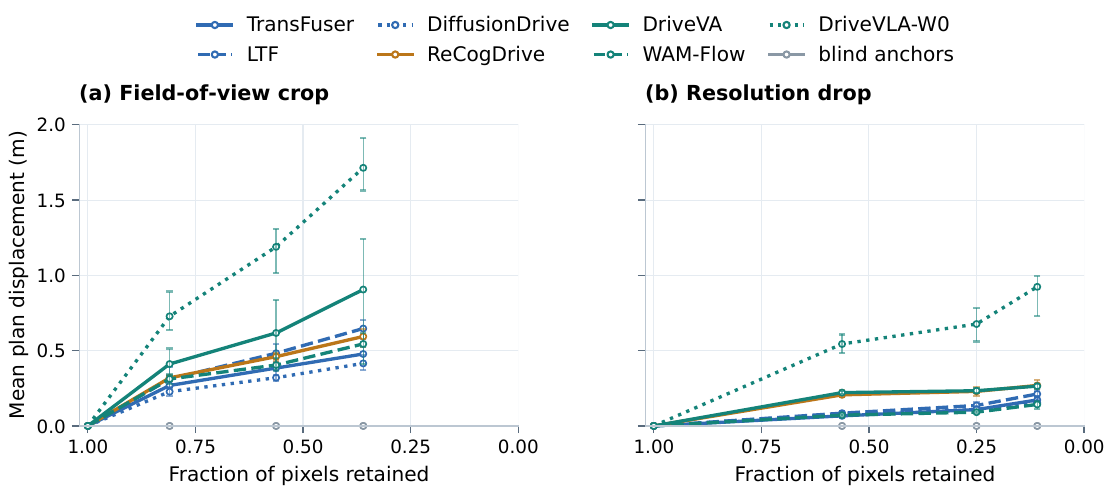}
\caption{\textbf{RQ1: losing view vs.\ losing sharpness.} \textbf{(a)}~Response to field-of-view crop (lines: four-city mean; bars: range across cities). \textbf{(b)}~Response to resolution reduction, same axes. Blind anchors read zero on both.}
\label{fig:infogate}
\end{figure}

\begin{figure}[!ht]
\centering
\includegraphics[width=0.91\linewidth]{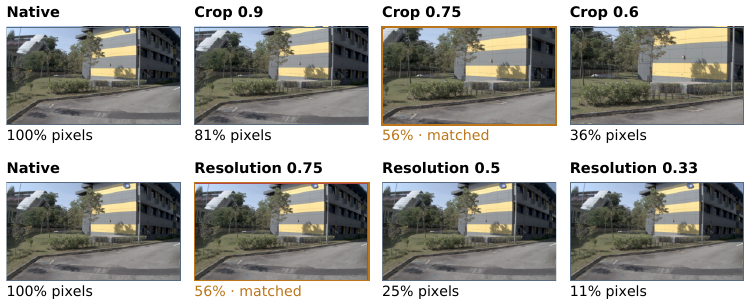}\\[4pt]
\includegraphics[width=0.91\linewidth]{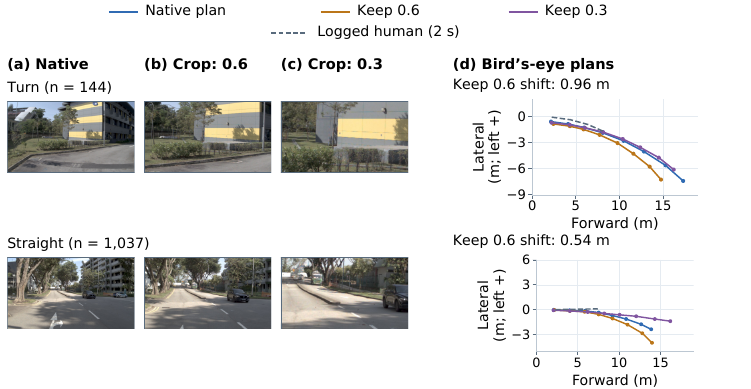}
\caption{\textbf{The matched pixel budget, and the same crop on a turn and on a straight road.} \emph{Top:} crop and resolution ladders on one Singapore turning frame; both keep \texttt{keep}$^2$ of the pixels and meet at \texttt{keep}~$=0.75$ (outlined), where \cref{fig:info}a is read. \textbf{(a--c)}~Median Singapore turning and straight frames, native and cropped. \textbf{(d)}~Their plans in the ego frame; dashed: logged human path at $2$\,s.}
\label{fig:matchedbudget}\label{fig:cropplan}
\end{figure}

\emph{Results.}
\textbf{Losing view costs $2.2$--$5.7\times$ more than losing sharpness.} Every sighted
model's plans displace further as the view narrows
(\cref{fig:infogate}a): DriveVLA-W0 climbs steepest, from $0.73$\,m at
the mildest crop to $1.71$\,m at the strongest, and the two
LiDAR-carrying stacks, TransFuser and DiffusionDrive, climb least
($0.21$ and $0.19$\,m). Resolution curves (\cref{fig:infogate}b) are
shallower on the same axes, and blind anchors sit at exactly zero in
both panels. With $56\%$ of pixels kept on both axes
(\cref{fig:info}a), the two inputs at the top of \cref{fig:matchedbudget}, losing them at the border costs $2.2$--$5.7\times$
more than losing them everywhere, for every sighted configuration and
in every city (the smallest city-level ratio is $1.9\times$). Six of
the seven respond less to $1/9$ of native resolution than to the
mildest crop, which removes only $19\%$ of pixels; DriveVLA-W0 is the
exception.

\textbf{The ratio does not track the sensor suite; the absolute stake
does.} The three largest ratios belong to the tri-camera-plus-LiDAR
TransFuser, its camera-only sibling LTF and the monocular WAM-Flow
(all $5.7\times$), the smallest to the monocular ReCogDrive and
DriveVLA-W0 ($2.2\times$). The closest comparison in the pool,
TransFuser against LTF, two separately trained checkpoints of one
architecture with and without LiDAR input, shows LiDAR lowering the
crop response by $20\%$ and the resolution response by $21\%$, which
leaves the ratio unchanged. The absolute response differs more: at
the matched budget the two largest are monocular, DriveVLA-W0
($1.19$\,m) and DriveVA ($0.62$\,m), while the other two monocular
candidates ($0.46$ and $0.40$\,m) sit among the traditional stacks.

\emph{Takeaway.} At equal pixel budgets, losing view costs several times more than losing sharpness for every configuration, whatever its sensors: this is why
the framework treats a narrower view as a gate and a lower resolution
as a format conversion. When forced to trade sensing,
give up resolution before field of view. The ratio says how uneven the trade-off is and the absolute response how much is at stake; the two
rank the pool differently.

\RQ{2}\textbf{: Can fine-tuning compensate a moderate sensory
deficit?}

\emph{Setup.} Open loop, real frames. We fine-tune TCP under the
$0.6$ field-of-view crop on each of its three stitched cameras,
evaluated under the crop, against a matched-budget uncropped run with identical data, hyperparameters, and epochs. TCP is used
because a paired design needs a public, exactly replicable training
recipe; RQ3 covers the other tiers in closed loop. We report turning frames separately, where the
missing lateral view should matter most ($252$ turning frames across
three cities; the fourth contributes none in this window). We repeat
the design on ReCogDrive, a vision-language-action model, with a LoRA \citep{hu2022lora} recipe we denote LoRA(LM)+head, used for every adapted ReCogDrive-2B: its vision-tower adapters did not train, so the vision encoder stays frozen; its set of turning frames is a different $222$ frames. The criterion: a sensing gap should leave a residual of at least $25\%$ of the uncropped repair.

\emph{Results.}
\textbf{Whether the residual survives depends on the base model.} Before
any adaptation the damage sits on turns: on ReCogDrive the crop keeping
$60\%$ raises tracking error against the human driver by $0.221$\,m on
turning frames and $0.091$\,m over all frames, $2.4\times$
concentrated. \cref{fig:cropplan} shows the same crop on one turning and one straight frame (a--c) and the plans it moves (d). On TCP both runs recover $73$--$76\%$ of their zero-shot
turn error (\cref{fig:repair}a), and the residual, though of the
predicted sign, is $1.3\%$ of the repair: the criterion fails by a factor of twenty. \trbnew{Repeating TCP's adaptation over crop size, local data size and training seed (\TrbEthreeNArms{} adapted runs) leaves this unchanged: with \TrbEthreeNmax{} local samples the crop raises the adapted model's turn error, \TrbEthreeNativeTurn\,m uncropped, by \TrbEthreePSixMean\,m at \texttt{keep} $=0.6$ (mean of three seeds; seed spread \TrbEthreeS\,m) and by \TrbEthreePThree\,m at $0.3$ (one seed), and every frame-resampled 95\% interval of this penalty includes zero. The penalty exceeds the seed spread at \texttt{keep} \TrbEthreeKeepsNR{}, where it is also not lower at \TrbEthreeNmax{} than at \TrbEthreeNmid{} samples by more than that spread, and lies within the seed spread at \TrbEthreeKeepsWithin; it is not monotone in crop size, so for TCP it is a residual of a few centimeters near the noise floor.} On ReCogDrive the residual on turning frames is $0.056$\,m against a repair of $0.130$\,m, or
$43\%$ \trbnew{(\cref{fig:repair}c and \cref{tab:trb-e4})}, and sixteen times the all-frame residual ($0.0035$\,m; about $0.8$\,mm on straight frames): straight driving is repaired, turns are not, and the criterion is met. The fine-tuning recipe left the vision encoder frozen, \trbnew{which accounts for most of this residual, as the runs with the encoder adapting show below}.

\textbf{The larger the gap, the less local data buys back.} Keeping
$30\%$ raises turn damage to $0.371$\,m and all-frame damage to
$0.251$\,m, so the damage is no longer turn-specific ($1.5\times$). The
residual is then $0.133$\,m, $101.8\%$ of the same repair
(\cref{fig:repair}c); the repair-normalized criterion passes on the pooled turning frames and in Singapore and Pittsburgh, and the crop-damage
criterion in every city with turns ($0.577$, $0.671$, $0.705$
compensated). From \texttt{keep} $=0.6$ to $0.3$ the zero-shot turn
damage grows $1.67\times$ while the residual grows $2.36\times$, so the
compensated fraction falls from $0.746$ to $0.642$. The severe residual
is only $5.1\times$ concentrated on turns ($16\times$ at the moderate
crop): the moderate crop locates the surviving deficit, the severe crop shows that one survives in every city. The residual is positive in all
three cities with turns ($+0.041$, $+0.149$, $+0.065$\,m at
\texttt{keep} $=0.6$\trbnew{; \cref{fig:repair}d shows the runs with the encoder adapting}), while the
residual-to-repair ratio is undefined in Boston, where uncropped
adaptation raises turn error from $0.793$ to $0.874$\,m. Per frame, the
damage, repair and residual agree in sign with their pooled values on
$73\%$, $70\%$ and $62\%$ of turning frames (all three on $31\%$), so we judge the pooled turning frames.

{\trbnewon\textbf{With the vision encoder adapting, the moderate crop is largely repaired.} With the vision adapters trained, the same recipe, on the same \TrbEfourNUsable{} usable samples with the same hyperparameters and epochs, updates all \TrbEfourNVision{} vision adapter tensors together with the language adapters and the head. We trained \TrbEfourNArms{} runs (uncropped and keep $0.6$ with two seeds each, keeps $0.75$, $0.45$ and $0.3$ with one) and evaluated them, with the zero-shot model at the same crops, on the same frames and with the same per-frame diffusion noise as above.

At keep $0.6$ the residual is \TrbEfourResSixA{} and \TrbEfourResSixB\,m in the two seeds, \TrbEfourShareSixA{} and \TrbEfourShareSixB\% of their repairs (\TrbEfourRepairA{} and \TrbEfourRepairB\,m), and the compensated fraction is \TrbEfourCfSixA{} and \TrbEfourCfSixB, against $0.746$ with the encoder frozen (\cref{tab:trb-e4}). Both residuals are positive and exceed the seed spread ($S = \TrbEfourS$\,m, the larger of the two seed-to-seed differences in turn error, uncropped and keep $0.6$), but their intervals reach down to \TrbEfourLoSixA{} and \TrbEfourLoSixB\,m, so neither residual is distinguishable from zero, and both are below the $25\%$ criterion of RQ2. The residual stays positive in all three cities with turns in both seeds, but it is only \TrbEfourConcLo{} to \TrbEfourConcHi{} times the all-frame residual, against $16$ times with the encoder frozen, and on straight frames the cropped runs err \TrbEfourStraightSixLo{} to \TrbEfourStraightSixHi\,mm more than the uncropped ones, against under a millimeter. Larger crops leave a residual whose interval lies above zero (one seed each): \TrbEfourResFourFive\,m (\TrbEfourShareFourFive\% of the repair) at keep $0.45$ and \TrbEfourResThree\,m (\TrbEfourShareThree\%) at $0.3$, with interval lower bounds of \TrbEfourLoFourFive{} and \TrbEfourLoThree\,m and compensated fractions of \TrbEfourCfFourFive{} and \TrbEfourCfThree. The mildest crop, keep $0.75$, leaves \TrbEfourResSevenFive\,m (\TrbEfourShareSevenFive\%, lower bound \TrbEfourLoSevenFive\,m), more than keep $0.6$, so the curve is not monotone at its mild end; its zero-shot damage of \TrbEfourDamageSevenFive\,m is below the $0.10$\,m at which the compensated fraction is defined. Most of the $43\%$ at keep $0.6$ therefore came from the frozen encoder. An adapting encoder recovers most of a moderate crop's cost on these frames, a residual remains at the other three crop sizes, and with either recipe the larger the crop, the smaller the share that adaptation removes.\par}

\begin{table}[!ht]
\centering\footnotesize\setlength{\tabcolsep}{3pt}
\trbnewon
\caption{\trbnew{\textbf{ReCogDrive under a crop, encoder frozen and adapting (RQ2).} Turn-frame L2 at $2$\,s, pooled over the $222$ turning frames. Damage: zero-shot crop minus zero-shot native; repair: zero-shot native minus adapted native; residual: adapted crop minus adapted native, with its 95\% interval; compensated: one minus residual over damage, where the damage is at least $0.10$\,m. Rows are compared within a block.}}
\label{tab:trb-e4}
\begin{tabular*}{\linewidth}{@{\extracolsep{\fill}}lcccccccc@{}}
\toprule
Encoder & Keep & Seed & Damage & Repair & Residual, turn (m) & \% of & Straight & Compen-\\
 & & & (m) & (m) & [95\% interval] & repair & (mm) & sated\\
\midrule
Frozen & 0.6 & 1 & 0.221 & 0.130 & 0.056 [0.020, 0.096] & 43 & 0.8 & 0.746\\
Frozen & 0.3 & 1 & 0.371 & 0.130 & 0.133 [0.073, 0.202] & 102 & 20.5 & 0.642\\
\midrule
Adapting & 0.75 & 1 & 0.087 & 0.117 & 0.041 [0.017, 0.064] & 35 & 5.6 & --\\
Adapting & 0.6 & 1 & 0.218 & 0.117 & 0.022 [$-$0.001, 0.045] & 19 & 13.9 & 0.900\\
Adapting & 0.6 & 2 & 0.218 & 0.125 & 0.026 [$-$0.001, 0.052] & 21 & 7.7 & 0.882\\
Adapting & 0.45 & 1 & 0.339 & 0.117 & 0.055 [0.014, 0.101] & 47 & 12.6 & 0.837\\
Adapting & 0.3 & 1 & 0.370 & 0.117 & 0.090 [0.044, 0.141] & 77 & 22.1 & 0.757\\
\bottomrule
\end{tabular*}
\end{table}

\emph{Takeaway.} Whether a moderate sensory deficit can be adapted
away depends on the base model, the fine-tuning recipe and the dose as well as on the deficit. The traditional stack, retrained end to end,
relearns to drive from the narrower view (criterion not met), consistent
with the measured-or-inferable criterion of \cref{sec:principle}.
The vision-language-action model under LoRA(LM)+head, its vision
encoder frozen, keeps a residual on the frames the theory predicts, the turns (criterion met). \trbnew{With its encoder adapting, it repairs most of the moderate crop as well (criterion not met) and keeps a residual at the other three crop sizes, the largest under the strongest crop.} Do not
conclude from open-loop metrics that a sensor downgrade has been
trained away; the next question is always the closed-loop one.

\RQ{3}\textbf{: Does a moderate sensory deficit harm safety even
when capability metrics look intact?}

\begin{figure}[!ht]
\centering
\includegraphics[width=0.99\linewidth]{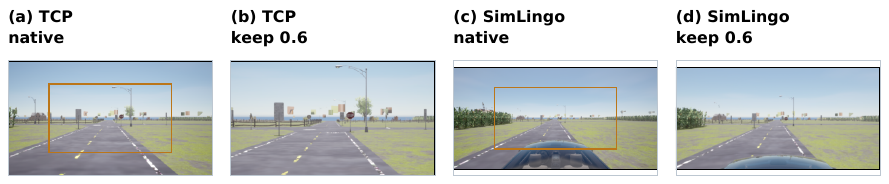}\\[4pt]
\includegraphics[width=0.99\linewidth]{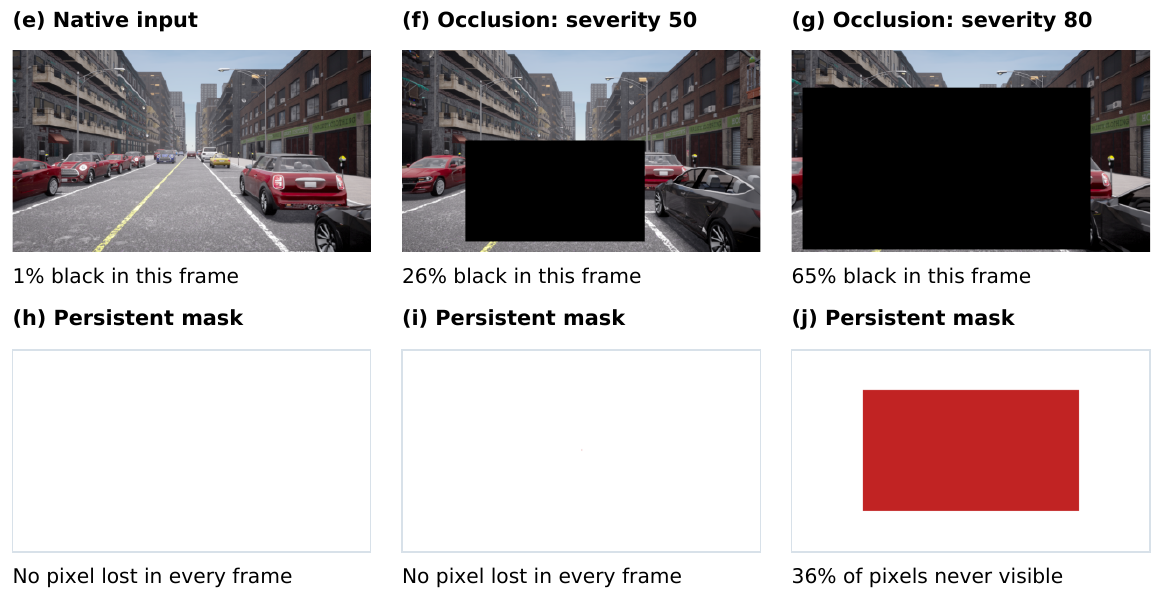}
\caption{\textbf{Closed-loop inputs for RQ3.} \textbf{(a--d)}~The crop keeping $60\%$: captured frame with the kept region outlined, and the frame as delivered, for TCP (one of its three cameras) and SimLingo. \textbf{(e--g)}~One delivered frame without occlusion and at severities $50$ and $80$. \textbf{(h--j)}~Pixels black in all $300$ consecutive frames: none at severity $50$, the central $36\%$ at severity $80$.}
\label{fig:cropframes}\label{fig:occlusionframes}
\end{figure}

\emph{Setup.} Closed loop, simulation, where ground-truth safety is
measurable. First probe, occlusion: a black rectangle is applied to
every camera the configuration declares, redrawn at an independent
random position per camera and frame, at the benchmark's severities
$50$ and $80$, per-dimension ratios that black out $25\%$ and $64\%$
of each frame. Second probe: the crop keeping $60\%$ of RQ1 and RQ2,
introduced zero-shot on two models spanning the pool's tiers, TCP (a
traditional stack) and SimLingo (a leading closed-loop VLA), over
$30$ runs (two configurations, five routes, three seeds) against
native baselines (inputs in \cref{fig:cropframes}a--d); ORION repeats the crop run on the same routes and seeds.

\begin{table}[!ht]
\centering\small
\caption{\textbf{Occlusion (RQ3).} Collision route rate / route completion, Town13, seeds pooled. Severities $50$ and $80$ black out $25\%$ and $64\%$ of every camera frame.}
\label{tab:occlusion}
\begin{tabular*}{\linewidth}{@{\extracolsep{\fill}}lccc@{}}
\toprule
Configuration & no occlusion & severity $50$ & severity $80$\\
 & & (area $25\%$) & (area $64\%$)\\
\midrule
TCP (three forward cameras, stitched) & 0.07 / 0.87 & 0.40 / 0.88 & 0.80 / 0.71\\
SimLingo (single wide camera) & 0.00 / 1.00 & 0.30 / 1.00 & 1.00 / 0.66\\
UniAD front-only (masked) & 0.40 / 0.77 & 0.40 / 0.76 & 1.00 / 0.63\\
UniAD (six-camera) & 0.40 / 0.91 & 0.40 / 0.80 & 1.00 / 0.75\\
VAD (six-camera) & 0.27 / 0.80 & 0.20 / 0.85 & 1.00 / 0.69\\
\bottomrule
\end{tabular*}
\end{table}

\emph{Results.}
\textbf{Occlusion: redundant cameras help only up to moderate severity.} At
severity $50$ the two forward-facing configurations (TCP, three
stitched cameras; SimLingo, one wide camera) see collision route rates
rise by $0.3$, while the two six-camera configurations hold or fall
(UniAD $0.40$ to $0.40$, VAD $0.27$ to $0.20$; \cref{tab:occlusion}).
At severity $80$ every configuration fails regardless of camera count.
\cref{fig:occlusionframes}e--j shows why the effect changes abruptly: essentially
no pixel is black in every frame at severity $50$, whereas the central
$36\%$ is black at severity $80$. Since every camera is occluded
independently, no view stays intact, and six overlapping, partially
occluded fields carry more recoverable scene than one or three forward
fields. The protection does not track the number of views reaching the
model, however: UniAD with five of its six views zeroed does not
degrade either ($0.40$ to $0.40$).

\textbf{The crop damages safety, in a channel that differs by model.}
For TCP and SimLingo, route completion stays within the seed-noise bound (TCP $-0.1$ and SimLingo $-11.7$ points, against
bounds of $8.4$ and $18.0$; \cref{fig:cropcl}c), while collision route
rate rises beyond it, all three seeds landing on the same elevated
value ($0.067$ to $0.400$ for TCP, $0.000$ to $0.200$ for SimLingo;
panel~d). SimLingo also acquires a failure its native runs never show:
$4$ of $15$ cropped routes end in a scenario timeout, running past the $200$\,s limit. ORION is damaged in a different channel. Its route
completion falls from $0.973$ to $0.770$ with a seed range of $0.0026$ (native $0.081$), and the routes abandoned with the vehicle
stuck rise from $1$ to $5$ of $15$; its collision route rate swings by
up to $0.4$ across seeds and is uninformative. In both
patterns the damaged channel has near-zero seed variance \trbnew{in these runs} while the
intact channel swings widely, so a single-seed study could misread
either model.

{\trbnewon\textbf{Crop against resolution on the closed-loop models' frames.} As the CARLA-frame counterpart of RQ1 for the closed-loop checkpoints, we replayed recorded Town13 frames ($\TrbEtwoNTurn$ turning and $\TrbEtwoNStraight$ straight, weighted by their share of the log) through TCP and SimLingo under the crop and the resolution cut of RQ1, at matched pixels and with operators identical to those of the closed-loop runs. Cropping moved the plans more than downscaling at every keep from $90$ to $60\%$ of each dimension: at the $56\%$ budget of RQ1 by \TrbEtwoTcpRatioMid{} times for TCP (95\% interval \TrbEtwoTcpRatioMidLo{} to \TrbEtwoTcpRatioMidHi) and \TrbEtwoSlRatioMid{} times for SimLingo (\TrbEtwoSlRatioMidLo{} to \TrbEtwoSlRatioMidHi), and over all three keeps by \TrbEtwoTcpRatioMin{} to \TrbEtwoTcpRatioMax{} and \TrbEtwoSlRatioMin{} to \TrbEtwoSlRatioMax{} times. TCP's repeated calls and identity crops returned exactly zero displacement. As released, the SimLingo agent keeps dropout active at inference, and our closed-loop runs use it unchanged; its run-to-run noise, \TrbEtwoSlNoise\,m (95\% interval \TrbEtwoSlNoiseLo{} to \TrbEtwoSlNoiseHi), is of the same order as its displacement under the resolution cut (\TrbEtwoSlDownMin{} to \TrbEtwoSlDownMax\,m), so these runs fix the random seed per frame across conditions and the displacement reflects only the image change. UniAD and VAD, run the same way, show the same pattern: over the three keeps cropping moved their plans \TrbEtwoUnRatioMin{} to \TrbEtwoUnRatioMax{} and \TrbEtwoVaRatioMin{} to \TrbEtwoVaRatioMax{} times more than downscaling, so across the four checkpoints the ratio lies between \TrbEtwoAllRatioMin{} and \TrbEtwoAllRatioMax.\par}

{\trbnewon\textbf{Crop against resolution in closed loop.} We drove the four closed-loop checkpoints on the five Town13 routes of RQ3 with three traffic seeds each, natively and under both cuts, with the same pixel-budget operators, the same evaluator settings as RQ3 and no planning deadline (\cref{tab:trb-e1}): TCP and SimLingo at keeps of $90$, $75$ and $60\%$, UniAD and VAD at $75$ and $60\%$, the last two with their camera intrinsics unchanged when cropped, as in the ORION run of RQ3. The cropped runs are new; at $60\%$ they repeat the crop of RQ3 and again raise the collision route rate of TCP and SimLingo, to $0.47$ and $0.33$, with per-seed values of $0.4$, $0.6$, $0.4$ and $0.4$, $0.4$, $0.2$. Every model scored lower under the crop than under the resolution cut at keeps of $75$ and $60\%$. At $60\%$, TCP scored \TrbEoneTcpCropCDs{} under the crop against \TrbEoneTcpDownCDs{} under the resolution cut (native \TrbEoneTcpNatDs) and collided on \TrbEoneTcpCropCCrr{} of its routes against \TrbEoneTcpDownCCrr; SimLingo scored \TrbEoneSlCropCDs{} against \TrbEoneSlDownCDs{} and collided on \TrbEoneSlCropCCrr{} of its routes against \TrbEoneSlDownCCrr; and VAD scored \TrbEoneVaCropCDs{} against \TrbEoneVaDownCDs{} (native \TrbEoneVaNatDs) and collided on \TrbEoneVaCropCCrr{} of its routes against \TrbEoneVaDownCCrr, a gap that exceeds twice its seed spread. Counted per kilometre driven, collisions at $60\%$ were \TrbEoneTcpCropCCpk{} under the crop against \TrbEoneTcpDownCCpk{} under the resolution cut for TCP, \TrbEoneSlCropCCpk{} against \TrbEoneSlDownCCpk{} for SimLingo and \TrbEoneVaCropCCpk{} against \TrbEoneVaDownCCpk{} for VAD. For UniAD the differences stay within its seed spread, which reaches \TrbEoneUnDsSdMax{} points: \TrbEoneUnCropBDs{} against \TrbEoneUnDownBDs{} at $75\%$ and \TrbEoneUnCropCDs{} against \TrbEoneUnDownCDs{} at $60\%$. Under every resolution cut SimLingo drove as it did natively, with the same score on every seed, and at $90\%$ the two cuts did not differ in driving score by more than twice the seed spread for TCP or SimLingo. Across the four models and \TrbEoneNPts{} points the direction agrees: in \TrbEoneAgree{} of the \TrbEoneNPairs{} matched crop-versus-resolution pairs (\TrbEoneTies{} tie) the cut that moved the plans more in open loop also cost more driving score (one-sided sign test $p = \TrbEoneSignP$), and all \TrbEoneResolved{} pairs that the route-and-seed bootstrap resolves agree (SimLingo at $75$ and $60\%$, VAD at $60\%$). Across models the association is weak: the Spearman correlation of open-loop displacement with driving-score loss is \TrbEoneRho{} (95\% interval \TrbEoneRhoLo{} to \TrbEoneRhoHi); it is \TrbEoneRhoTurn{} (\TrbEoneRhoTurnLo{} to \TrbEoneRhoTurnHi) with displacement on turning frames and \TrbEoneCrrRho{} (\TrbEoneCrrRhoLo{} to \TrbEoneCrrRhoHi) with collision route rate. Within each model, then, the cut that displaces the plans more in open loop also costs more in closed loop, while the size of the open-loop displacement does not reliably order the closed-loop loss across models.\par}

\begin{table}[!ht]
\centering\small
\trbnewon
\caption{\trbnew{\textbf{Closed-loop crop against resolution cut at matched pixels.} Bench2Drive Town13, five routes, three traffic seeds per condition. Keep: fraction of each image dimension retained. DS: driving score, mean over seeds (standard deviation in parentheses); RC: route completion (\%); CRR: share of routes with at least one collision.}}
\label{tab:trb-e1}\label{tab:trb-e1b}
\begin{tabular*}{\linewidth}{@{\extracolsep{\fill}}lccccccc@{}}
\toprule
 & & \multicolumn{3}{c}{TCP} & \multicolumn{3}{c}{SimLingo}\\
\cmidrule(lr){3-5}\cmidrule(lr){6-8}
Cut & Keep & DS & RC & CRR & DS & RC & CRR \\
\midrule
Native & -- & \TrbEoneTcpNatDs{} (\TrbEoneTcpNatDsSd) & \TrbEoneTcpNatRc & \TrbEoneTcpNatCrr & \TrbEoneSlNatDs{} (\TrbEoneSlNatDsSd) & \TrbEoneSlNatRc & \TrbEoneSlNatCrr \\
\midrule
Crop & $0.9$ & \TrbEoneTcpCropADs{} (\TrbEoneTcpCropADsSd) & \TrbEoneTcpCropARc & \TrbEoneTcpCropACrr & \TrbEoneSlCropADs{} (\TrbEoneSlCropADsSd) & \TrbEoneSlCropARc & \TrbEoneSlCropACrr \\
Crop & $0.75$ & \TrbEoneTcpCropBDs{} (\TrbEoneTcpCropBDsSd) & \TrbEoneTcpCropBRc & \TrbEoneTcpCropBCrr & \TrbEoneSlCropBDs{} (\TrbEoneSlCropBDsSd) & \TrbEoneSlCropBRc & \TrbEoneSlCropBCrr \\
Crop & $0.6$ & \TrbEoneTcpCropCDs{} (\TrbEoneTcpCropCDsSd) & \TrbEoneTcpCropCRc & \TrbEoneTcpCropCCrr & \TrbEoneSlCropCDs{} (\TrbEoneSlCropCDsSd) & \TrbEoneSlCropCRc & \TrbEoneSlCropCCrr \\
\midrule
Resolution & $0.9$ & \TrbEoneTcpDownADs{} (\TrbEoneTcpDownADsSd) & \TrbEoneTcpDownARc & \TrbEoneTcpDownACrr & \TrbEoneSlDownADs{} (\TrbEoneSlDownADsSd) & \TrbEoneSlDownARc & \TrbEoneSlDownACrr \\
Resolution & $0.75$ & \TrbEoneTcpDownBDs{} (\TrbEoneTcpDownBDsSd) & \TrbEoneTcpDownBRc & \TrbEoneTcpDownBCrr & \TrbEoneSlDownBDs{} (\TrbEoneSlDownBDsSd) & \TrbEoneSlDownBRc & \TrbEoneSlDownBCrr \\
Resolution & $0.6$ & \TrbEoneTcpDownCDs{} (\TrbEoneTcpDownCDsSd) & \TrbEoneTcpDownCRc & \TrbEoneTcpDownCCrr & \TrbEoneSlDownCDs{} (\TrbEoneSlDownCDsSd) & \TrbEoneSlDownCRc & \TrbEoneSlDownCCrr \\
\midrule\midrule
 & & \multicolumn{3}{c}{UniAD} & \multicolumn{3}{c}{VAD}\\
\cmidrule(lr){3-5}\cmidrule(lr){6-8}
Cut & Keep & DS & RC & CRR & DS & RC & CRR \\
\midrule
Native & -- & \TrbEoneUnNatDs{} (\TrbEoneUnNatDsSd) & \TrbEoneUnNatRc & \TrbEoneUnNatCrr & \TrbEoneVaNatDs{} (\TrbEoneVaNatDsSd) & \TrbEoneVaNatRc & \TrbEoneVaNatCrr \\
\midrule
Crop & $0.75$ & \TrbEoneUnCropBDs{} (\TrbEoneUnCropBDsSd) & \TrbEoneUnCropBRc & \TrbEoneUnCropBCrr & \TrbEoneVaCropBDs{} (\TrbEoneVaCropBDsSd) & \TrbEoneVaCropBRc & \TrbEoneVaCropBCrr \\
Crop & $0.6$ & \TrbEoneUnCropCDs{} (\TrbEoneUnCropCDsSd) & \TrbEoneUnCropCRc & \TrbEoneUnCropCCrr & \TrbEoneVaCropCDs{} (\TrbEoneVaCropCDsSd) & \TrbEoneVaCropCRc & \TrbEoneVaCropCCrr \\
\midrule
Resolution & $0.75$ & \TrbEoneUnDownBDs{} (\TrbEoneUnDownBDsSd) & \TrbEoneUnDownBRc & \TrbEoneUnDownBCrr & \TrbEoneVaDownBDs{} (\TrbEoneVaDownBDsSd) & \TrbEoneVaDownBRc & \TrbEoneVaDownBCrr \\
Resolution & $0.6$ & \TrbEoneUnDownCDs{} (\TrbEoneUnDownCDsSd) & \TrbEoneUnDownCRc & \TrbEoneUnDownCCrr & \TrbEoneVaDownCDs{} (\TrbEoneVaDownCDsSd) & \TrbEoneVaDownCRc & \TrbEoneVaDownCCrr \\
\bottomrule
\end{tabular*}
\end{table}

\emph{Takeaway.} Open loop, adaptation removes most of the crop's damage on one base model and leaves a residual on another; closed loop,
the same crop on released checkpoints damages safety through a
channel that differs by model. Open-loop repair therefore does not certify closed-loop safety. Moderate deficits are therefore tested in closed loop, on several channels and seeds,
and structural ones gate outright; occlusion is priced as occluded
area on every camera.

\subsection{The latency gate: a safety property of the loop}
\label{app:ext-time}

This section validates Stage~2 of \cref{fig:hero}d.

\RQ{4}\textbf{: Is a latency violation a safety event, what decides
its failure mode, and can wrappers or fine-tuning repair it?}
\emph{Setup.} Closed loop in simulation; latency measured on real
hardware. We time every model per frame with no other process on the GPU, then enforce inference deadlines in closed loop on five
routes per town: $500$\,ms in Town03 and Town13 (clear weather, three
seeds) and Town12 (fog, two seeds), and $200$\,ms in Town03. Under a deadline, a missed control step replays the previous actuation command (command \emph{hold}). To separate staleness from how stale outputs are consumed, a second axis freezes each camera's input in random bursts of up to 1 or 3\,s while the model keeps replanning on fresh speed and position and the controller tracks with fresh vehicle state (\cref{app:instruments}). Finally we attempt two repairs. One is an asynchronous-planning wrapper delivering stale plans to the unchanged controller, implemented three ways (coordinate-frozen, raw-GPS
pose, filtered pose); the other is fine-tuning, whose effect on
latency we measure directly.

\emph{Results.}
\textbf{A missed deadline is a safety event whose form depends on the
model.} Latency spans five orders of magnitude on one GPU
(\cref{fig:timegate}a): $0.01$ and $0.12$\,ms for the blind anchors,
$63$--$85$\,ms for the traditional stacks of \cref{tab:pool} and $0.49$--$4.1$\,s for every VLA and world model, so of these only the traditional stacks fit a real-time budget
as shipped. In Town13, where SimLingo completes every route without a
collision when no deadline is enforced (\cref{fig:timegate}b), a
$500$\,ms deadline leaves it still moving but colliding: it covers $48\%$ of route length and collides on $0.8$ of routes, and the
pattern repeats in Town03 ($48\%$, $0.8$) and in fog in Town12
($58\%$, $0.9$). At $200$\,ms (Town03) it does not move and so does
not collide. TCP, inside its budget, keeps its completion at both
deadlines. ORION, a second modern VLA, stops rather than collides, as expected when latency exceeds the deadline: its $1{,}020$\,ms latency (p99 $1{,}032$\,ms) exceeds both deadlines, so all $36{,}000$ of its scheduled calls miss and over thirty routes it never leaves the start (completion and collision route rate both
$0.00$, against $97\%$ completion without a deadline).

\textbf{Stale images are tolerated; stale commands are not.} Freezing the camera input in bursts of up to $1$ or $3$\,s while the loop keeps acting on fresh vehicle state (\cref{fig:timegate}c; two seeds) leaves
collision route rates near baseline: SimLingo stays at $0.00$, UniAD at
$0.40$, VAD moves from $0.27$ to $0.30$ and TCP from $0.07$ to $0.10$
and $0.20$. Replaying the stale command at $500$\,ms instead raises
SimLingo's rate to $0.80$. Planning from an image up to a few seconds old is tolerated; holding an old steering angle is not.

\textbf{Neither a wrapper nor fine-tuning moves a candidate across the
gate.} None of the three asynchronous-planning wrappers works: the coordinate-frozen one did not run, the raw-GPS one stalled at $4\%$ of its first
route, and the filtered-pose one, even without a deadline, completes $0.07$ of route
length with a collision on every route (two seeds;
\cref{fig:timegate}d). The mechanism
is shared: the unchanged controller assumes a plan starts at the vehicle's
current position, so a stale plan's early waypoints fall behind it and
the interpolated trajectory folds back on itself. Fine-tuning does not
move the clock either: it changed TransFuser's forward pass by
$+0.18\%$ ($18.456$ to $18.488$\,ms). The adapted ReCogDrive, timed under the same conditions as the zero-shot model, is $13.3\%$ slower, which makes the gate harder to clear. Its adapters are mounted rather than merged,
adding computation at every adapted module.

\emph{Takeaway.} A missed deadline changes the failure mode as well as the score: one modern model keeps driving and collides, another
never moves, so there is no single degradation curve beyond the gate
and latency belongs in a gate rather than a ranking feature.
The repair lies in the consumer, fresh-state tracking of
low-frequency plans; neither a wrapper nor fine-tuning moves a
candidate across the gate. Measure latency on the deployment GPU and
treat a violation as an exclusion, not a discount.

\subsection{The distribution gap: how large, how structured, and
what repair costs}
\label{app:ext-dist}

This section sizes the gap that Stages~3--5 of \cref{fig:hero}d
exist to measure and shrink (RQ5), then prices its repair (RQ6).

\RQ{5}\textbf{: How large and how structured is the gap that training data and environment create?}
\emph{Setup.} Open loop, real frames: we contrast public
checkpoints of the \emph{same} TransFuser architecture, one trained in CARLA and one trained on real data, under an identical probe code, cities, budget, and output horizon. \cref{app:twins}
documents the pool-wide search for such pairs; its one
vision-language-action candidate differs architecturally across
its two releases at the tensor level and so is not a controlled pair. To test
whether the effect is architecture-specific, we repeat the
contrast on the two ecosystem releases of UniAD, which pass the
same comparison. Closed loop, simulation: a six-environment matrix
(five configurations $\times$ five routes $\times$ three seeds per
environment) including a masked intervention, \texttt{front-zero},
a six-camera model consuming only its front view.

\begin{figure}[!ht]
\centering
\includegraphics[width=0.92\linewidth]{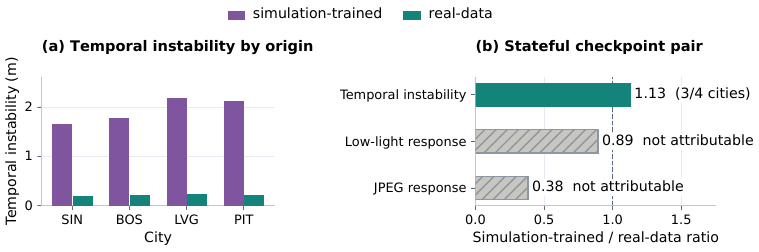}
\caption{\textbf{RQ5: temporal instability and cross-architecture attribution.} \textbf{(a)}~Stateless pair (TransFuser): temporal instability on real frames by training source. \textbf{(b)}~Stateful pair (UniAD): only the temporal measure transfers; hatched measures fall below the run-to-run noise.}
\label{fig:distribution}
\end{figure}

\begin{figure}[!ht]
\centering
\includegraphics[width=\linewidth]{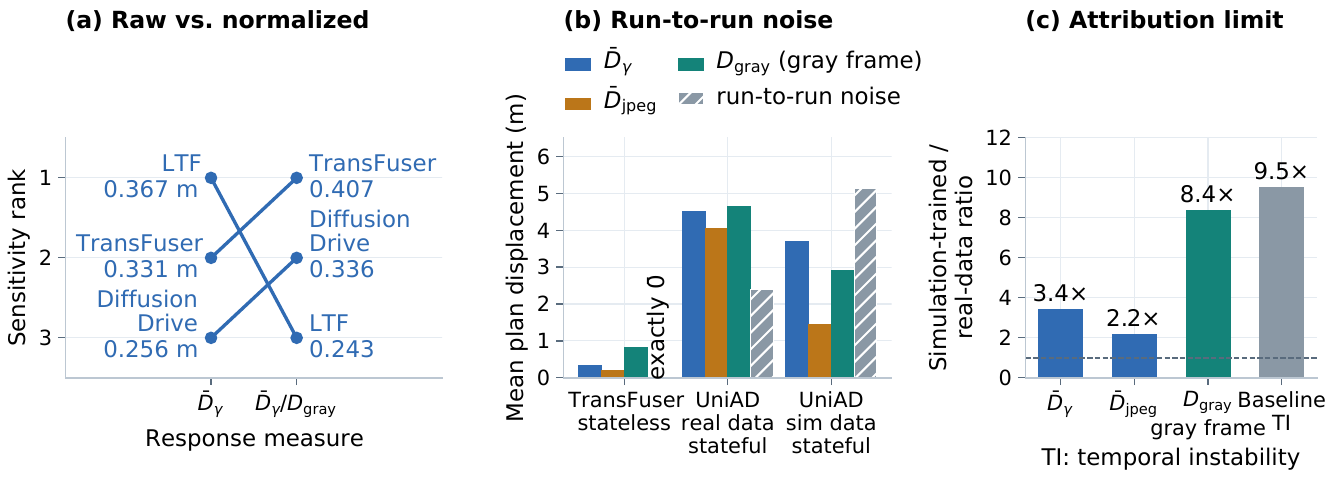}
\caption{\textbf{Raw probe sensitivity is not comparable across models.} \textbf{(a)}~Ranks by raw and by normalized response. \textbf{(b)}~Run-to-run noise: zero on the stateless architecture, above every measure on the stateful simulation-trained checkpoint. \textbf{(c)}~On RQ5's pair the gray-frame gap is matched by the instability gap.}
\label{fig:engagement}
\end{figure}

\emph{Results.}
\textbf{Training data separates identical architectures.} The
simulation-trained TransFuser reacts $1.5$--$4.9\times$ more strongly to
darkened and compressed real frames in all four cities (low-light
response in \cref{fig:dist}a), and its plans disagree with their own
previous frame $8.6$--$10.3\times$ more, again in four of four cities
(\cref{fig:distribution}a). This architecture computes each frame independently: six consecutive calls on identical input return exactly
zero displacement (checked on $120$ Singapore frames; \cref{fig:engagement}b), so run-to-run drift contaminates neither measure. The response to a constant gray frame (RQ7) is inconclusive on this pair
and leaves the gap standing: the simulation-trained side responds
$8.4\times$ more to losing the image, but its baseline instability is
$9.5\times$ larger, and dividing one by the other leaves the two sides
within $12\%$ (\cref{fig:engagement}c).

\begin{table}[!ht]
\centering\small
\caption{\textbf{Six closed-loop environments (RQ5).} Route completion (mean $\pm$ seed s.d., three seeds) and collision route rate (CRR), without perturbations or deadlines. Ranks are plotted in \cref{fig:dist}b.}
\label{tab:sixenv}

\begin{tabular*}{\linewidth}{@{\extracolsep{\fill}}lrrrrrr@{}}
\toprule
& \multicolumn{2}{c}{T03 clear} & \multicolumn{2}{c}{T03 fog} & \multicolumn{2}{c}{T12 clear}\\
\cmidrule(lr){2-3} \cmidrule(lr){4-5} \cmidrule(lr){6-7}
Configuration & RC & CRR & RC & CRR & RC & CRR\\
\midrule
SimLingo & 84.2$\pm$5.2 & 0.07 & 98.0$\pm$3.4 & 0.13 & 95.5$\pm$0.4 & 0.33\\
TCP & 57.6$\pm$0.5 & 0.60 & 58.4$\pm$0.0 & 0.60 & 80.6$\pm$0.0 & 0.60\\
VAD & 66.9$\pm$9.1 & 0.53 & 63.0$\pm$1.5 & 0.47 & 49.3$\pm$0.3 & 0.33\\
UniAD & 33.6$\pm$5.1 & 0.33 & 40.9$\pm$7.1 & 0.47 & 45.7$\pm$10.1 & 0.53\\
UniAD front-only & 47.3$\pm$9.4 & 0.33 & 53.3$\pm$15.5 & 0.27 & 43.7$\pm$3.0 & 0.53\\
\midrule
& \multicolumn{2}{c}{T12 fog} & \multicolumn{2}{c}{T13 clear} & \multicolumn{2}{c}{T13 fog}\\
\cmidrule(lr){2-3} \cmidrule(lr){4-5} \cmidrule(lr){6-7}
Configuration & RC & CRR & RC & CRR & RC & CRR\\
\midrule
SimLingo & 88.0$\pm$6.9 & 0.27 & 100.0$\pm$0.0 & 0.00 & 100.0$\pm$0.0 & 0.07\\
TCP & 80.6$\pm$0.0 & 0.60 & 86.7$\pm$0.3 & 0.07 & 88.2$\pm$1.1 & 0.40\\
VAD & 40.7$\pm$6.4 & 0.20 & 79.7$\pm$8.2 & 0.27 & 79.4$\pm$8.3 & 0.33\\
UniAD & 54.4$\pm$4.9 & 0.67 & 90.6$\pm$16.2 & 0.40 & 84.8$\pm$14.2 & 0.47\\
UniAD front-only & 54.5$\pm$4.3 & 0.80 & 76.7$\pm$4.9 & 0.40 & 71.9$\pm$0.0 & 0.40\\
\bottomrule
\end{tabular*}
\end{table}

\textbf{Environment reorders a closed-loop pool beyond noise.} Across
the six environments (\cref{fig:dist}b, \cref{tab:sixenv}), $13$ of
$15$ environment pairs change some pairwise order among the native
configurations ($14$ of $15$ with the masked line), and each
configuration's spread across environments ($15.8$--$57.0$ completion
points) dwarfs its seed noise ($0.2$--$8.6$ points). The masked
intervention beats its own parent in $3$ of $6$ environments, by
$12$--$14$ points in Town03, a gain that clears seed noise in clear
weather and not in fog, and it never overtakes a native front-view
model.

\textbf{A second architecture narrows the claim.} The simulation-trained
UniAD is again less steady on real frames, in the predicted direction
in three of four cities, but by $1.13\times$ rather than
$8.6$--$10.3\times$. Its two perturbation measures are not readable:
UniAD holds state across calls, and six consecutive calls on identical
input ($120$ frames per city) move its plan by $5.12$\,m on the
simulation-trained side and $2.39$\,m on the real-data side, a floor that
on the simulation-trained side exceeds its own gamma and compression
readings ($3.69$ and $1.44$\,m; \cref{fig:distribution}b). The temporal-instability measure comes from the baseline
pass, is unaffected, and is the one measure that transfers. In the three cities with enough turning frames, the simulation-trained
UniAD agrees with the human driver's turn direction less often
($0.62$--$0.72$ against $0.85$--$0.92$).

\emph{Takeaway.} The distribution gap is large and structured:
training data alone separates identical architectures by up to an
order of magnitude on real frames, and environment alone reorders a
closed-loop pool beyond noise, so rankings do not transfer across
environments. On a second architecture the training-data effect is
real but only one of three measures transfers, the other two being
unreadable on a stateful architecture rather than negative. A masked intervention can beat its parent but does not overtake native sensor contracts.
Prefer the checkpoint trained closest to the domain, measure its gap on
local frames above the candidate's own run-to-run noise, and re-check a
masked or re-wired model as a new candidate.

\RQ{6}\textbf{: What does fine-tuning repair, and at what cost?}
\emph{Setup.} Open loop, real frames. We behavior-clone the
simulation-trained TransFuser on real target-city frames, log-level
disjoint from all evaluation windows, at nested doses of 1k, 5k and
20k frames. We read probe
sensitivities, temporal instability, and tracking error against the real-data checkpoint, which gives the level a dose curve should approach. We replicate the top dose with
an independent seed and run an LP-FT variant \citep{kumar2022finetuning}, the standard remedy for fine-tuning's
robustness cost in classification.

\begin{figure}[!ht]
\centering
\includegraphics[width=0.9\linewidth]{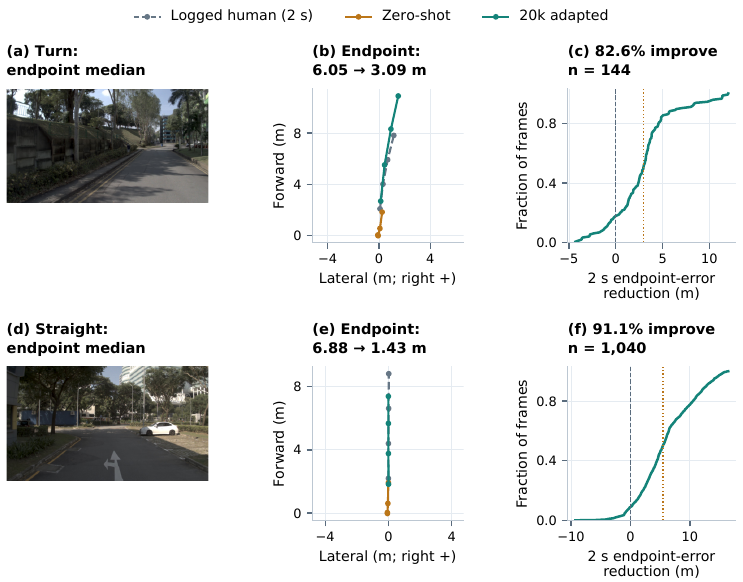}
\caption{\textbf{Adaptation measured by 2\,s endpoint error.} Simulation-trained TransFuser, Singapore. \textbf{(a,b,d,e)}~Median turning and straight frames by the zero-shot-to-20k error reduction. \textbf{(c,f)}~The full distributions; $82.6\%$ and $91.1\%$ of frames improve.}
\label{fig:repairframescdf}
\end{figure}

\emph{Results.}
\textbf{On TransFuser the repair costs robustness at small doses.}
Stability behaves as a deployer would hope (\cref{fig:repairdist}c):
the sim-versus-real instability gap falls monotonically to $25\%$ of
its zero-shot size ($75\%$ closure) by 20k frames, and an independent
seed, run at 20k only, lands within $4\%$ of it and leaves every conclusion unchanged. The two sensitivities do the
opposite. After 1k frames the model reacts $2.4\times$ more strongly
to darkened and $3.4\times$ more strongly to compressed inputs than
before fine-tuning, in every city ($1.4$--$3.3\times$ and
$3.0$--$3.6\times$), leaving a gap to the real-data checkpoint $3.0$ to
$4.7\times$ its zero-shot size; at 20k both are still above their
zero-shot values, so the closure criterion is not met ($1/3$ components). LP-FT enlarges the small-dose spike: at 1k its responses are
$1.96\times$ (darkening) and $1.60\times$ (compression) those of plain
fine-tuning, higher in all four cities. Tracking error against the
human driver nonetheless falls monotonically from $8.0$ to $1.4$\,m
(\cref{fig:repairdist}d), and frame by frame the gain is broad:
$82.6\%$ of turning and $91.1\%$ of straight frames improve
(\cref{fig:repairframescdf}c,\,f; median frames in a, b, d and~e),
and mean trajectory error improves on $86.8\%$ of turning and $93.5\%$ of straight frames, though monotonically across doses on only $28.5\%$ and $27.4\%$, so the repair works even while
robustness dips.

\begin{figure}[!ht]
\centering
\includegraphics[width=0.97\linewidth]{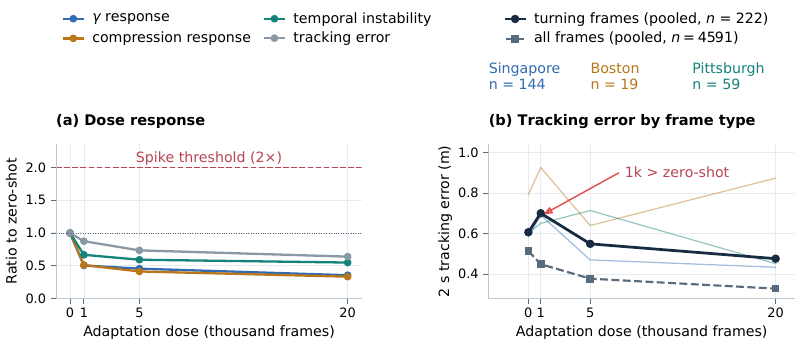}
\caption{\textbf{The dose series on ReCogDrive (LoRA(LM)+head).} \textbf{(a)}~Every measure falls with dose, far below the spike threshold. \textbf{(b)}~Tracking error on turning frames (pooled) and over all frames; thin lines: cities. On turns the smallest dose is worse than zero-shot.}
\label{fig:dose}
\end{figure}

\textbf{On ReCogDrive no spike appears, but a small dose moves turns
backwards.} Under the LoRA(LM)+head recipe at the same budgets
(\cref{fig:dose}a), the response to darkening falls to $0.50$, $0.46$
and $0.36$ of zero-shot, the response to compression to $0.51$,
$0.41$ and $0.33$, and temporal instability to $0.67$, $0.59$ and
$0.55$; the largest darkening or compression ratio, $0.51$, is a quarter of the spike threshold of $2.0$, which applies to those two responses. Tracking error improves to $0.87$, $0.73$ and $0.64$ of
zero-shot over all frames, but on turning frames the smallest dose is
worse than no fine-tuning ($1.15$) before improving to $0.91$ and
$0.79$ (\cref{fig:dose}b). Boston, $19$ of the $222$ pooled turning frames, is the
only city whose largest dose regresses.

\textbf{Part of that fall is the model attending less to the image.}
Falling sensitivity fits a more robust model and equally one that
leans more on ego state and its own priors. The response to a constant gray frame (RQ7) separates the two, against a null control that repeats the
smallest dose with only the training seed changed: reseeding alone
moves the response to losing the image to $0.868$ of the original,
and combining that seed half-width with the pool's city dispersion gives a band of chance variation of $[0.709, 1.160]$ (\cref{app:instruments}). Every variant tested falls below it:
$0.705$, $0.526$ and $0.473$ at the three doses, $0.438$ and $0.346$
on the two cropped runs.

\emph{Takeaway.} Fine-tuning repairs the class the framework assigns
it, but at a cost. On TransFuser it raises perturbation sensitivity at small doses, and LP-FT makes this
worse; on ReCogDrive every response falls, partly because the model
attends less to the image, which a sensitivity measure alone would
score as an improvement. The spike is thus a property of the fine-tuning recipe and the base model, not of post-training, and \cref{alg:triage}'s monitored fine-tuning step must watch reliance on the image beside sensitivity. Monitor both at every
dose against a run that changes only the training seed, budget past the small doses,
and check separately the frames the deployment depends on.

\subsection{Selection with the full framework}
\label{app:ext-select}

We now \trbnew{compose the gates and the selection stages} (RQ7) and test whether
it admits new candidates (RQ8).

\RQ{7}\textbf{: Does \textsc{Triage} select better than standard
practice, and where does cheap evidence stop?}
\emph{Setup.} We compare \trbnew{picking the top benchmark score without gates
(score-first), picking the closed-loop champion (closed-loop-first), and
\cref{alg:triage}, whose Stage~4 is a gate-restricted benchmark ranking,}
across four deployment profiles
(monocular or full camera set; real-time, a $150$\,ms deadline, or
low-speed, a $1.2$\,s replanning deadline under plan tracking). \trbnew{Offline, the latency gate is applied to the compute-only p50 of \cref{tab:pool}, a lower bound on the end-to-end latency $\hat\tau_c$: its exclusions stand, and its passes would still need the end-to-end latency measured on the platform.} Fixed predictor families then select within a tier from probe features under
leave-one-city-out, to locate where cheap evidence stops resolving.

\emph{Results.}
\textbf{The gates do the selection work.} Score-first selection
violates the latency gate in both real-time profiles, and closed-loop-first
selection violates the sensing gate in all four, because neither profile's cameras provide the champion's single $110^\circ$ view
(\cref{fig:deploymentprofiles}). \textsc{Triage} is feasible in three
profiles and agrees with the scoreboard wherever every gate passes. In
the monocular real-time profile it has nothing to return: every
sighted candidate inside the $150$\,ms budget needs three cameras and every
monocular one exceeds it, so it returns each candidate's failing gate instead (\cref{alg:triage}, line~4). This empty pool admits no public
checkpoint until the deployer adopts plan tracking or changes the sensors;
a higher score does not change that.

\textbf{Cheap probes place tiers but cannot rank within one.} In the leave-one-city-out test (\cref{fig:selection}b), nine configurations
in four cities give $36$ \trbnew{rows and \TrbLocoSelections{} selections, one per held-out city,} with ten fixed
features (three environment offsets, seven probe measures) and four
predictor families chosen within each fold\trbnew{; no gate is applied}. The probes never pick below the top tier, but within it they lose to picking the top-scoring candidate: $0.48$ against $0.18$ PDMS points of mean regret. \trbnew{With \TrbLocoSelections{} selections the two rules cannot be separated; the per-fold differences are \TrbPairedFolds{} points. With the gates applied (\cref{tab:trb-selection}), the gate-restricted benchmark ranking gives up \TrbGatedScoreMin{} to \TrbGatedScoreMax{} points in the three profiles with feasible candidates (ReCogDrive and DriveVA in the monocular $1.2$\,s profile; TransFuser, LTF and DiffusionDrive in the full $150$\,ms one; all five in the full $1.2$\,s one); the probe predictor restricted to the feasible candidates gives up \TrbGatedProbeMin{} to \TrbGatedProbeMax, the same predictor refit on their rows only \TrbRefitMin{} to \TrbRefitMax, and a uniform pick among them \TrbRandomMin{} to \TrbRandomMax{} in expectation. Ranking by the score without the held-out city, or by the best candidate of the training city nearest to the held-out one, with any of three feature sets, makes the same picks as the top score, and so does adding DriveVLA-W0 and WAM-Flow, which have no rows in this test, to the candidate list.} Adding latency as a feature makes selection worse
($0.48\to1.29$), since a constant across cities cannot help decide
between them. This test uses none of the per-candidate information we record (sensor contract, latency, license), so feasibility is
decided by the gates alone.

\begin{table}[!ht]
\centering\small
\caption{\trbnew{\textbf{Selection rules under leave-one-city-out (RQ7).} Mean regret in PDMS points over the four folds, against the best candidate that passes both gates in the held-out city, on the nine configurations of \cref{fig:select}b. Profiles: monocular or full camera set, $150$\,ms or $1.2$\,s deadline, with the latency gate on the compute-only p50 of \cref{tab:pool}. In parentheses, folds whose pick fails the sensing (S) or latency (L) gate; a negative regret means that an infeasible pick outscores every feasible one. --: no feasible candidate.}}
\label{tab:trb-selection}
\begin{tabular*}{\linewidth}{@{\extracolsep{\fill}}lcccc@{}}
\toprule
\trbnew{Rule} & \trbnew{Mono, $150$\,ms} & \trbnew{Mono, $1.2$\,s} & \trbnew{Full, $150$\,ms} & \trbnew{Full, $1.2$\,s} \\
\midrule
\multicolumn{5}{@{}l}{\trbnew{\emph{Without gates}}}\\
\trbnew{Top benchmark score (score-first)} & \trbnew{4\,L} & 0.18 & \trbnew{$\ensuremath{-}3.45$ (4\,L)} & 0.18 \\
\trbnew{Probe predictor (\cref{fig:select}b)} & \trbnew{2\,S, 2\,L} & 0.48 \trbnew{(2\,S)} & \trbnew{$\ensuremath{-}3.14$ (4\,L)} & 0.48 \trbnew{(2\,L)} \\
\trbnew{Closed-loop champion (closed-loop-first)} & \trbnew{no score (4\,S)} & \trbnew{no score (4\,S)} & \trbnew{no score (4\,S)} & \trbnew{no score (4\,S)} \\
\midrule
\multicolumn{5}{@{}l}{\trbnew{\emph{Among the candidates that pass both gates}}}\\
\trbnew{Top benchmark score (Stage~4)} & \trbnew{--} & \trbnew{0.18} & \trbnew{0.00} & \trbnew{0.18} \\
\trbnew{Probe predictor} & \trbnew{--} & \trbnew{0.67} & \trbnew{1.69} & \trbnew{0.67} \\
\trbnew{Probe predictor with latency} & \trbnew{--} & \trbnew{0.65} & \trbnew{0.00} & \trbnew{0.65} \\
\trbnew{Probe predictor refit on these candidates} & \trbnew{--} & \trbnew{0.18} & \trbnew{0.00} & \trbnew{0.65} \\
\trbnew{Uniform random pick (expected)} & \trbnew{--} & \trbnew{0.66} & \trbnew{3.16} & \trbnew{4.34} \\
\trbnew{Oracle} & \trbnew{--} & \trbnew{0.00} & \trbnew{0.00} & \trbnew{0.00} \\
\bottomrule
\end{tabular*}
\end{table}

\textbf{Raw probe readings compare within a model, not across models.}
One photometric perturbation level is large for a model that barely uses the image and small for a model that leans on it
(\cref{fig:engagement}a). Normalizing each response by the model's
response to a constant gray frame reorders the three stateless
traditional stacks from LTF, TransFuser, DiffusionDrive to
TransFuser, DiffusionDrive, LTF: no model keeps its rank. LTF looks the most
sensitive and is the least, relative to how much it uses the image.
On the stateful architecture of RQ5, whose run-to-run noise lies above its measures (\cref{fig:engagement}b), the normalizer is not
established, since the response to a gray frame there is confounded with how well the model falls back on its own history.

\emph{Transferability estimators.} Estimators such as LEEP \citep{nguyen2020leep} and LogME \citep{you2021logme} score how well a source model's predictions or features fit labeled target data. In \textsc{Triage} they would be candidates for the Stage~4 proxy $\widehat{\Delta}_{\mathrm{dom}}$, not replacements for the gates: they assume the candidate can already consume the target inputs and answer in time, which is what the sensing and latency gates test, and here the gates alone decide feasibility in every profile (\cref{fig:select}a). They also need a feature or label space shared across candidates, which per-waypoint regressors, token-decoding vision-language-action models and video world models do not expose in the same form, and labeled target data, of which a deployer has little at the start. Within the top tier, where such a proxy would act, ten probe features and four predictor families already lose to the top score (\cref{fig:select}b); whether a transferability estimator does better there is left to future work.

\emph{Takeaway.} The gates do measurable selection work, and
latency's value lies entirely in the gate. Cheap probes place tiers
but cannot separate models of the same generation, so \trbnew{Stage~4 of} \cref{alg:triage}
\trbnew{is a gate-restricted benchmark ranking, with the probes scoped to tier placement, vetoes and sanity checks}. Normalize probe readings by each candidate's
response to complete input removal before comparing candidates,
and check that they exceed the model's run-to-run noise.

\RQ{8}\textbf{: Does the framework absorb new state-of-the-art
candidates?} \emph{Setup.} Four NAVSIM candidates joined after the pool was assembled. Two members of
NVIDIA's Alpamayo family had their sensor contracts checked, and the one admitted
was probed with a camera-count ablation. The ablation tests the model card's note that with a front camera only the model is ``expected to struggle to see cross-traffic in a right turn''. Two world models, WAM-Flow and
DriveVLA-W0, were evaluated in full: score reproduction on all 12{,}146 test tokens, four-city probes, both perturbation axes and latency.

\begin{table}[!ht]
\centering\small
\caption{\textbf{Camera-count ablation on Alpamayo-1.5 (RQ8).} Front-only against three cameras on identical frames, four cities pooled; turn frames: heading change above $2^\circ$ over $2$\,s.}
\label{tab:alpamayo}
\begin{tabular*}{\linewidth}{@{\extracolsep{\fill}}lcc@{}}
\toprule
Measure & front-only & three-camera\\
\midrule
Sign-diagnostic agreement, turn frames & 0.915 & 0.954\\
Temporal instability, turning frames (m) & 1.431 & 1.544\\
Temporal instability, straight frames (m) & 0.921 & 0.934\\
Low-light response, turning frames ($\gamma{=}2.6$) & 0.792 &
1.079\\
Latency p50 (ms, GPU not shared) & \multicolumn{2}{c}{959}\\
Probe readings (4 cities), frame failures & \multicolumn{2}{c}{8/8 sets, 0 failures}\\
\bottomrule
\end{tabular*}
\end{table}

\emph{Results.}
\textbf{Sensor and memory requirements decide the Alpamayo family before any GPU work.} Alpamayo~2 Super does not fit this platform on paper, consistent with its model card \citep{alpamayo2}: its reported $72{,}115$\,MiB
footprint ($70$\,GiB), documented for an $80$\,GB H100, does not fit
the $48$\,GB ($49{,}140$\,MiB) of the RTX A6000 deployment GPU, and its six-camera sensor contract fails on a monocular camera setup. Alpamayo-1.5 (10B, $24$\,GB, flexible cameras) was admitted:
eight city$\times$configuration sets of probe readings, zero frame failures, and a p50 latency of $959$\,ms, the VLA tier. Its camera-count
ablation (\cref{tab:alpamayo}) refines the model card's caution. The
front-only variant is not less self-consistent; it is slightly smoother on both turning and straight frames ($1.43$ against $1.54$\,m on turns, $0.92$
against $0.93$\,m on straight roads). Its agreement with the human
driver's turn direction is lower, $0.915$ against $0.954$ (the sign
diagnostic is defined on turning frames only; $771$ of $843$ and $809$
of $848$ frames). The caution is thus consistent with a cost in turn correctness, while plan stability is not degraded. The two channels are worth separating, and the probe readings do so.

\textbf{Both world models reproduce their published scores and RQ1's asymmetry.} Both
reproduced their published scores before entering the pool (WAM-Flow
$90.32$ against $90.3$; DriveVLA-W0 $90.30$ against $90.2$), both
landed in the VLA latency tier ($781$ and $485$\,ms), and both
replicated RQ1's asymmetry. The DriveVLA-W0 check also matched the published score to its checkpoint: the paper's $90.2$ corresponds to the query-based checkpoint with anchors, the one admitted here, while another released checkpoint, labeled with a PDMS of $87.2$, is a flow-matching variant. \cref{app:twins} records a related case: UniDriveVLA's two benchmark releases differ in architecture.

\emph{Takeaway.} A new candidate costs a sensor-contract check plus at
most a GPU-day, and infeasible ones are excluded without any GPU. Probes on recorded logs add detail that complements a model card, here separating turn correctness from plan stability. Checking sensor and memory requirements before downloading weights is the cheapest step in the procedure.

\subsection{Transfer to manipulation}
\label{app:ext-gen}

We transplant the sensor-contract and latency checks and the probes unchanged to a second embodiment, to test whether the driving findings of \cref{sec:experiments}
are properties of the framework or of driving.

\begin{table}[!ht]
\centering\small
\caption{\textbf{Sensor contracts of the manipulation pool (RQ9).} Octo is release 1.5. Sources: OpenVLA \citep{kim2024openvla}, $\pi_0$ \citep{black2024pi0}, SmolVLA \citep{shukor2025smolvla}, GR00T~N1.5 \citep{bjorck2025groot}, Octo \citep{octo2024}.}
\label{tab:manipcontracts}

\begin{tabularx}{\linewidth}{@{}ll>{\raggedright\arraybackslash}p{0.17\linewidth}>{\raggedright\arraybackslash}p{0.12\linewidth}>{\raggedright\arraybackslash}X@{}}
\toprule
Model & Venue & Cameras & Wrist cam & Action space\\
\midrule
OpenVLA-7B & CoRL'24 & 1 (third-person) & no & 7-DoF EEF delta,
256-bin discretized tokens, per-robot de-normalization\\
$\pi_0$ & arXiv'24 & 3 & yes & continuous actions padded to 32-d,
flow-matching chunk; base checkpoint ships no de-normalization
statistics\\
SmolVLA & arXiv'25 & 3 (generic slots) & platform-dependent & 6-d
SO-10x joint-space targets, checkpoint-stat de-normalized,
flow-matching chunk\\
GR00T N1.5 & arXiv'25 & embodiment-dependent (oxe\_droid: 2 ext.\ +
1 wrist) & head-dependent & embodiment-indexed continuous chunk
($\le$32-d), flow matching\\
Octo-small/base & RSS'24 & 1--2 (primary + optional wrist) &
optional slot & 7-d [dxyz, drpy, gripper] continuous, diffusion
head, per-dataset unnormalization (incl.\ bridge)\\
\bottomrule
\end{tabularx}
\end{table}

\begin{figure}[!ht]
\centering
\includegraphics[width=0.91\linewidth]{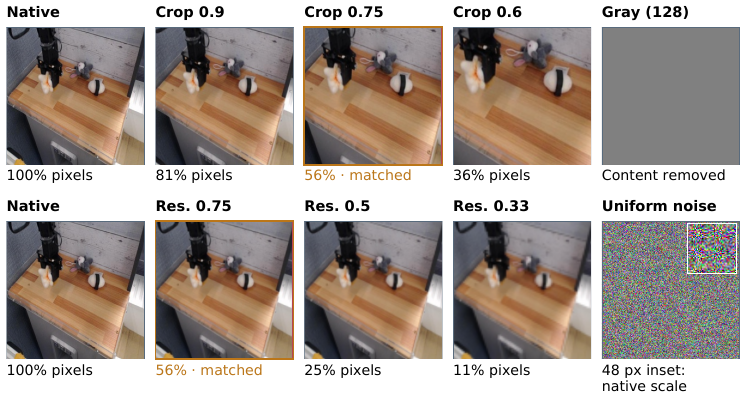}
\caption{\textbf{The same interventions on a manipulation frame} (Bridge V2, median-index frame of the $120$-frame probe subset). The ladders meet at \texttt{keep}~$=0.75$ (outlined), where \cref{fig:info}b is read.}
\label{fig:manipstimulus}
\end{figure}

\RQ{9}\textbf{: Do the instruments and the sensory asymmetry
transfer to a second embodiment?}
\emph{Setup.} The same checks and probes run unmodified on six
manipulation models (Octo-small/base, GR00T~N1.5, SmolVLA,
OpenVLA-7B, $\pi_0$; sensor contracts in \cref{tab:manipcontracts}), under offline inference on recorded real robot frames (\cref{fig:manipstimulus}).
We read action displacement in place of plan displacement, as
within-model ratios so heterogeneous action spaces compare validly;
a constant-action policy is the zero-response anchor. We test whether two driving findings carry over: a latency spread of at least two orders of magnitude, and the field-of-view asymmetry in most sighted models.

\emph{Results.}
\textbf{The sensory asymmetry transfers; the latency span does not.}
The six latency measurements (\cref{fig:portability}c) span $27.5$\,ms
(Octo-small) to $385$\,ms ($\pi_0$), a $14\times$ spread from a
$\sim$36\,Hz replanner to a $\sim$2.6\,Hz one. That falls short of two
orders of magnitude; driving's
five-decade spread came from blind anchors and multi-second reasoning
VLAs absent here. The asymmetry holds in all six models:
cropping the camera view moves the actions $1.6$--$2.4\times$ more than
downscaling it at the same pixel budget (\cref{fig:info}b), milder than
driving's $2.2$--$5.7\times$ but unanimous. The constant-action anchor
responds exactly zero at every perturbation level, every model's run-to-run displacement is exactly zero, these being stateless policies, and all
paired comparisons reproduce exactly.

\begin{figure}[!ht]
\centering
\includegraphics[width=0.95\linewidth]{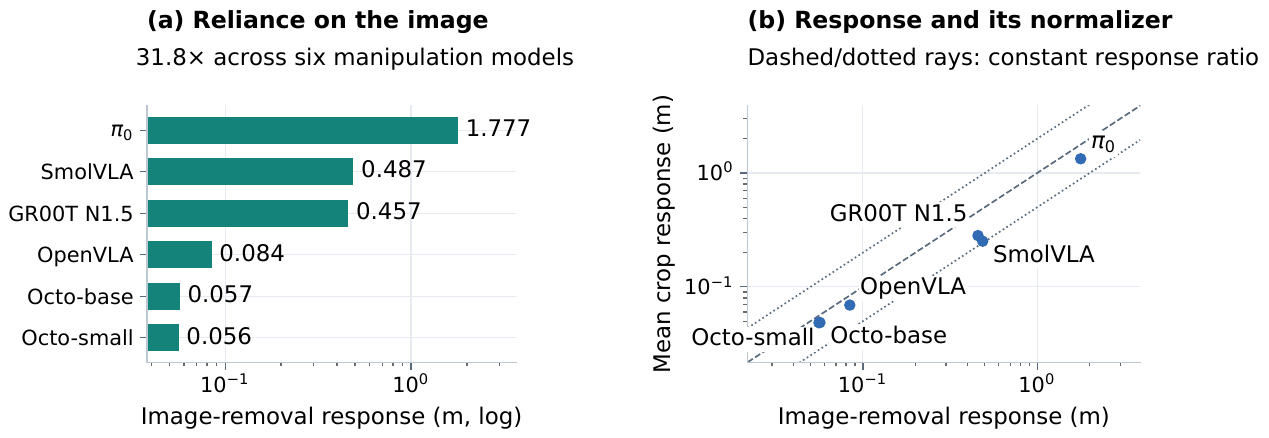}
\caption{\textbf{Reliance on the image and normalized responses (RQ9).} \textbf{(a)}~Response to image removal, which spans $31.8\times$ across the six models; the constant-action policy reads zero. \textbf{(b)}~Crop response against image-removal response: crop responses, averaged over the three crop levels, span $27.6\times$ across the six models, and normalized by the image-removal response they span $1.67\times$.}
\label{fig:crossembodiment}
\label{fig:engagementmech}
\end{figure}

\textbf{The probe's cross-model limitation transfers too.} Responses
to losing the image span $31.8\times$, from $0.056$ to $1.777$
(\cref{fig:crossembodiment}a), so a fixed perturbation level is a very different fraction of each model's signal. Ranking the six by raw field-of-view
response and by the same response normalized to each model's own
ablation gives near-opposite orders (\cref{fig:crossembodiment}b): $\pi_0$ drops from first to
fourth and Octo-small rises from sixth to first. This limitation is a property of the probe design, not of driving, and so is its repair.

\begin{figure}[!ht]
\centering
\includegraphics[width=0.91\linewidth]{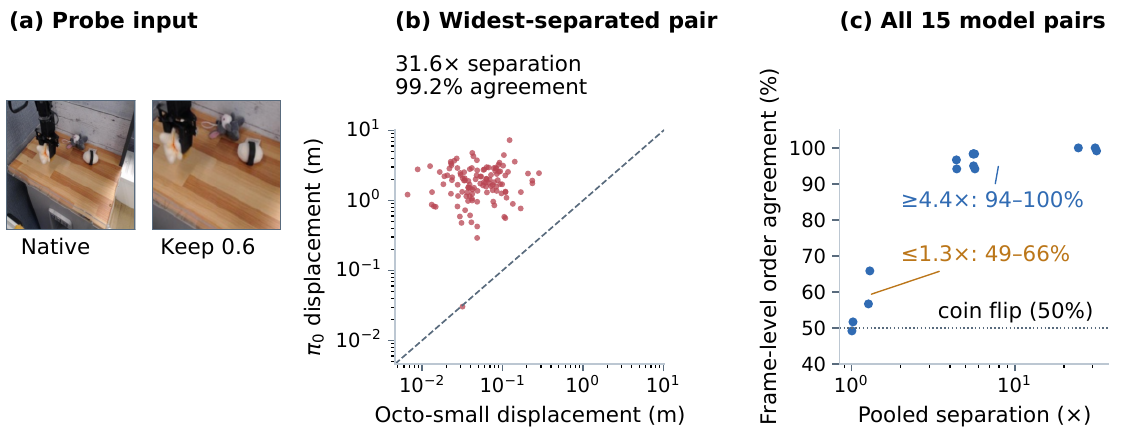}
\caption{\textbf{What the probe resolves frame by frame (RQ9).} \textbf{(a)}~The stimulus. \textbf{(b)}~The widest pair, whose mean crop responses at the strongest crop (keeping $60\%$) differ $31.6\times$, agrees with that order on $99.2\%$ of frames. \textbf{(c)}~All $15$ pairs: pooled separation against frame-level agreement.}
\label{fig:probelimit}
\end{figure}

\textbf{The probe resolves several-fold differences, not small ones.}
Frame by frame (stimulus in \cref{fig:probelimit}a), every pair of
models whose pooled crop responses differ by at least $4.4\times$
agrees with the pooled order on $94$--$100\%$ of the $120$ frames
(the widest pair in \cref{fig:probelimit}b), and every pair within
$1.3\times$ on only $49$--$66\%$, close to a coin flip
(\cref{fig:probelimit}c); the
full six-model raw ordering appears on $14$ frames and the normalized
one on none. This is the manipulation counterpart of RQ7, where probes
place tiers but cannot separate models of the same generation.

\emph{Takeaway.} The instruments and the central sensory asymmetry
are not driving artifacts: the field-of-view asymmetry holds in all six models, and the latency spread is $14\times$, short of driving's. Transplant the instruments unchanged and
validate the zero-response anchors before trusting any reading.
Normalize before comparing candidates: normalization rearranged the
ordering in both embodiments and nearly reversed it in
manipulation.

\FloatBarrier
\section{Measurement definitions and checks}
\label{app:instruments}

This appendix defines every quantity reported for the driving and manipulation experiments.

\paragraph{Frames and plans.} Open-loop measures use nuPlan logs at
2\,Hz. Each city contributes a ten-minute window of $1{,}200$ frames,
drawn with a fixed seed, and each model receives a four-frame history.
A plan $\xi_t$ is eight waypoints at $0.5$\,s spacing ($0.5$ to
$4$\,s) in the ego frame ($x$ forward, $y$ left); the
simulation-trained TransFuser outputs four (to $2$\,s), so its real-data counterpart is truncated to four whenever the two are compared. Baseline
plans, temporal instability, the sign diagnostic and tracking error
use every frame of the window; perturbations and the gray and noise inputs use every tenth frame, about $120$ per city. The \emph{probe readings} are the measures below, computed on one city's window by offline inference on the recorded logs. A four-city value is the mean of the four city means unless stated otherwise. Frames a model cannot process are kept with their reason and never imputed: of ReCogDrive's $4{,}800$ window frames, $146$ are missing because the logs carry a command class its deployment interface does not map ($134$ of them in one city) and $30$ because fewer than four frames of history precede them at a log's start.

\paragraph{Benchmark scores.} PDMS (NAVSIM v1.1) is computed per
token as $\mathrm{NC}\times\mathrm{DAC}\times(5\,\mathrm{EP}+5\,\mathrm{TTC}+2\,\mathrm{C})/12$
over a $4$\,s horizon, where no-collision (NC; $0.5$ for a static
object) and drivable-area compliance (DAC) act as multiplicative gates
on ego progress (EP), time to collision (TTC) and comfort (C). A
configuration's PDMS is the mean over tokens that returned a plan;
tokens with no plan are excluded, not scored zero. \emph{Regret}
(RQ7) is computed leave one city out: in each of the four folds, the
PDMS of the best candidate in the held-out city minus the PDMS of the
candidate the selector picks, averaged over folds. The candidate set is the whole candidate list, blind anchors included\trbnew{; \cref{tab:trb-selection} restricts it to each profile's feasible set}. Picking the top-scoring candidate means taking the configuration with the highest global PDMS.

\paragraph{Plan displacement and probe responses.} For a perturbation
$P_\rho$ at strength $\rho$ applied to image $I_t$, the displacement is the waypoint-averaged
distance between the plan on the perturbed frame and the plan on the
same frame unperturbed,
\begin{equation}
D(\rho) = \operatorname*{mean}_{t}\ \frac{1}{H}\sum_{j=1}^{H}
\bigl\lVert \xi_{t,j}\bigl(P_\rho(I_t)\bigr) - \xi_{t,j}(I_t)
\bigr\rVert_{2}.
\label{eq:probes-formal}
\end{equation}
Every camera the model consumes is perturbed, except that AutoVLA's probes perturb its front view only. The field-of-view crop
keeps $90$, $75$ and $60\%$ of each image dimension around the center
and resizes back; the resolution drop scales each dimension to $75$,
$50$ and $33\%$ and back; darkening applies gamma $1.8$ and $2.6$ with
Gaussian noise of standard deviation $4$ gray levels; compression uses
JPEG quality $50$ and $20$. The matched-budget ratio is
$\kappa = D_{\mathrm{fov}}(0.75)/D_{\mathrm{res}}(0.75)$, the two perturbation levels that keep $56\%$ of the pixels, taken as a ratio of four-city
means (\cref{fig:info}a); no city ratio falls below $1.9$. For
darkening and compression the reported \emph{response} is the area
under the displacement-versus-severity curve, with severity scaled to $[0,1]$ from the unperturbed reference (gamma $1$, JPEG quality $90$):
$A_\gamma = \tfrac{1}{2}\,D(1.8) + \tfrac{1}{4}\,D(2.6)$ and
$A_{\mathrm{jpeg}} = \tfrac{1}{2}\,D(50) + \tfrac{3}{14}\,D(20)$. The plain average of the responses at the two severities, $\bar D$, is used for normalization (below) and for comparison with the run-to-run noise. In manipulation, $\xi$ is the first step of the action
chunk over the pose dimensions, gripper excluded, in each model's
native units, averaged over $120$ evenly spaced Bridge frames; the
constant-action policy returns a fixed action and so reads zero.

\paragraph{Temporal instability.} For consecutive frames of one log,
$0.5$\,s apart, the earlier plan is moved into the later ego frame
with the logged ego motion $E_{t\to t+1}$, and waypoint $j{+}1$ of
the earlier plan is compared with waypoint $j$ of the later one, since
both refer to the same future instant:
\begin{equation}
\mathrm{TI} = \operatorname*{mean}_{t}\ \frac{1}{H-1}\sum_{j=1}^{H-1}
\bigl\lVert E_{t\to t+1}\,\xi_{t,j+1} - \xi_{t+1,j} \bigr\rVert_{2}.
\end{equation}
RQ5's $8.6$--$10.3\times$ is the per-city ratio of TI between the
simulation-trained and real-data TransFuser, and $1.5$--$4.9\times$ is
the range of the per-city ratios of the darkening and compression
responses.

\paragraph{Tracking error and turning frames.} Tracking error is the
distance, in the ego frame, between the plan's waypoint at $2$\,s and
the logged ego position $2$\,s later, over every window frame with a
logged future. A turning frame is one whose logged heading changes by
at least $15^\circ$ over the next $2$\,s. RQ2's values pool frames
across cities; RQ6's TransFuser curve ($8.0$ to $1.4$\,m) is the mean
of city means, while ReCogDrive's tracking ratios pool frames.
Mean trajectory error in RQ6 averages over the waypoints to $2$\,s instead of taking the $2$\,s endpoint. Turning frames are unevenly spread: $144$, $62$, $46$ and $0$ in Singapore, Pittsburgh, Boston and Las Vegas for TCP's windows, and $144$, $59$, $19$ and $0$ for ReCogDrive's, so no per-city turn reading is made for Las Vegas.

\paragraph{Repair and closure.} RQ2 uses four settings, zero-shot (zs) or
fine-tuned (ft) and native (nat) or cropped (crop), each scored by
turn-frame tracking error $e$: damage $=e_{\mathrm{zs,crop}}-e_{\mathrm{zs,nat}}$,
repair $=e_{\mathrm{zs,nat}}-e_{\mathrm{ft,nat}}$, residual
$=e_{\mathrm{ft,crop}}-e_{\mathrm{ft,nat}}$, and compensated fraction
$=1-\mathrm{residual}/\mathrm{damage}$; the share of the zero-shot error of a setting $s$ that fine-tuning removes (the $73$--$76\%$ of RQ2) is
$\eta_{s}=(e_{\mathrm{zs}}-e_{\mathrm{ft}})/e_{\mathrm{zs}}$.
The criterion is a positive residual of at least $0.25\times$ the repair. For RQ6, with zero-shot value $m(0)$, dose-$N$
value $m(N)$ and real-data reference $m_{\mathrm{ref}}$, all four-city
means,
\begin{equation}
\mathrm{cl}(N) = \frac{m(0) - m(N)}{m(0) - m_{\mathrm{ref}}}.
\label{eq:closure}
\end{equation}
The criterion requires at least two of the three measures
(darkening response, compression response, TI) to fall monotonically
from a positive gap and reach $\mathrm{cl}(20\mathrm{k})\ge0.5$; only
TI did ($0.753$). A \emph{spike} is a response at a dose divided by
its zero-shot value; on ReCogDrive the spike threshold is a darkening or compression ratio of at least $2.0$ at 1k or 5k frames.

\paragraph{Response to a gray frame and run-to-run noise.} To measure how much a plan relies on the image, we replace every camera view with constant gray, or with seeded uniform noise (rightmost column of \cref{fig:manipstimulus}), and measure the displacement $D_{\mathrm{gray}}$ with the same metric
and frames; LiDAR is left intact. Dividing a response by it makes
responses comparable across models: RQ7's normalized sensitivity is
$\bar D_\gamma/D_{\mathrm{gray}}$ per city, averaged over cities, and RQ6's \emph{reliance on the image} at a dose is
$D_{\mathrm{gray}}$ at that dose over $D_{\mathrm{gray}}$ at zero
shot, as four-city means. \emph{Run-to-run noise} is the mean displacement of calls two to six from the first, over six consecutive calls on the same unperturbed input (four in manipulation); a model that keeps internal state across calls shows non-zero run-to-run noise, and a perturbation reading below it cannot be
attributed to the perturbation (RQ5). The gray-frame response behaves monotonically on models that plan each
frame independently, where removing the whole image moves the plan
further than any perturbation level; on a model that keeps state it does not,
because such a model falls back on its own history when the image
carries nothing.

\paragraph{Band of chance variation for reliance on the image.} The null control repeats the
smallest dose with only the training seed changed, giving per-city
ratios of $0.795$, $0.892$, $0.941$ and $0.846$ (mean $m=0.868$). The
seed half-width is twice their standard error, $0.063$; the city
half-width is twice the zero-shot coefficient of variation of
$D_{\mathrm{gray}}$ across cities divided by $\sqrt4$, $0.147$; the
two combine in quadrature to $h=0.160$. The band is
$[\min(1,m)-h,\ \max(1,m)+h] = [0.709, 1.160]$, anchored at $1$ on the
upper side. It is applied to four-city means; per-city values are descriptive only.

\paragraph{Sign diagnostic and blind anchors.} The sign diagnostic
takes the window frames whose logged heading changes by more than
$2^\circ$ over the next $2$\,s and whose planned final lateral offset
exceeds $0.05$\,m, and reports the share on which the plan turns the
same way as the logged heading. It is recorded with every set of probe readings
and enforced as a gate at $0.8$ on the ground-truth labels before
every fine-tuning run and on Alpamayo's readings; the turn-frame rows of
\cref{tab:alpamayo} use this $2^\circ$ definition. Blind anchors
(constant velocity and the ego-status MLP) must respond exactly zero
to every image perturbation, which verifies that a response measures reliance on the image rather than pipeline noise.

\paragraph{What is recorded per candidate, and latency.} For each candidate we record its training ecosystem, sensor contract (the cameras its inference code consumes and the preprocessing it applies), license and per-frame latency. Latency is measured on the deployment
GPU with no other compute process present: five warm-up calls, then $200$ consecutive frames, each timed from agent
input to returned plan (for models that read images from disk,
decoding falls inside); p50 and p99 are percentiles of these samples.
Image acquisition, transport, control and actuation lie outside the
timed boundary\trbnew{, so these values are lower bounds on the end-to-end latency $\hat\tau_c$
of the latency gate (\cref{app:timegap}), which only the deployment
platform can measure}. Two latency measurements differ: TCP's uses $279$ samples after
$20$ warm-ups and includes its JPEG preprocessing, and ORION's $300$
samples after $20$ warm-ups and includes its controller. For manipulation models the latency measurement times the whole action chunk as one call.

\paragraph{Closed-loop metrics.} In Bench2Drive, route completion is
the share of the route driven; the collision route rate is the
fraction of route-and-seed runs with at least one collision with a
vehicle, pedestrian or static object; the driving score is route completion times Bench2Drive's infraction penalty, averaged over routes. A route is
\emph{stuck} when the vehicle stays below $0.1$\,m/s for $60$\,s, and
ends in a \emph{scenario timeout} when it runs past $4{,}000$ simulation steps
($200$\,s). RQ3's seed-noise bound is twice the larger of the native and cropped runs' seed standard deviation of five-route mean
completion ($8.4$ points for TCP, $18.0$ for SimLingo). In RQ5 a
configuration's spread is the range of its seed-mean completion
across the six environments, its seed noise the median over
environments of the per-environment seed standard deviation, and an
environment pair changes the order if the completion ranking of the
four native configurations differs. An occlusion of severity $s$
blacks out a rectangle of $s\%$ of each image dimension, so $s^2$ of
the area, at a new random position for every camera and frame.

\paragraph{Deadlines and stale inputs (RQ4).} Under a deadline, each
planning call is charged the next latency sample from the model's latency record, in order; if it exceeds the deadline, the new control is
discarded and the last on-time control is applied again (hold), and
before any control has arrived the vehicle brakes. On the frame-drop
axis each camera independently starts a burst with probability $0.01$ per step, lasting a uniformly drawn $2$ to $20$ or $2$ to $60$ steps
(up to $1$ or $3$\,s at $20$\,Hz); during a burst the camera returns
its last frame, while the model keeps planning every step with fresh
speed and position and the controller tracks with fresh vehicle state.
The asynchronous wrapper delivers each plan $\lceil\tau/50\,\mathrm{ms}\rceil$ steps after it was computed and always uses the newest plan.

\FloatBarrier
\section{Reproducibility details}
\label{app:protocols}

\paragraph{Fixed versions and admission.} Closed loop runs on CARLA 0.9.15 with Bench2Drive at a fixed version, with every metric definition fixed before the first run. Open loop runs the NAVSIM v1.1
evaluator at a fixed version on the \texttt{navtest} split (136 logs, 12{,}146 tokens), with fixed probe definitions and a fixed sampling seed. A configuration with a published score was admitted
only after that score was reproduced under that evaluator (for example EgoStatusMLP $65.5165$ against the published $65.6$). ORION's reproduction on Bench2Drive's ten-route Dev10 subset came out four points below the published mean ($71.19$ against $75.19$), beyond the two-point admission threshold; the whole gap is one collision on one route, whose score ranges from $60$ to $100$ over two further seeds, so ORION was admitted by exception.

\paragraph{Data handling.} Closed-loop and open-loop measurements are
never pooled; published scores never stand in for missing measurements;
missing measurements are never imputed; and fine-tuned checkpoints
never receive official benchmark scores, because their fine-tuning
pool shares the split. Fine-tuning pools are log-level disjoint from all evaluation windows, checked automatically before training, ground-truth waypoints pass
a 20-frame sign diagnostic before any gradient step, and every dose
starts from zero-shot weights.

\FloatBarrier
\section{Checkpoint pairs that differ only in training data}
\label{app:twins}

\begin{table}[t]
\centering\small
\caption{\textbf{Public checkpoints of one architecture in both ecosystems are rare.} The open driving pool by publisher. ``Sim'': trained in CARLA/Bench2Drive; ``real'': trained on nuScenes \citep{caesar2020nuscenes} or nuPlan. Other candidates: MindDrive \citep{fu2026minddrive}, DriveWorld-VLA \citep{jia2026driveworldvla}, SparseDriveV2 \citep{sun2026sparsedrivev2}.}
\label{tab:twinsearch}
\begin{tabularx}{\linewidth}{@{}>{\raggedright\arraybackslash}p{0.19\linewidth}>{\raggedright\arraybackslash}p{0.12\linewidth}cc>{\raggedright\arraybackslash}X@{}}
\toprule
Candidate & Family & Sim & Real & Status for a training-data contrast \\
\midrule
UniDriveVLA & VLA & \checkmark & \checkmark & both sides public; architectures differ (below); not used \\
DriveStack-VLA & VLA & -- & -- & scores on both; weights not public \\
ORION, MindDrive & VLA & \checkmark & & simulation side only \\
ReCogDrive & VLA & & \checkmark & real side only (sim release announced) \\
AutoVLA, Curious-VLA & VLA & & \checkmark & real side only \\
DriveVLA-W0, WAM-Flow & world model & & \checkmark & real side only \\
DriveWorld-VLA & VLA & & \checkmark & real side only \\
SparseDriveV2 & vision only & \checkmark & \checkmark & both sides public; not a VLA; not checked \\
LEAD / TFv6 & traditional & \checkmark & \checkmark & both sides public;
modality symmetry unresolved, and one description indicates real panoramic data in the simulation side's pre-training, so the separation of training data could not be confirmed; not used \\
\midrule
TransFuser & traditional & \checkmark & \checkmark & controlled pair used in
RQ5 \\
UniAD & traditional & \checkmark & \checkmark & controlled pair used in RQ5
\\
\bottomrule
\end{tabularx}
\end{table}

RQ5 measures how the training data changes a model. That needs one
architecture with public checkpoints trained in simulation and on real
data and otherwise alike. We searched the open driving pool by
publisher, on code hosts and weight hubs, rather than by model name
(\cref{tab:twinsearch}). Five architectures publish both sides.
TransFuser and UniAD are the controlled pairs used in RQ5. LEAD/TFv6 \citep{nguyen2026lead} is not used, because one of its descriptions indicates that the simulation side's pre-training included real panoramic data, so we could not confirm that the two sides differ in training data. SparseDriveV2 also publishes both sides but is not a vision-language-action model, and whether its two checkpoints share one architecture was not checked. UniDriveVLA is the only vision-language-action model with both sides public. DriveStack-VLA \citep{wang2026drivestack} reports scores on both benchmarks; its weights are not public, so it could not be included.

\paragraph{UniDriveVLA's two checkpoints differ in architecture.} Its two public checkpoints \citep{unidrivevla2026} share sensor contracts, preprocessing, model class, planning head, planning objective and vision-language co-training corpus, but not an architecture: comparing parameter names and shapes, $260$ tensors exist on one side only and $52$ differ in shape ($2{,}809$ shared), adding up to an occupancy branch, an ego-history input pathway, twelve extra decoder layers and a different sampling geometry in the ego branch. The UniAD pair, by contrast, has no tensor on one side only and differs only in the counts and widths of task-head outputs ($35$ shape mismatches in $2{,}457$ shared tensors), so it is used as a controlled pair. Releasing a separately adapted model for each benchmark is common practice. We list the differences because a training-data contrast needs a controlled pair, which these two checkpoints were not released to be.

\end{document}